%% file: main.tex
\documentclass[twoside]{article}

\usepackage[preprint]{aistats2026}

\usepackage[round]{natbib}

\usepackage[utf8]{inputenc} % allow utf-8 input
\usepackage[T1]{fontenc}    % use 8-bit T1 fonts
\usepackage{hyperref}       % hyperlinks
\usepackage{url}            % simple URL typesetting
\usepackage{booktabs}       % professional-quality tables
\usepackage{nicefrac}       % compact symbols for 1/2, etc.
\usepackage{microtype}      % microtypography
\usepackage{graphicx}
\usepackage{pgffor}
\usepackage{doi}
\usepackage{xfrac}
\usepackage{xcolor}
\usepackage{amsfonts}
\usepackage{amsmath}
\usepackage{bm}
\usepackage{bbold}
\usepackage{bbm}
\usepackage{subcaption}
\usepackage{array}
\usepackage{algorithm}
\usepackage{algpseudocode}
\usepackage{multirow}
\usepackage{tikz}
\usepackage{mwe}
\hypersetup{
    colorlinks=true,
    linkcolor=blue,
    filecolor=magenta,
    urlcolor=blue,
    citecolor=blue,
    }
\newcommand{\thb}{\boldsymbol{\theta}}
\newcommand{\ub}{\mathbf{u}}
\newcommand{\yb}{\mathbf{y}}
\newcommand{\vb}{\mathbf{v}}
\newcommand{\jac}{\mathbf{J}}
\newcommand{\data}{\mathbf{y}^o}
\newcommand{\accregioni}{C_{\epsilon}^i}

\newcommand{\E}{\mathbb{E}}

\newcommand{\rromc}{R2OMC}

\newcommand{\NPE}{NPE}

\newcommand{\CtwoST}{\texttt{C2ST}}
\newcommand{\mask}{\mathbf{m}_\tau}

\newcommand{\approptoinn}[2]{\mathrel{\vcenter{
  \offinterlineskip\halign{\hfil$##$\cr
    #1\propto\cr\noalign{\kern2pt}#1\sim\cr\noalign{\kern-2pt}}}}}

\newcommand{\imagerow}[3]{% #1 = prefix, #2 = label
  \raisebox{-.5\height}{\includegraphics[width=.08\textwidth]{./figures/#1/fig_0_#2.png}} &
  \raisebox{-.5\height}{\includegraphics[width=.08\textwidth]{./figures/#1/fig_1_#2.png}} &
  \raisebox{-.5\height}{\includegraphics[width=.08\textwidth]{./figures/#1/fig_2_#2.png}} &
  \raisebox{-.5\height}{\includegraphics[width=.08\textwidth]{./figures/#1/fig_3_#2.png}} &
  \raisebox{-.5\height}{\includegraphics[width=.08\textwidth]{./figures/#1/fig_4_#2.png}} &
  \raisebox{-.5\height}{\includegraphics[width=.08\textwidth]{./figures/#1/fig_5_#2.png}} &
  \raisebox{-.5\height}{\includegraphics[width=.08\textwidth]{./figures/#1/fig_6_#2.png}} &
  \raisebox{-.5\height}{\includegraphics[width=.08\textwidth]{./figures/#1/fig_7_#2.png}} &
  \raisebox{-.5\height}{\includegraphics[width=.08\textwidth]{./figures/#1/fig_8_#2.png}} &
  \raisebox{-.5\height}{\includegraphics[width=.08\textwidth]{./figures/#1/fig_9_#2.png}} &
  \textbf{#3} \\
}

\begin{document}


% \runningtitle{I use this title instead because the last one was very long}


\twocolumn[

\aistatstitle{Fast and Robust Simulation-Based Inference With Optimization Monte Carlo}

\aistatsauthor{ Vasilis Gkolemis \And Christos Diou \And Michael U.\ Gutmann }

\aistatsaddress{ Harokopio University of Athens \\ ATHENA Research Center \And  Harokopio University of Athens \And University of Edinburgh } ]

\begin{abstract}
Bayesian parameter inference for complex stochastic simulators is challenging due to intractable likelihood functions.
Existing simulation-based inference methods often require large number of simulations and become costly to use in high-dimensional parameter spaces or in problems with partially uninformative outputs.
We propose a new method for differentiable simulators that delivers accurate posterior inference with substantially reduced runtimes.
Building on the Optimization Monte Carlo framework, our approach reformulates inference for stochastic simulators in terms of deterministic optimization problems.
Gradient-based methods are then applied to efficiently navigate toward high-density posterior regions and avoid wasteful simulations in low-probability areas.
A JAX-based implementation further enhances the performance through vectorization of key method components.
Extensive experiments, including high-dimensional parameter spaces, uninformative outputs, multiple observations and multimodal posteriors show that our method consistently matches, and often exceeds, the accuracy of state-of-the-art approaches, while reducing the runtime by a substantial margin.
\end{abstract}

\section{Introduction}

% DONE: (i) General intro to LFI (ii) SBI is challenging when model becomes complex
Parametric stochastic simulators are widely used to model complex processes across biology, physics, health sciences, and many other disciplines.
As simulators become more realistic, they also become more complex, often involving many free parameters and high-dimensional outputs.
This complexity, in turn, poses significant challenges for parameter inference.

% Backgroung in SBI (existing solutions) with a special focus on neural-based methods
A variety of methods, collectively referred to as Likelihood-Free (LFI) or Simulation-Based (SBI) Inference methods \citep{lintusaari2017fundamentals, sisson2018handbook, cranmer2020frontier, deistler2025simulation, arruda2025diffusion}, have been developed to infer simulator parameters.
These include Approximate Bayesian Computation \citep[ABC,][]{Marin2012}, summary statistic-based approaches \citep{Chen2021a, Chen2023}, parametric modeling \citep{wood2010statistical, price2018bayesian, pacchiardi2024generalized}, ratio estimation \citep{thomas2022likelihood, durkan2020contrastive}, and surrogate model optimisation \citep{gutmann2016bayesian, jarvenpaa2025surrogate}.
Recently, neural-based methods have become prominent, with deep neural networks used to estimate the posterior \citep{greenberg2019automatic, papamakarios2016fast, radev2020bayesflow, wildberger2023flow} or the likelihood \citep{papamakarios2019sequential} function.
State-of-the-art methods include Simformer~\citep{gloeckler2024allinone} and tabular-foundation-model SBI~\citep{vetter2025effortless}, which achieve high accuracy by learning arbitrary conditionals of the joint distribution.

% Neural-based methods are computationally expensive    
Neural-based approaches have greatly broadened the applicability of SBI to complex problems.
However, they are data-intensive, as they require large volumes of simulated datasets to train their neural estimators.
Since both data generation %, using calls to the simulator, 
and neural network training are computationally costly, these methods often incur significant runtimes, which in turn restricts experimental flexibility and hinders practical deployment.

\begin{figure*}[ht]
  \centering
  \setlength{\tabcolsep}{0.1pt} % Adjust column spacing
  \renewcommand{\arraystretch}{.6} % Adjust row spacing
  \begin{tabular}{c *{8}{c}} % First column for row titles
    % Column titles (empty first cell for row labels)
    & \scriptsize{Reference} & \scriptsize{R2OMC} & \scriptsize{NPE-low} & \scriptsize{NPE-high} & \scriptsize{BayesFlow-low} & \scriptsize{BayesFlow-high} & \scriptsize{FMPE-low} & \scriptsize{FMPE-high} \\

    % First row
    \rotatebox{90}{\scriptsize{\quad Simple}} &
    \includegraphics[width=.12\textwidth]{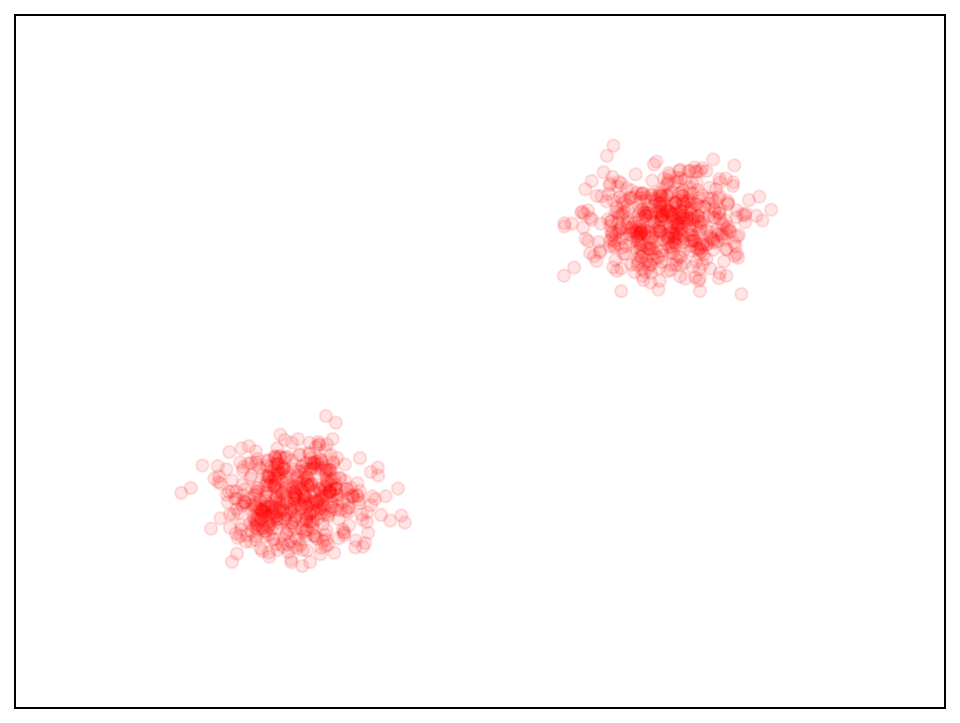} &
    \includegraphics[width=.12\textwidth]{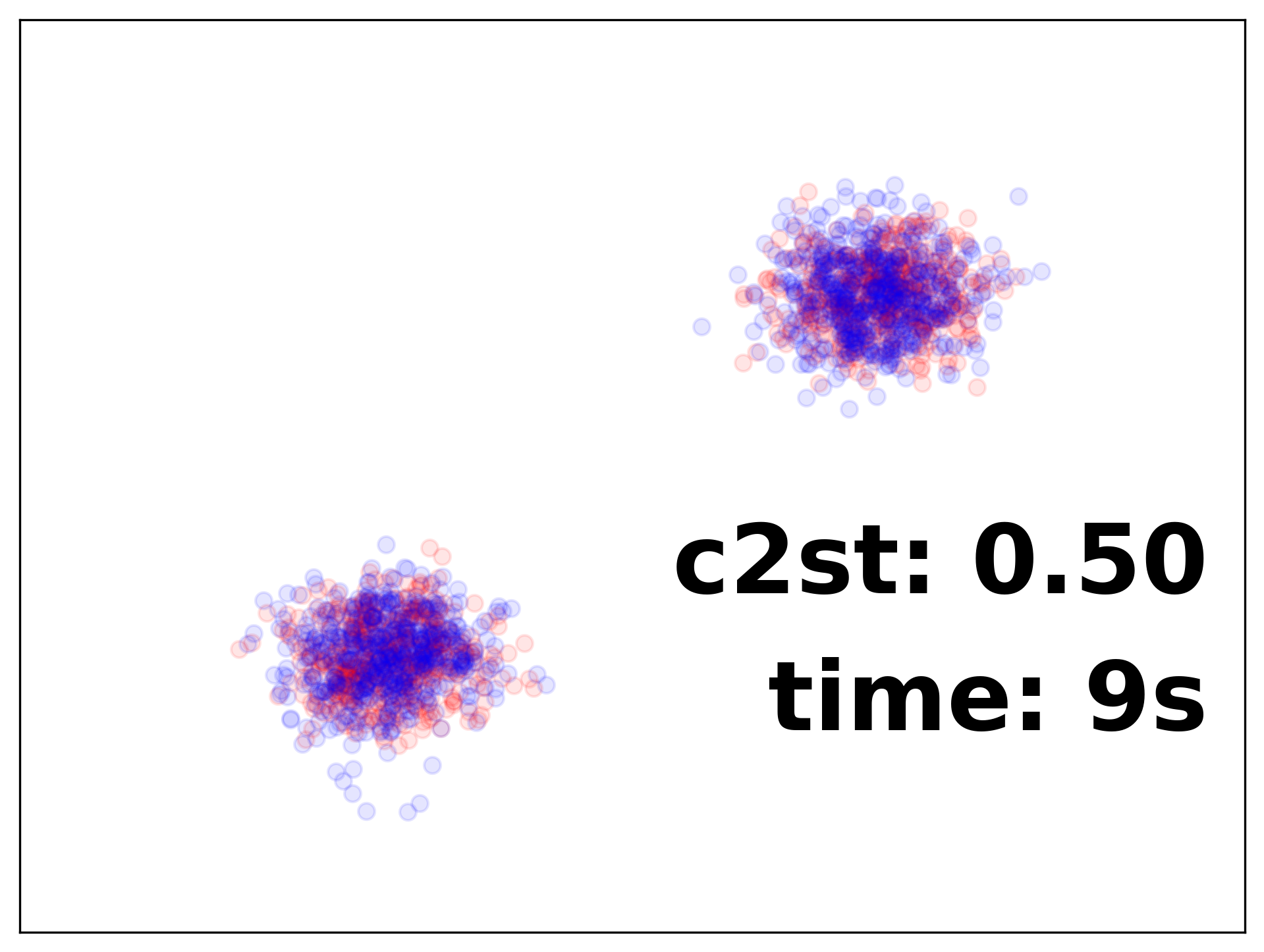} &
    \includegraphics[width=.12\textwidth]{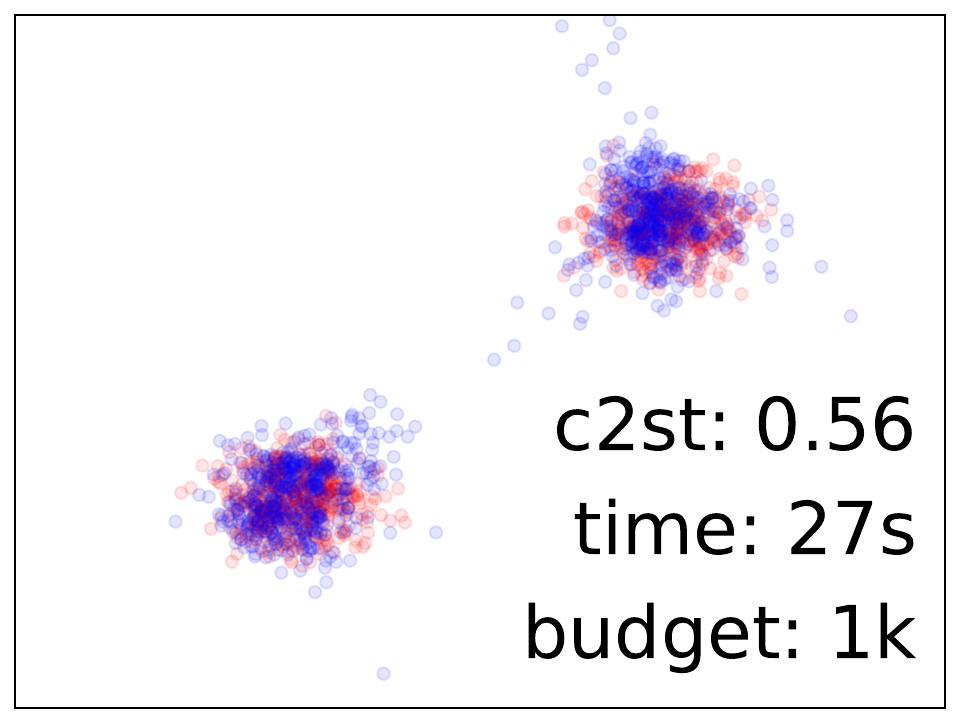} &
    \includegraphics[width=.12\textwidth]{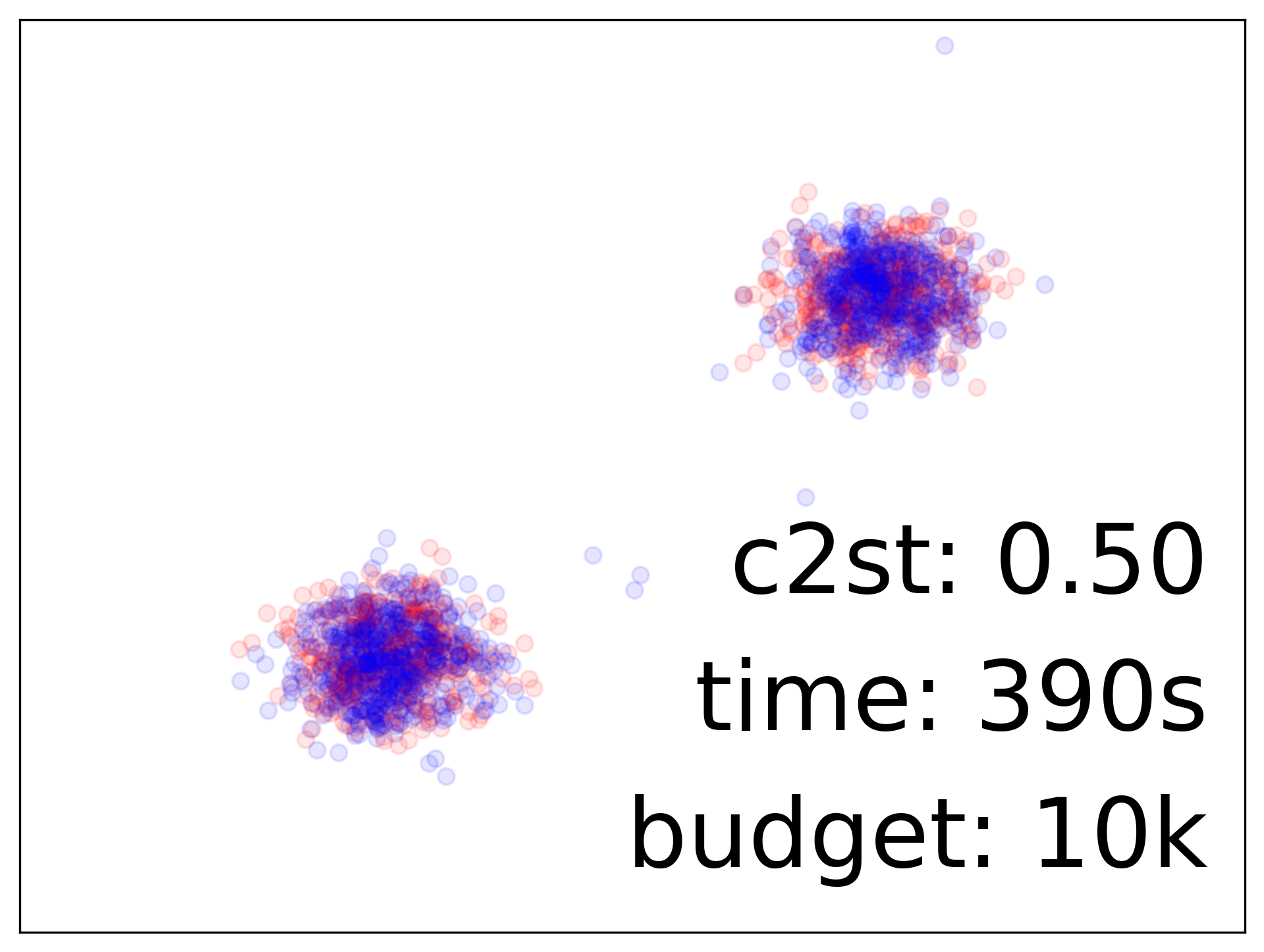} &
    \includegraphics[width=.12\textwidth]{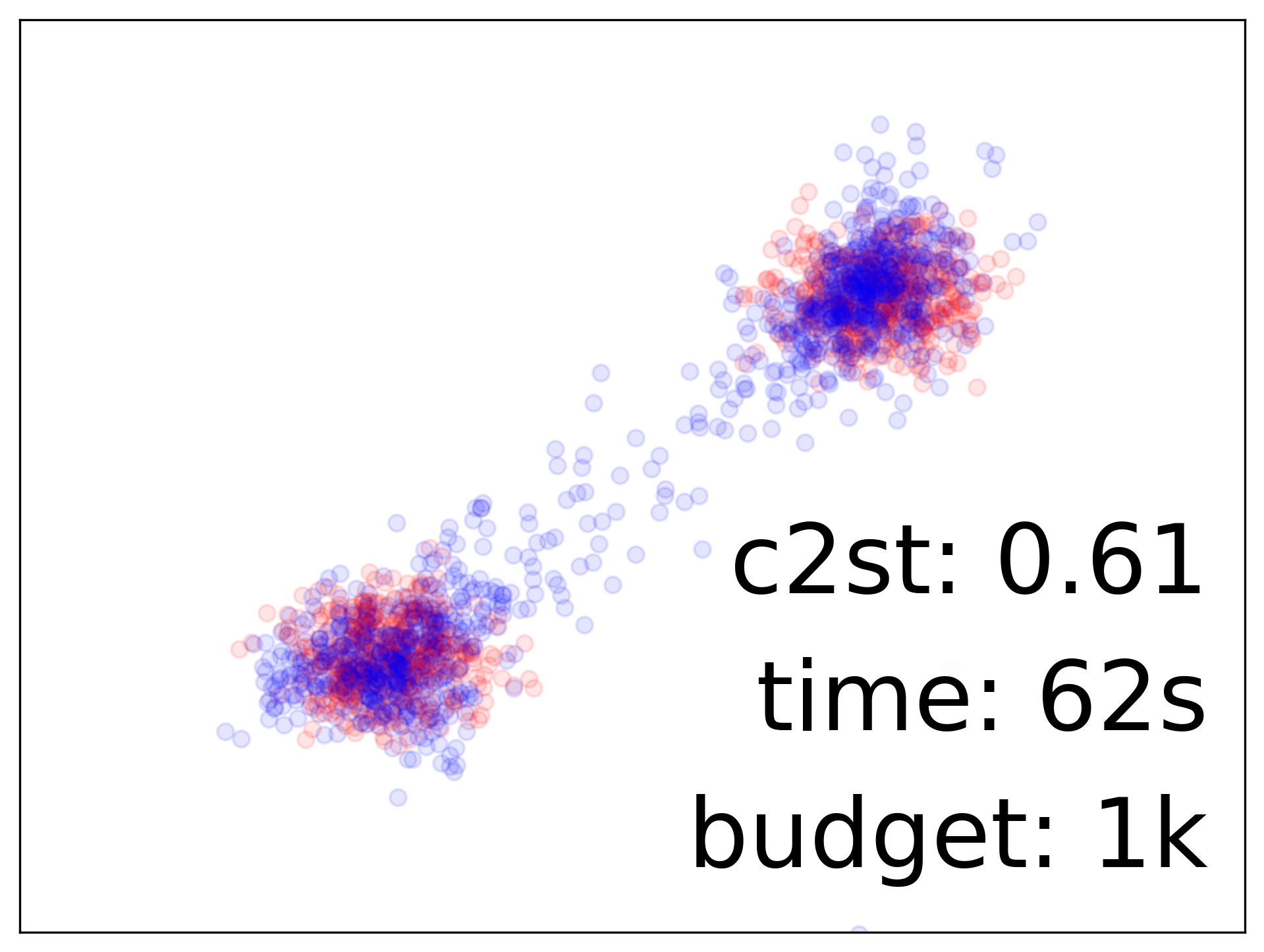} &
    \includegraphics[width=.12\textwidth]{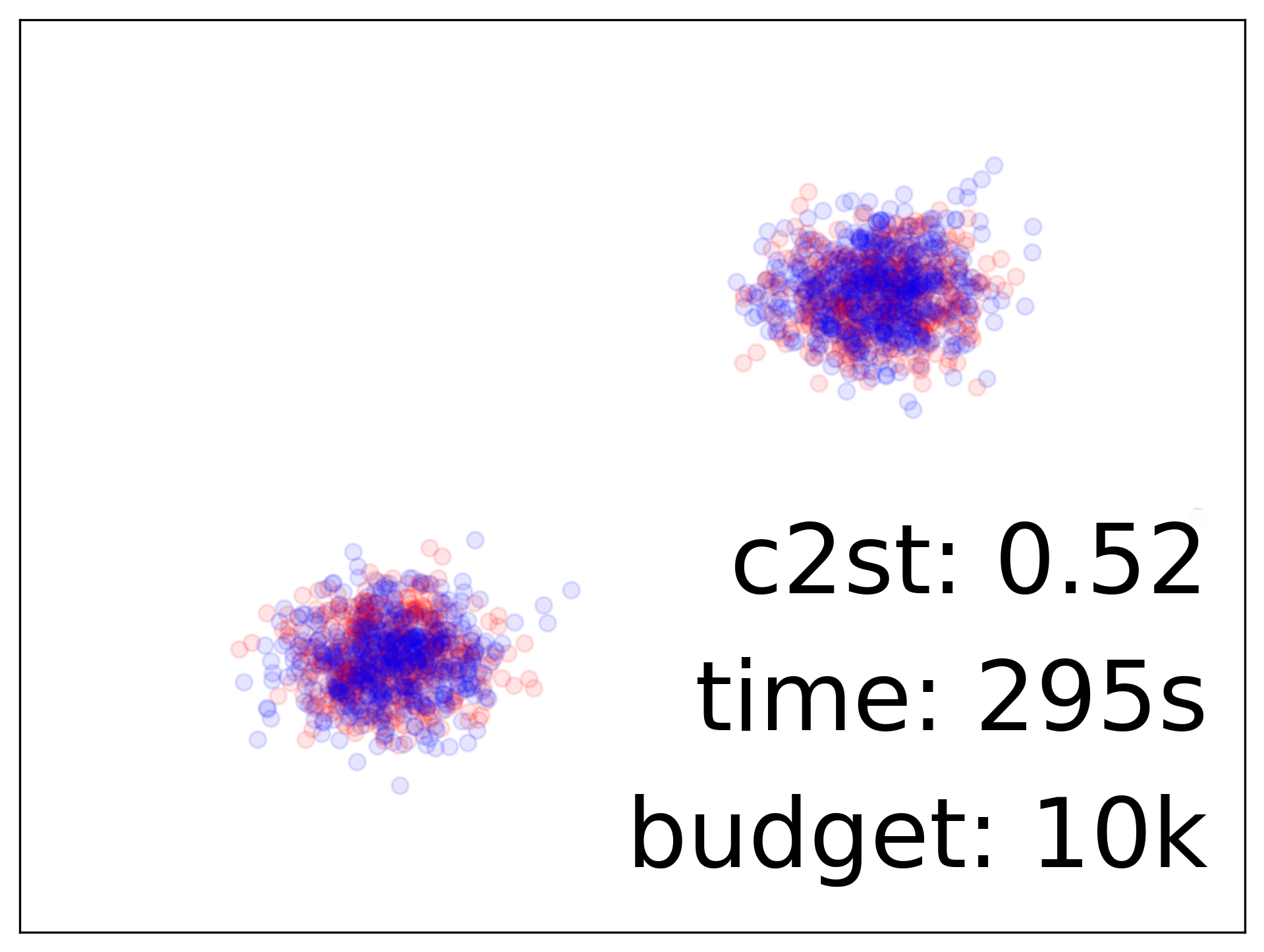} &
    \includegraphics[width=.12\textwidth]{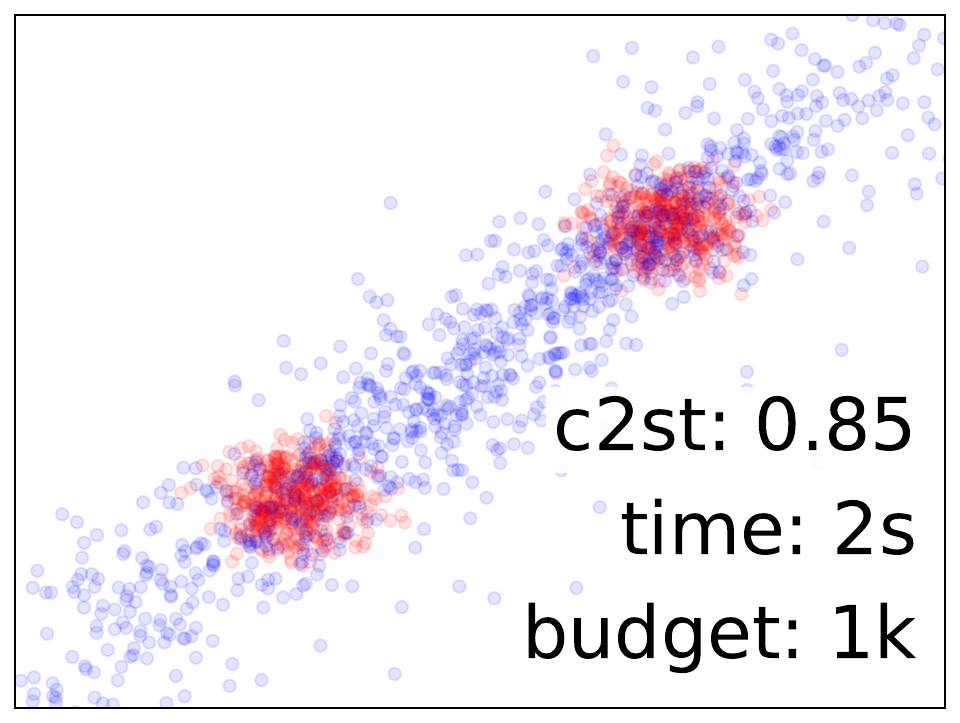} &
    \includegraphics[width=.12\textwidth]{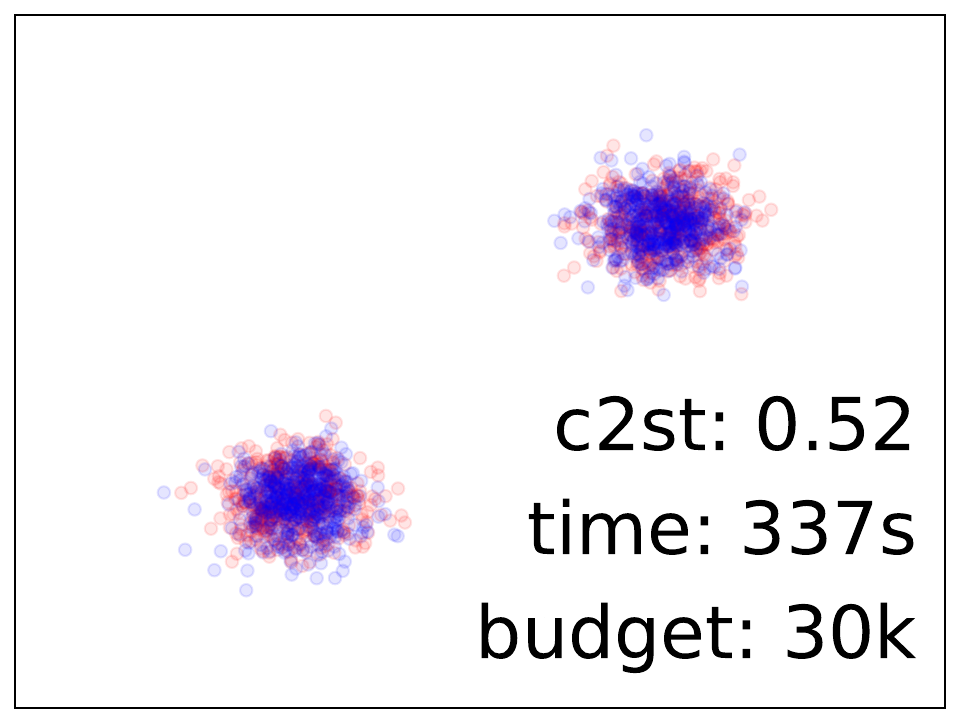} \\
    % Second row example
    \rotatebox{90}{\scriptsize{Distractors}} &
    \includegraphics[width=.12\textwidth]{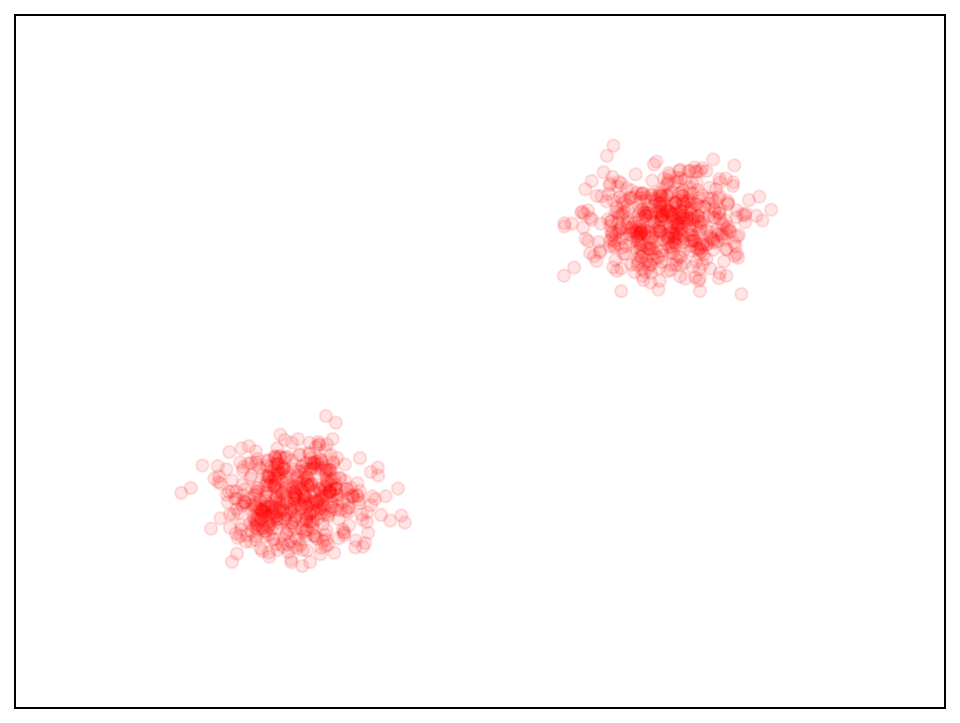} &
    \includegraphics[width=.12\textwidth]{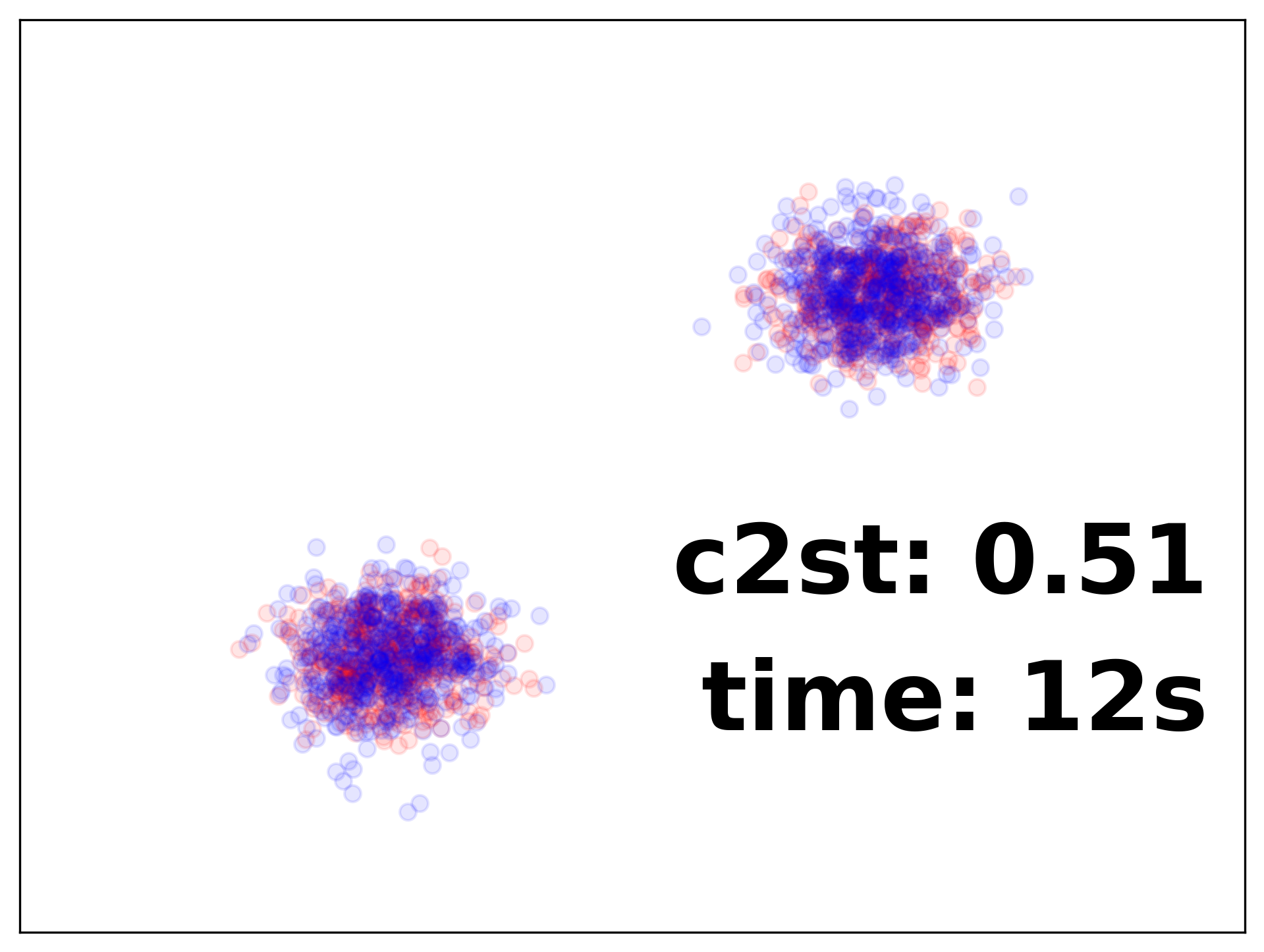} &
    \includegraphics[width=.12\textwidth]{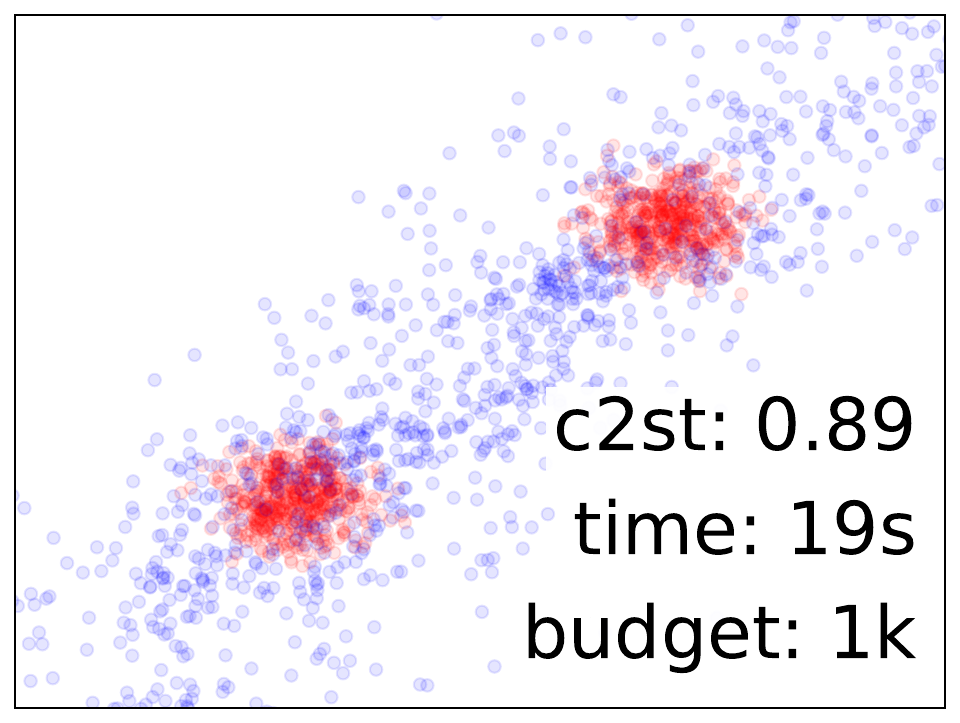} &
    \includegraphics[width=.12\textwidth]{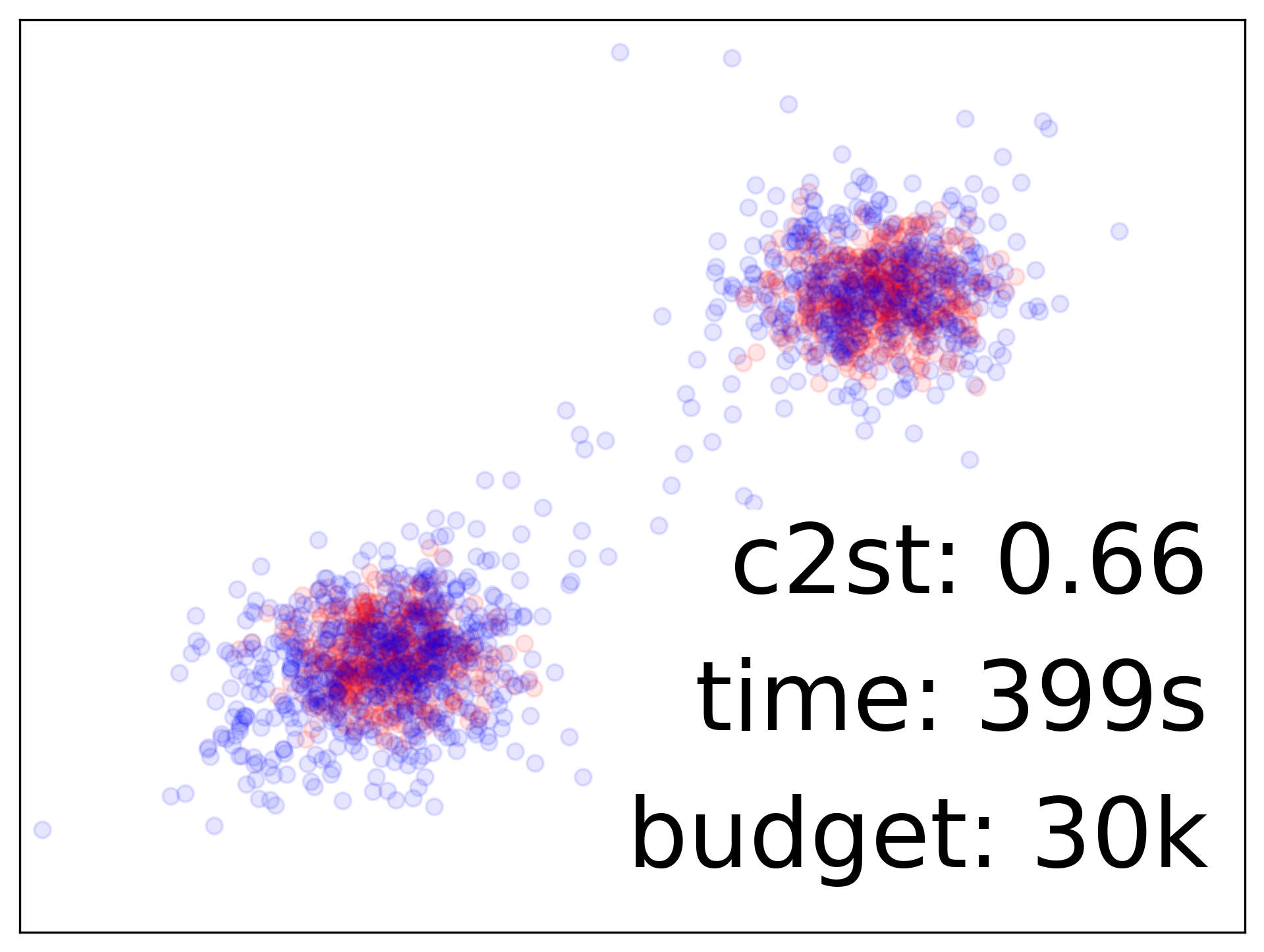} &
    \includegraphics[width=.12\textwidth]{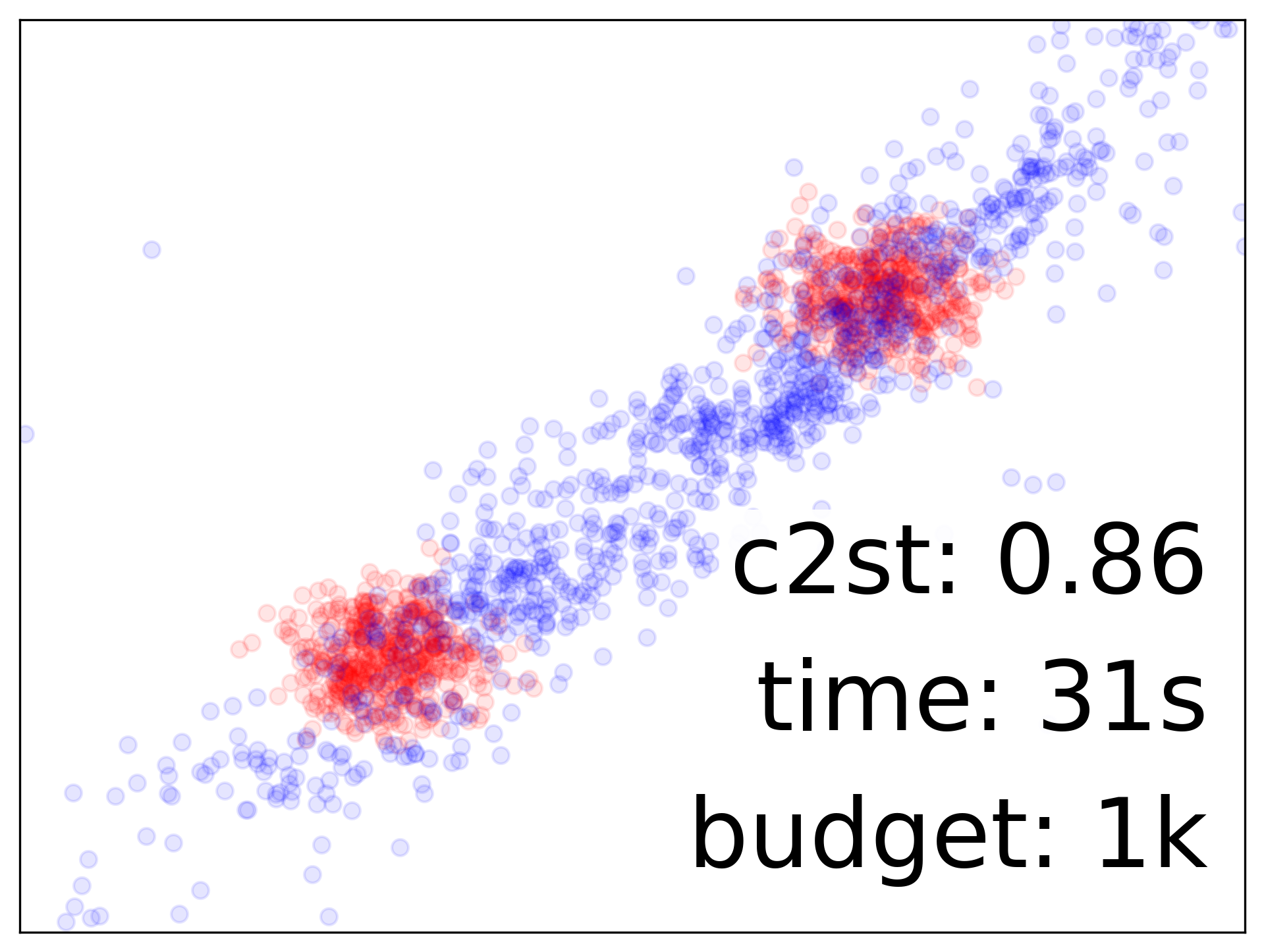} &
    \includegraphics[width=.12\textwidth]{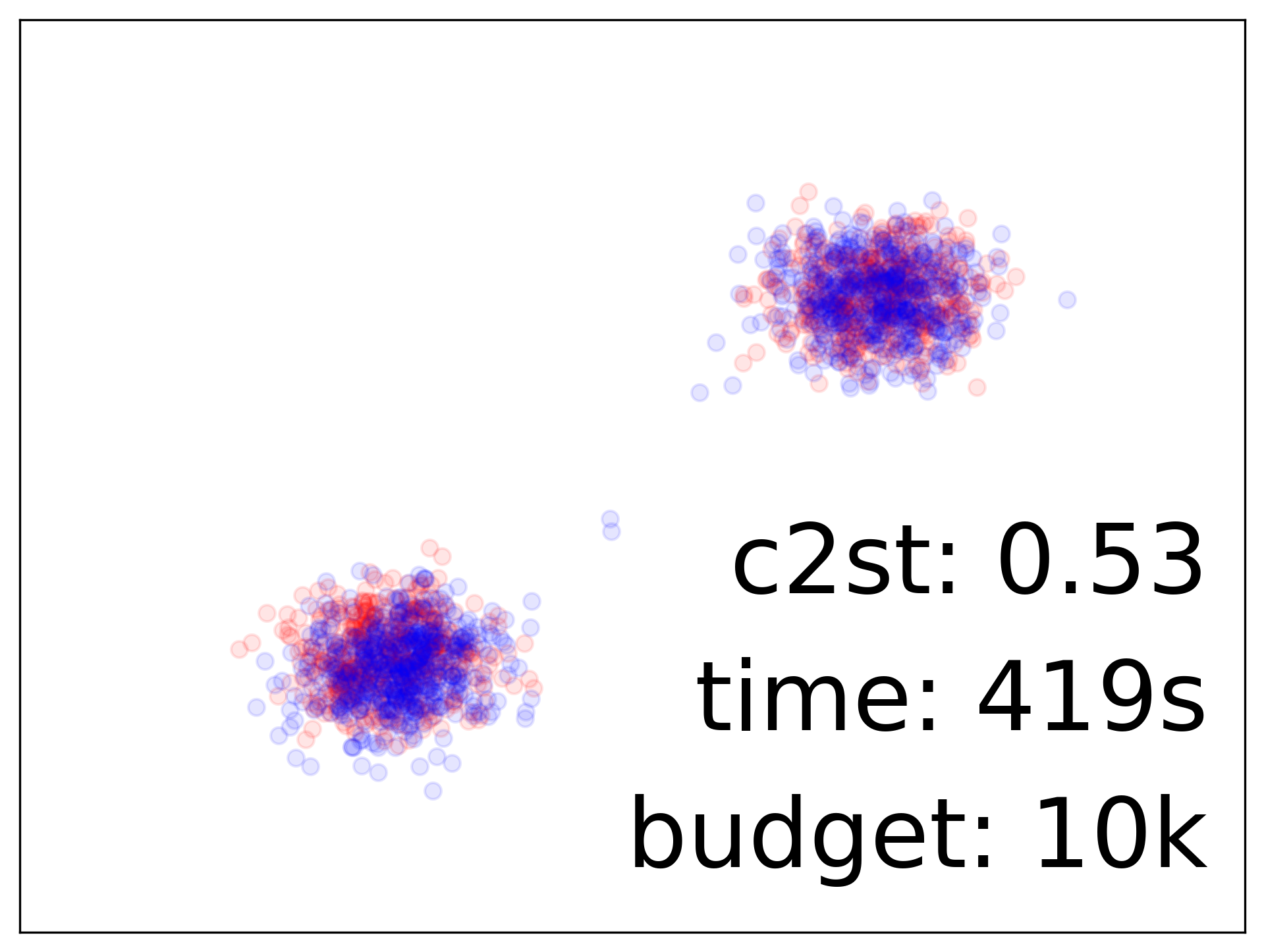} &
    \includegraphics[width=.12\textwidth]{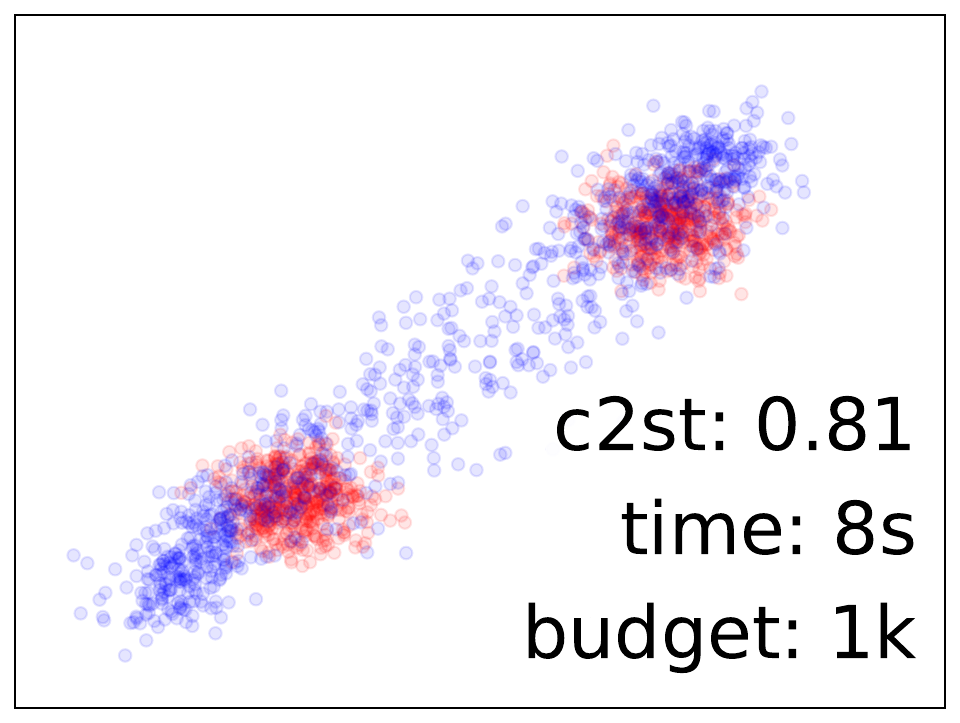} &
    \includegraphics[width=.12\textwidth]{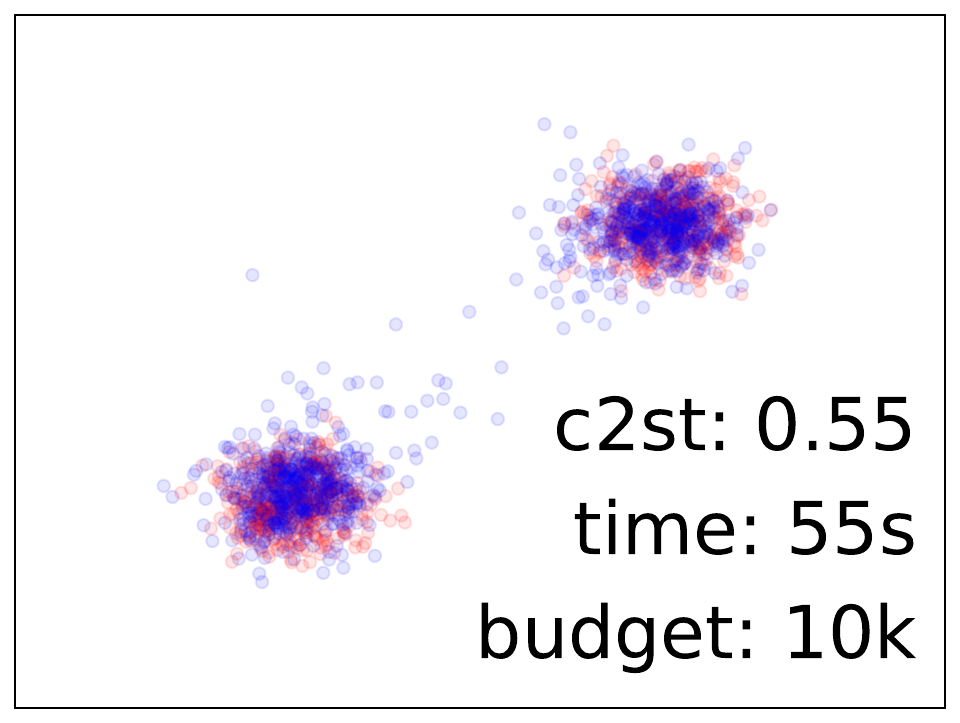} \\
    % Second row example
    \rotatebox{90}{\scriptsize{\; High-dim}} &
    \includegraphics[width=.12\textwidth]{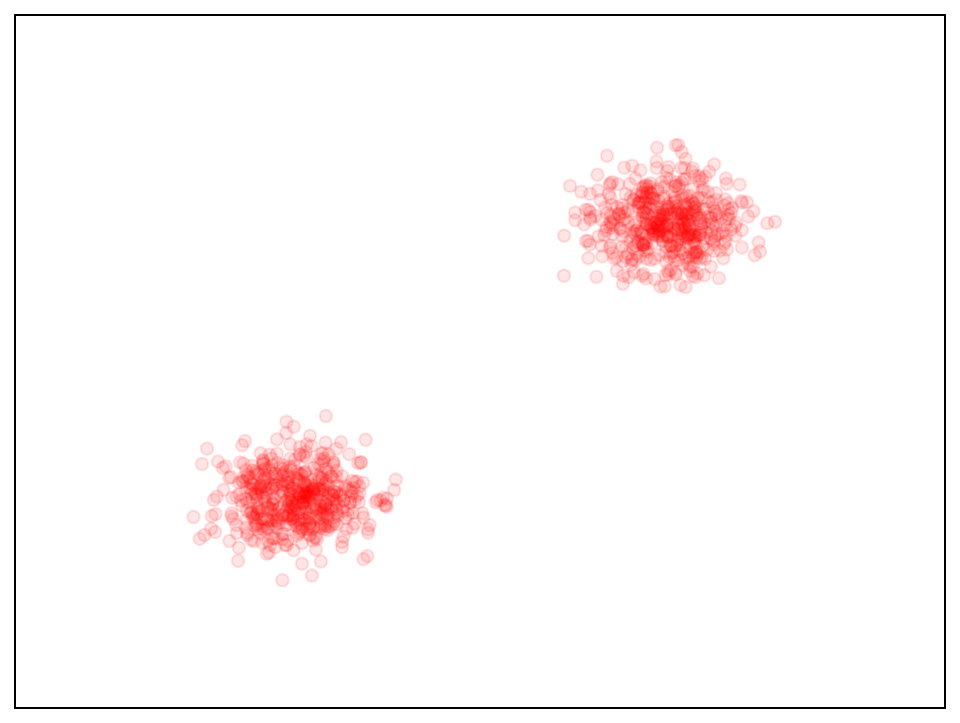} &
    \includegraphics[width=.12\textwidth]{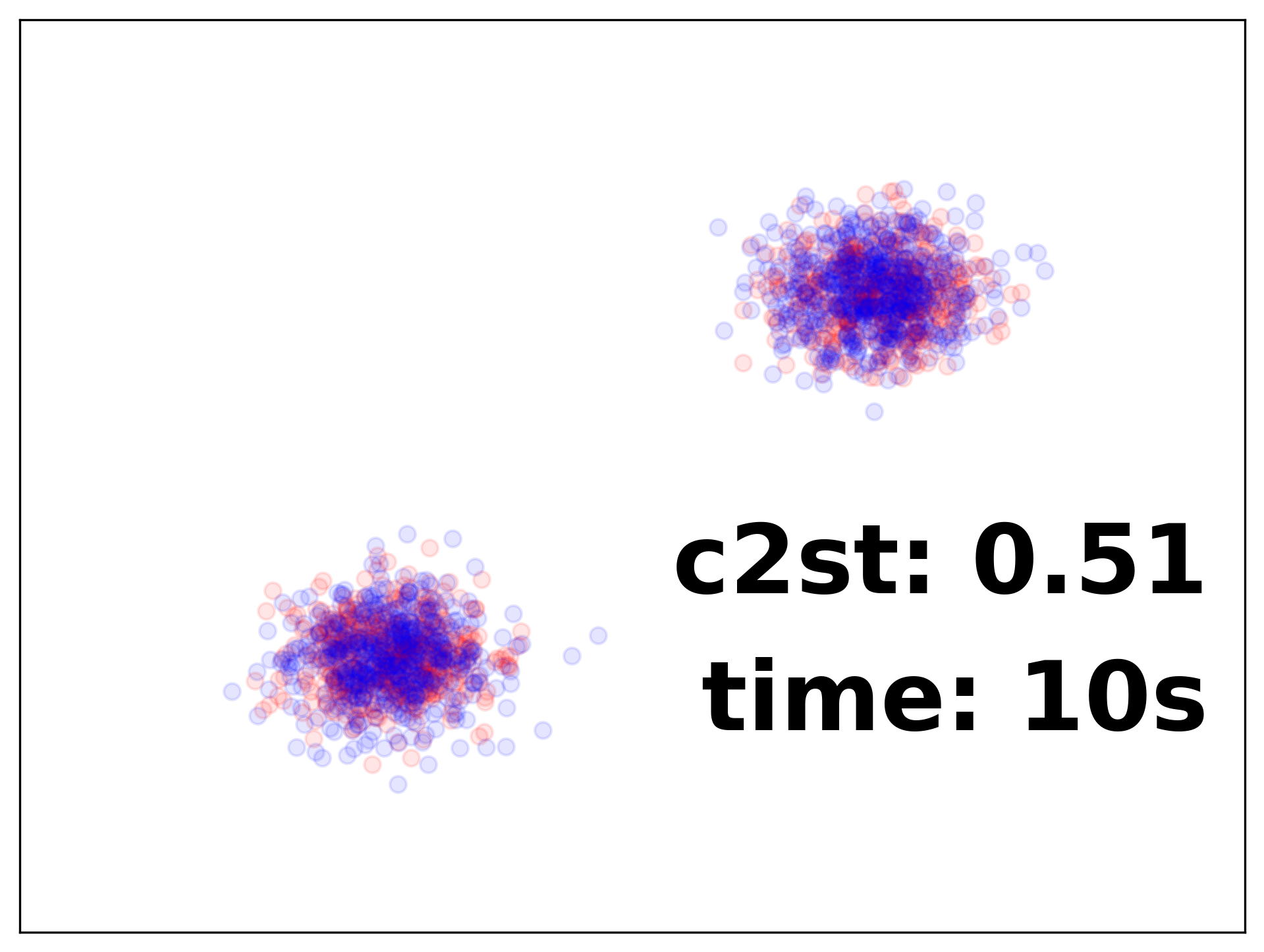} &
    \includegraphics[width=.12\textwidth]{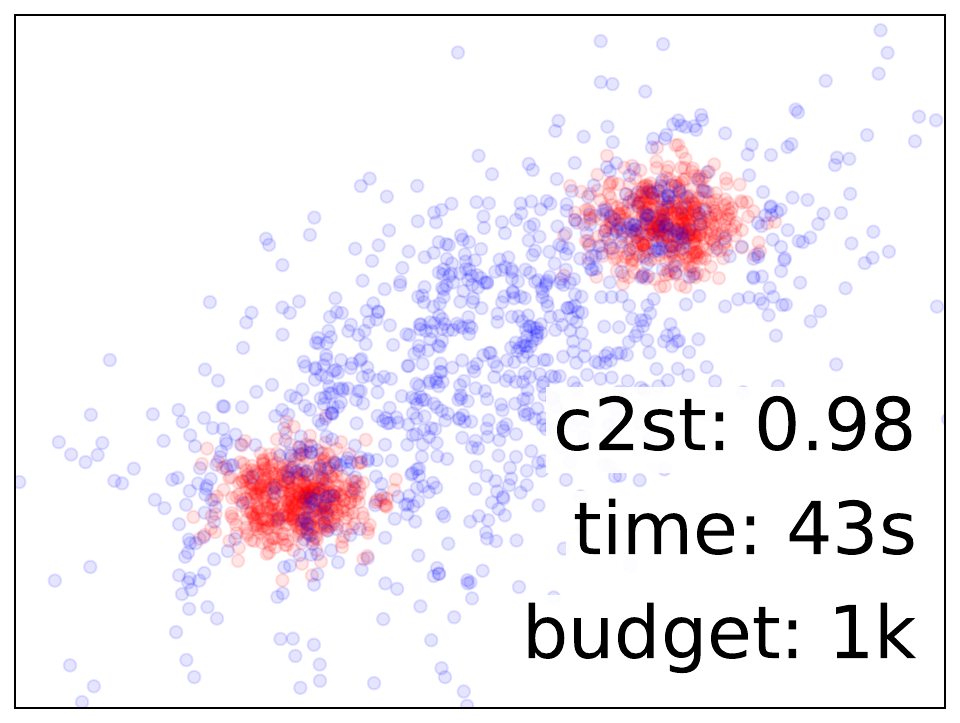} &
    \includegraphics[width=.12\textwidth]{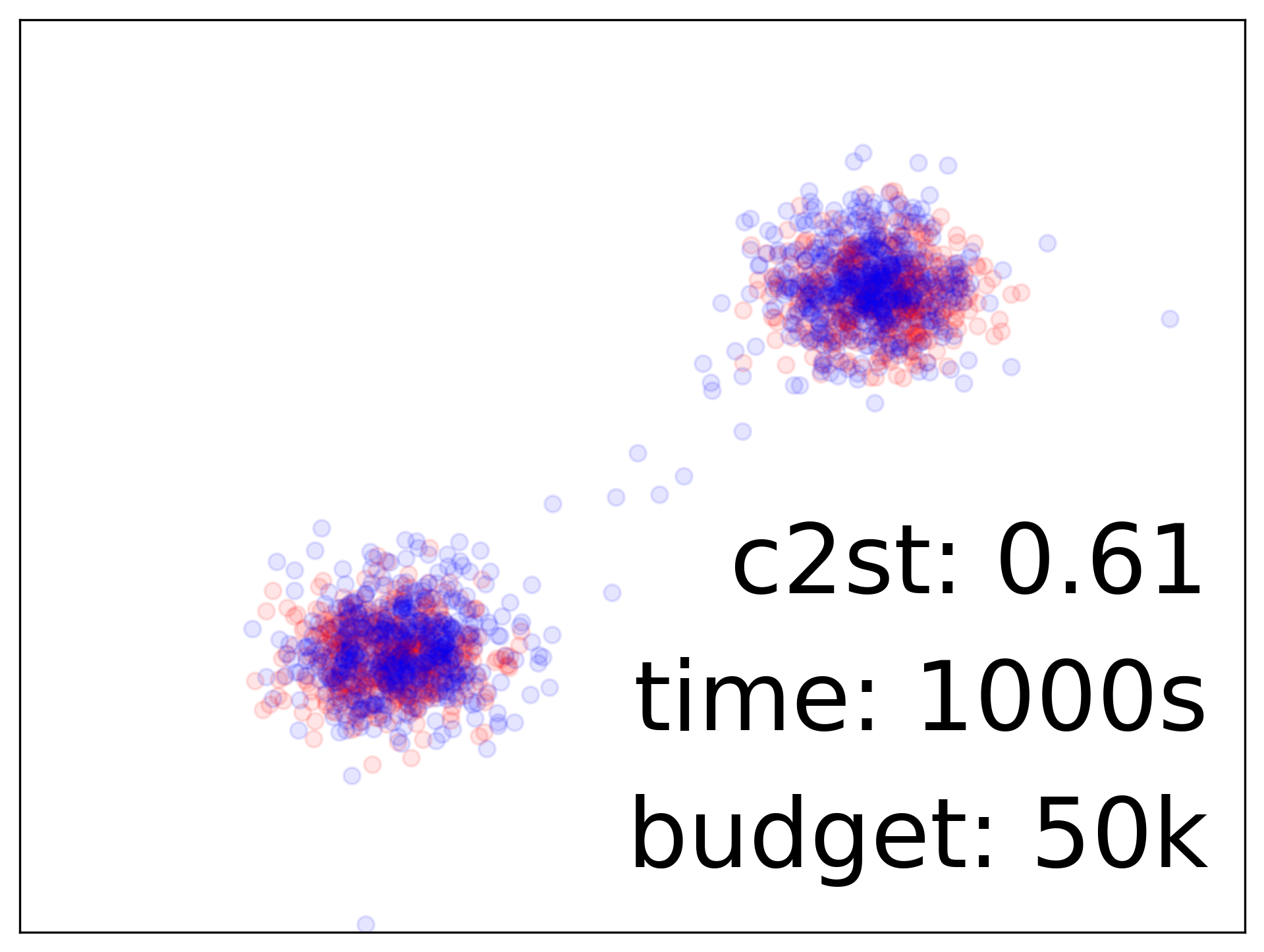} &
    \includegraphics[width=.12\textwidth]{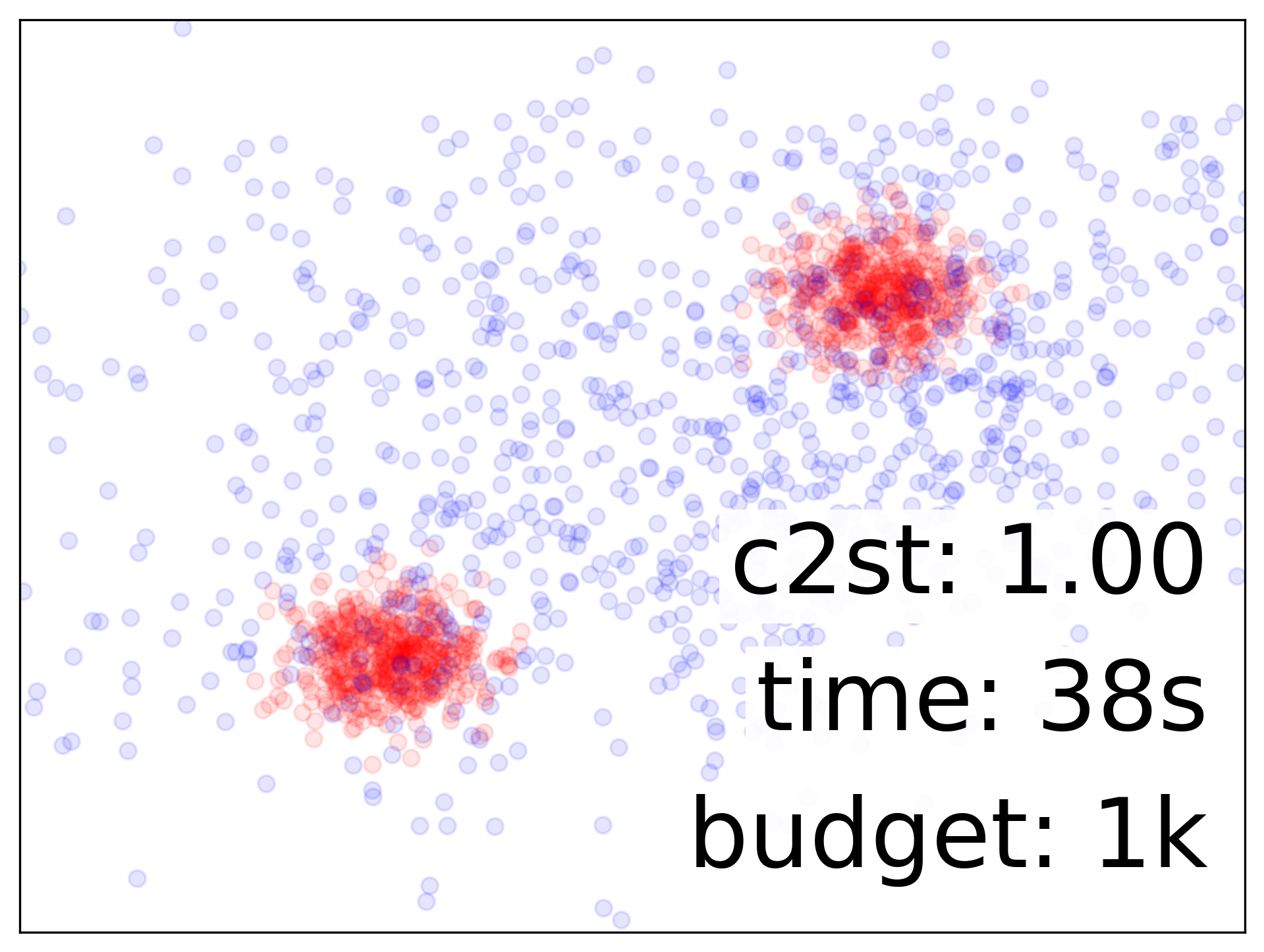} &
    \includegraphics[width=.12\textwidth]{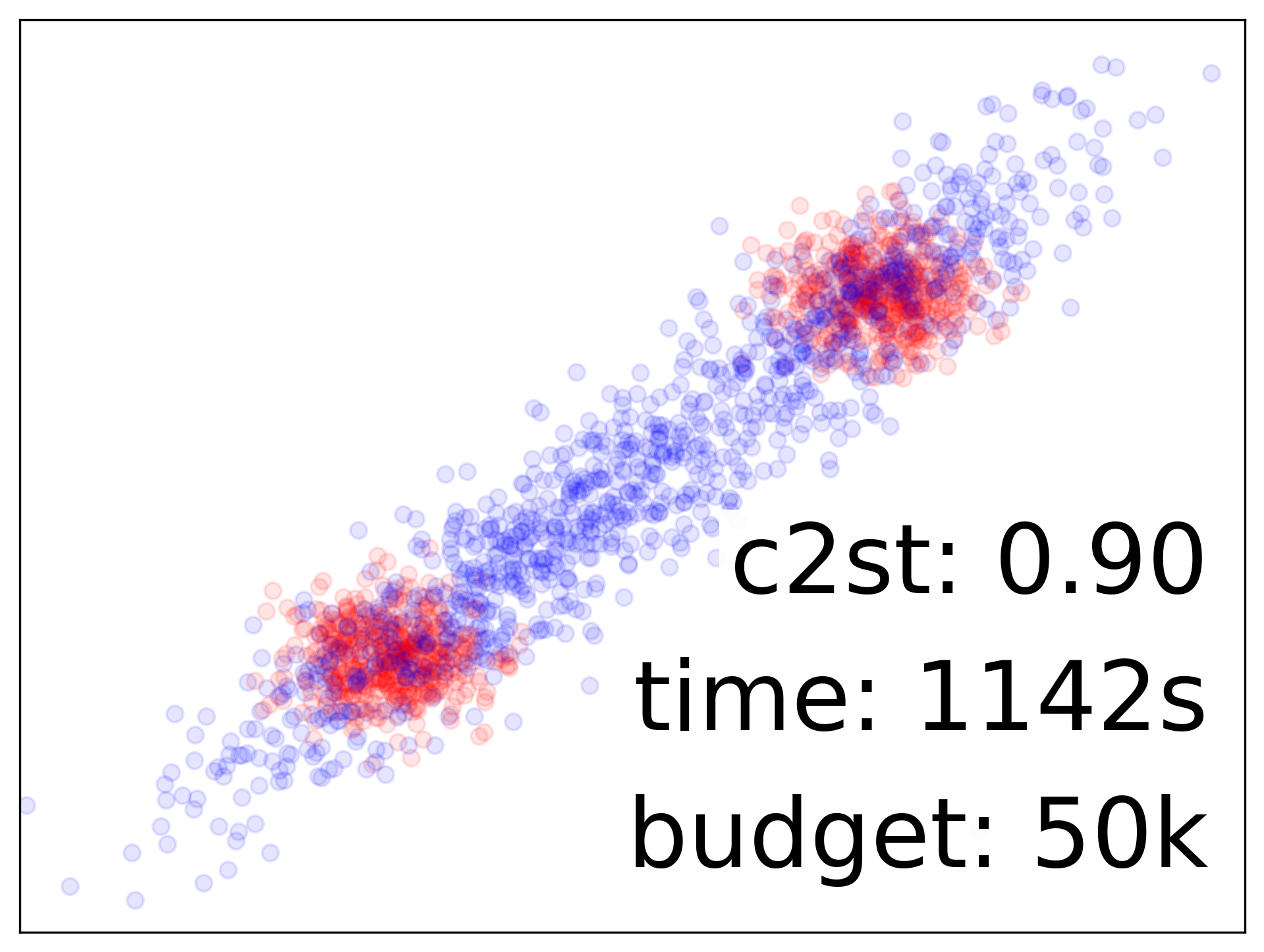} &
    \includegraphics[width=.12\textwidth]{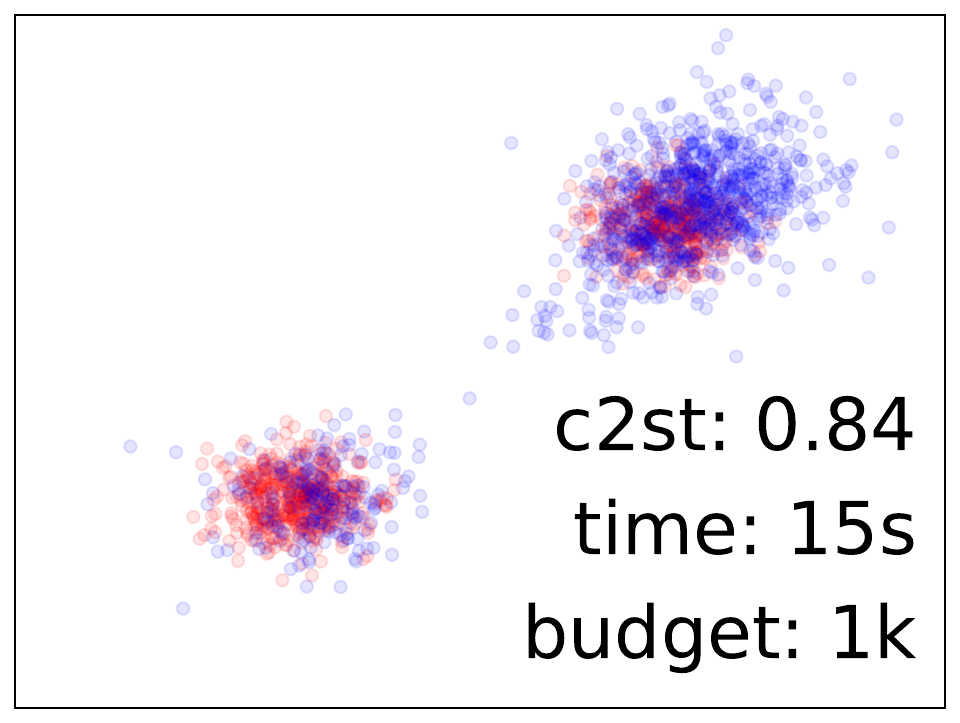} &
    \includegraphics[width=.12\textwidth]{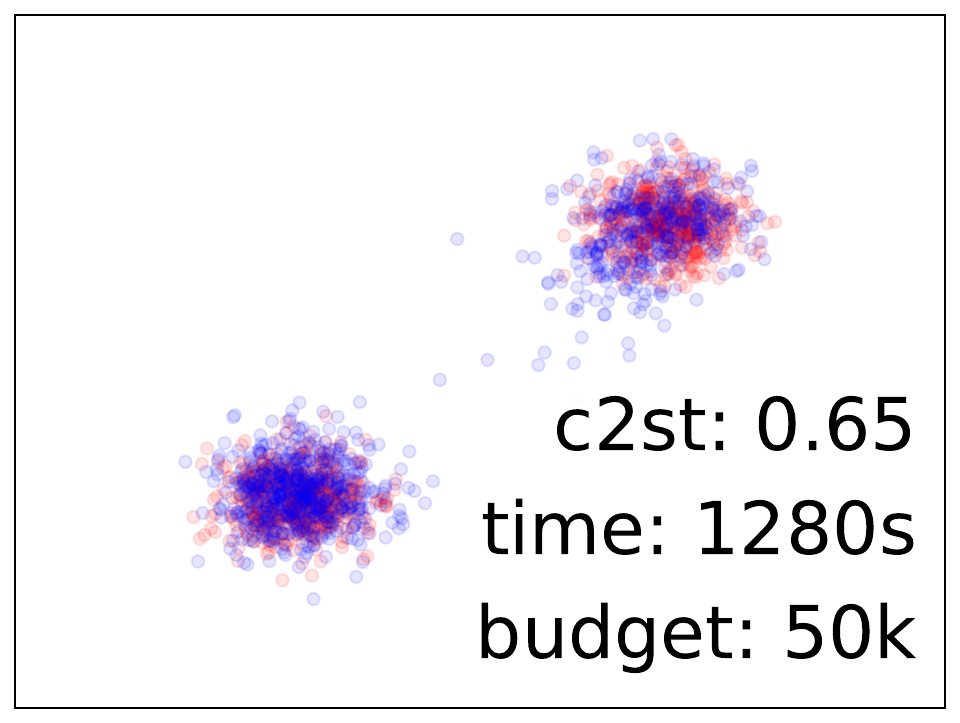} \\
  \end{tabular}
  \caption{
  Posterior inference on a two-Gaussian mixture simulator. 
  Rows correspond to simple ($D=D_y=2$), distractor ($D=2, D_y=18$), and high-dimensional ($D=D_y=10$) settings.
  Columns show the reference posterior, and samples obtained from \rromc~(proposed), and three established neural-based methods—Neural Posterior Estimator~\citep[NPE,][]{greenberg2019automatic}, BayesFlow~\citep{radev2020bayesflow}, and Flow Matching Posterior Estimator~\citep[FMPE,][]{wildberger2023flow}—each evaluated at low and high runtime. 
  Unlike the neural methods, which require high runtime in the distractor and high-dimensional settings, \rromc~produces accurate posterior samples with low runtime. See Appendix~\ref{app-sec:intro-example} for details on the experimental setup.
  }
  \label{fig:concept_example}
\end{figure*}

% Especially in two particular challenges: high-dimensional parameters and distractors
The computational cost of SBI methods is particularly pronounced in two scenarios.
First, in high-dimensional parameter spaces, they face the curse of dimensionality, as they rely on densely populating both parameter and data spaces with samples \citep{lintusaari2017fundamentals, sisson2018handbook}.
Second, when a subset of the output dimensions depends weakly, or not at all, on the simulator parameters \citep{Saltelli2008, MonsalveBravo2022}, these uninformative outputs act as ``distractors'', complicating inference \citep{lueckmann2021benchmarking}.
As illustrated in Figure~\ref{fig:concept_example}, in both cases existing neural-based approaches demand large simulation budgets, resulting in substantial runtime for an accurate posterior estimate.

% R2OMC: our proposal
We propose \rromc,~a novel LFI method that overcomes these challenges (see Figure~\ref{fig:concept_example}).
Building on the Robust Optimization Monte Carlo framework \citep[ROMC,][]{ikonomov2020robust},
 it reformulates the inference problem in terms of deterministic optimization problems and uses gradient descent to obtain posterior samples.
Gradients, as in standard optimization, provide efficient navigation in high-dimensional parameter spaces, so they rapidly guide us toward high-density posterior regions.
Moreover, they enable sensitivity analyses which allows the identification and filtering of distractor dimensions, improving inference quality.

% ROMC: the basis of our idea
ROMC, the foundation of our method, establishes the theory for casting inference as deterministic optimization but has key limitations: %  that constrain its applicability to complex inference tasks:
it lacks a strategy for defining an appropriate distance function in problems with distractors or multiple iid observations
and
its latest implementation~\citep{gkolemis2023extendable} does not exploit gradients or other computational accelerations, restricting its applicability to low-dimensional problems.
% Summarize R2OMC
\rromc~overcomes these limitations, providing an SBI method for differentiable simulators that:

\begin{itemize}
 \item enables posterior sampling in high-dimensional parameter spaces with efficient runtimes,
 \item automatically identifies and filters out uninformative dimensions (distractors),
 \item handles multiple iid observations effectively,
 \item runs efficiently through a \texttt{JAX} implementation that uses automatic differentiation, vectorization, and just-in-time compilation.
 \end{itemize}

 \rromc~is systematically compared to state-of-the-art SBI methods, demonstrating accurate posterior inference with substantially lower runtime across a wide range of scenarios.

 Code for using R2OMC is publicly available at \href{https://github.com/givasile/lfi}{github.com/givasile/lfi}.
 To reproduce the experiments exactly, use the branch \href{https://github.com/givasile/lfi/tree/aistats2026}{github.com/givasile/lfi @ aistats2026}.

\section{Background and Motivation}

% Neural-based methods are computationally intensive
Neural-based SBI methods are data-intensive, with their accuracy depending on access to large simulated datasets.
This incurs computational costs on two fronts:
generating massive datasets from complex simulators and
training deep neural estimators on such volumes.
Prior work~\citep{lueckmann2021benchmarking} has shown that it is the second cost, neural network training, that often dominates SBI.
Moreover, modern frameworks, like \texttt{JAX}, are used in SBI~\citep{dirmeier2024simulation, quera2023blackbirds} to accelerate simulation through vectorization, which can massively reduce the first cost but not the second.
Appendix~\ref{app-sec:compute-budget} expands on both points. 
This motivates us to develop a method that avoids the cost of training an external neural estimator.
 
\subsection{ROMC framework}
\label{subsec:ROMC}
ROMC, introduced by \citet{ikonomov2020robust}, improves upon Optimization Monte Carlo of~\citet{meeds2015optimization} by addressing its robustness issues. 
It belongs to the larger family of ABC methods that approximate the likelihood function as the probability of simulating an output $\epsilon$-close to the observation,
\begin{equation}
L_\epsilon(\thb) = \Pr(d(\yb, \data) \leq \epsilon \mid \thb).
\label{eq:abc_likelihood}
\end{equation}
Here, $d(\cdot, \cdot)$ is a distance, $\yb$ the simulator output, and $\data$ the observed data. In the following, we denote by $g(\thb, \ub)$ the simulator that maps noise (nuisance) variables $\ub \sim p(\ub)$ and parameters $\thb$ to multivariate data $\yb = g(\thb, \ub)$. 
In a computer program, sampling $\ub$ corresponds to sampling a random seed.
Introducing the acceptance region $C_\epsilon = \{(\thb, \ub): d(g(\thb, \ub), \data)\le \epsilon\}$ and using the law of the unconscious statistician, (\ref{eq:abc_likelihood}) can be written as
\begin{equation}
    L_\epsilon(\thb) = \E_{p(\ub)}\left[\mathbb{1}_{C_\epsilon}(\thb, \ub) \right],
\end{equation}
where $\mathbb{1}_{C_\epsilon}$ denotes the indicator function that is one if $(\thb, \ub)\in C_\epsilon$ and zero otherwise. When approximating the expectation as a sample average, we obtain
\begin{equation}
    L_\epsilon(\thb) \approx \frac{1}{S} \sum_{i=1}^S \mathbb{1}_{\accregioni}(\thb),
    \label{eq:omc_likelihood}
\end{equation}
which is the proportion of acceptance regions $\accregioni = \{\thb : d(g(\thb, \ub_i), \data) \leq \epsilon \}$ that contain $\thb$.

ROMC takes (\ref{eq:omc_likelihood}) as starting point and focuses on characterizing the acceptance regions $\accregioni$, because once the $\accregioni$ are known, we can not only evaluate the approximate likelihood function, but also obtain posterior samples efficiently. ROMC introduces uniform proposal distributions $q_i$ with support on $\accregioni$ only. Samples from $\thb_{ij} \sim q_i$ for $j \in 1,\ldots, M$, are then equal to posterior samples if weighted with the (unnormalized) weights $w_{ij}$, $i=1, \ldots, S, j=1, \ldots, M$,
\begin{equation}
w_{ij} = \frac{p(\thb_{ij})}{q_i(\thb_{ij})} \mathbb{1}_{C^i_{\epsilon}}(\thb_{ij}) \label{eq:romc_weight}
\end{equation}

where $p(\thb)$ is the prior. For the characterization of the acceptance regions $\accregioni$, ROMC first determines the minima $\thb_i^*$,
\begin{equation}
 \thb_i^* = \text{argmin}_{\thb} d_i(\thb),\quad d_i(\thb) = d(g(\thb, \ub_i), \data),       
 \label{eq:optim_end_points}
\end{equation}

where $d_i(\thb)$ are deterministic distance functions because the sources of randomness (seeds) are set to $\ub_i$ and kept fixed.
Next, it starts from the optimization end points $\thb_i^*$ to discover $\accregioni$ selecting $\epsilon$ as the value derived from the quantiles of the minimum distance functions, $d_i^*=d_i(\thb_i^*), i=1, \ldots, S$ \citep[see][for details]{ikonomov2020robust}. While several options are possible \citep{ikonomov2020robust}, here we use hyperboxes centered at $\thb_i^*$, which can be constructed with a line search algorithm for each parameter dimension. For details, please see Appendix \ref{app-sec:methods}.

% Limitations of ROMC - TODO: maybe it can be shortened a bit
The power of the ROMC framework stems from the possibility to use optimization, in particular gradient-based methods, to determine the optimization end points $\thb_i^*$, and hence the acceptance regions $\accregioni$ and the proposal distributions $q_i$. 
However, ROMC has important limitations, which we analyze in the next subsection.

\subsection{ROMC challenges and limitations}
\label{sec:romc-limitations}

ROMC suffers from three main limitations;
(a) it fails in the presence of distractors,
(b) it does not provide a strategy for handling multiple iid observations, and
(c) its existing implementation does not use gradients and other computational accelerations, so it can only be applied to low-dimensional problems.

\textbf{Uninformative dimensions (distractors).}
\label{sec:romc-limitation-1}
Distractors are output dimensions that do not depend on the parameters of interest. For instance, consider a graphics renderer where the parameter of interest controls a scene element, e.g.\ the appearance of vase, so that most pixel values (output dimensions) are unrelated to the parameter and thus serve as distractors.
Formally, consider an observation vector $\yb^o = (\yb^{o, (1)}, \yb^{o, (2)})$, 
where only $\yb^{o, (1)}$ depends on $\thb$.
Posterior inference depends solely on $\yb^{o, (1)}$, so the elements of $\yb^{o, (2)}$ act as distractor dimensions.

% Why ROMC fails with distractors
ROMC is vulnerable to failure in this setting because typical distance functions $d$ consider all output dimensions equally.
Therefore, the minimal distances $d_i^*$ will generally be large for seeds that generate $\yb^{(2)}$ significantly different from $\yb^{o, (2)}$.
This results in an inflated $\epsilon$, and in turn to a much broader posterior than the ground truth.

\begin{figure*}
  \centering
  \includegraphics[width=.95\linewidth]{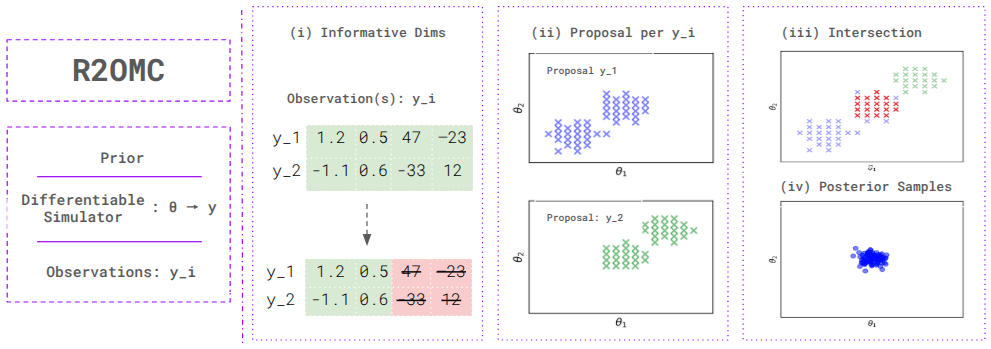}
  \caption{
    \rromc~overview: Distractors are automatically filtered out and proposal distributions $q_i^n$ are created for each observation (here two): the blue-dotted region indicates where the proposal distributions for the first observation are nonzero, and the green-dotted region indicates the same for the second observation. Only samples that fall within the intersection of both regions are given a positive weight and become posterior samples.}
  \label{fig:concept_image}
\end{figure*}

\textbf{Multiple iid observations.}
In many applications, we have access to multiple iid observations $\{\data_n\}_{n=1}^N$.
For instance, in neuroscience, researchers often collect several recordings of the same neural response to a stimulus.
Properly incorporating these repeated measurements is crucial for accurate inference, however, most simulation-based methods, including ROMC, are designed for single observations, making it non-trivial to aggregate data and estimate $p(\thb; \{\data_n\}_{n=1}^N)$ correctly.

% Naive approaches will fail
Naive approaches are prone to failure. 
For example, simulating $N$ outputs with fixed seeds $\{\ub_{i,n}\}$ and averaging pairwise distances, i.e., $d_i(\thb) = 1/N \sum_n \tilde{d}(g(\thb, \ub_{i,n}), \data_n)$—or concatenating all observations into a single vector—are sensitive to the arbitrary ordering of data points, violating permutation invariance.
Such artificial sensitivity inflate the minimal achievable distances $d^{*}_i$, forcing an excessively large $\epsilon$ and yielding posteriors that are much broader than the ground truth. 
Similarly, simple summary statistics, like computing the mean of each feature across observations, discard higher-order information and variability, producing biased or overconfident posteriors.

\textbf{Computational efficiency.}
Current ROMC implementations \citep{gkolemis2023extendable} suffer from significant computational bottlenecks that restrict their applicability to low-dimensional problems.
First, optimization relies on gradient-free methods, e.g., Bayesian optimization, which scale poorly with the number of parameters. % and require many simulator evaluations.
Second, the construction of the acceptance regions $\accregioni$ with a line-search algorithm (see Appendix \ref{app-sec:methods}), requires repeated simulator evaluations per seed and per dimension, which also scales poorly with the parameter dimensions.

\section{Proposed method: \rromc}
\label{sec:rromc}

We here present our new method for Bayesian parameter inference in simulator models. The method uses gradients to obtain posterior samples even in high-dimensional parameter spaces. The method belongs to the ROMC framework so that we start with discussing how we deal with distractors and iid observations before discussing its implementation.

\subsection{Uninformative dimensions (distractors)}
\label{subsec:distractors}
Distractors can negatively affect the quality of the posterior approximation. To counteract their influence, we measure the sensitivity of the output dimensions with respect to the parameters $\thb$ by computing the average norm of their derivative with respect to $\thb$, and checking that the value is above a minimal value (threshold) $\tau$, 
\begin{equation}
  \label{eq:informative_dimensions}
  I_i = \E_{p(\thb)p(\ub)}\left[ || \nabla_{\thb} g_i(\thb, \ub) || \right] > \tau.
\end{equation}
The expectation equals the expectation with respect to the joint distribution of the prior and the $i$-th dimension of the simulator output, $g_i(\thb, \ub)$. In our experiments, we approximate the expectation using $50$ samples from $p(\thb)$ and $p(\ub)$ each. Collecting all binary indicator variables $I_i$ into the vector $\mask$, we can filter out uninformative dimensions by computing the distance between observed and simulated data for informative dimensions only. To compute $\thb_{i,n}^*$ in (\ref{eq:rromc_optim_end_points}), we minimize
\begin{equation}
d_i^n(\thb) = d(g(\thb, \ub_{i,n})\odot \mask, \data_n \odot \mask).
\label{eq:rromc-masked-distance}
\end{equation}
The procedure is illustrated in Figure~\ref{fig:concept_image}, step (i). 

The threshold $\tau$ controls the sensitivity of the test. In our experiments, we set $\tau$ to machine precision; more sophisticated approaches consist in monitoring the distribution of the expected norm of the gradient, or by assessing false vs true positives under control conditions. This is possible since the mask $\mask$ can be computed prior to receiving observed data.

\subsection{Multiple iid observations}
\label{subsec:multiple-observations}

The likelihood for multiple iid observations $\{\data_n\}_{n=1}^N$ is the product of the likelihoods for each data point. Approximating the likelihood for each data point as in (\ref{eq:omc_likelihood}) and multiplying-in the prior $p(\thb)$, we can obtain a formal expression for the approximate posterior
\begin{equation}
  \label{eq:rromc_posterior}
  p_{\epsilon}(\thb | \{\data_n\}) \propto p(\thb) \prod_{n}^{N} \sum_{i}^S \mathbb{1}_{C_{\epsilon}^{i,n}}(\thb),
\end{equation}
where $C_{\epsilon}^{i,n}$ is the acceptance region for data point $n$ and source of randomness (seed) $\ub_{i,n}$,
\begin{equation}
C_{\epsilon}^{i,n} = \{\thb : d(g(\thb, \ub_{i,n}), \data_n) \leq \epsilon \}.
\end{equation}
We have $S$ sources of randomness for each data point $\data_n$, and so in total $S*N$ acceptance regions $C_{\epsilon}^{i,n}$.

We now address the question how to efficiently sample from $p_{\epsilon}(\thb | \{\data_n\})$. To achieve this, we consider the task of computing the posterior expectation $\bar{h} = \E_{p_{\epsilon}}[h(\thb)]$ under $p_{\epsilon}(\thb | \{\data_n\})$. Plugging in the expression in (\ref{eq:rromc_posterior}) gives, after normalization,
\begin{equation}
  \label{eq:rromc_expectation}
  \bar{h} = \frac{ \int h(\thb) p(\thb) \prod_n^N \sum_i^S \mathbb{1}_{C_{\epsilon}^{i,n}}(\thb) d\thb}{\int p(\thb) \prod_n^N \sum_i^S \mathbb{1}_{C_{\epsilon}^{i,n}}(\thb) d\thb}.
\end{equation}

The integrals in the equation are typically intractable. But they correspond to expectations with respect to the prior $p(\thb)$ and hence could be approximated with a sample average. However, this is very inefficient. Few (or no) prior samples will generate outputs $\epsilon$-close to all observations, so most (or all) will get zero weight, leading to a very low effective sample size. We thus employ importance sampling with an adaptive proposal distribution obtained via optimization. 

We first treat each data point $\data_n$ separately and construct proposal distributions $q_i^n(\thb)$, $i=1, \ldots, S$, as in Section \ref{subsec:ROMC}: We use optimization to find the minimizers $\thb_{i,n}^*$ of the deterministic functions $d_i^n(\thb) = d(g(\thb, \ub_{i,n}), \data_n)$,       
\begin{equation}
 \thb_{i,n}^* = \text{argmin}_{\thb} d_i^n(\thb),
 \label{eq:rromc_optim_end_points}
\end{equation}
and then build a hyperbox around each acceptance region $C_{\epsilon}^{i,n}$ and define $q_i^n(\thb)$ to be the uniform distribution over it. The optimization problem in (\ref{eq:rromc_optim_end_points}) is deterministic and we solve it with gradient-based methods to enable scalability to large parameter spaces. Note that treating each data point $\data_n$ separately allows us to avoid the issues explained in Section \ref{sec:romc-limitations}.

We propose to combine the per-data point proposal distributions $q_i^n(\thb)$ in form of a mixture distribution,
\begin{equation}
    q(\thb) = \frac{1}{N} \frac{1}{S} \sum_{n=1}^N \sum_{i=1}^S q_i^n (\thb).
    \label{eq:R2OMC-proposal}
\end{equation}
The posterior expectation in (\ref{eq:rromc_expectation}) can then be approximated in form of a weighted average 
\begin{equation}
  \label{eq:rromc_sample_expectation}
  \bar{h} = \frac{\sum_p w_p h(\thb_p)}{\sum_p w_p}, \quad \thb_p \sim q(\thb)
\end{equation}
where the weights $w_p$, $p=1, \ldots, P$, are given by
\begin{equation}
  \label{eq:rromc_weight}
w_p= \frac{p(\thb_p)}{q(\thb_p)} \prod_{n=1}^N \sum_{i=1}^S \mathbb{1}_{C_{\epsilon}^{i,n}}(\thb_p). 
\end{equation}
This means that the $\thb_p \sim q(\thb)$, weighted by $w_p$, represent samples from the posterior $p_{\epsilon}(\thb | \{\data_n\})$ in (\ref{eq:rromc_posterior}).

% TODO add something about what happens when little overlap exists
The rationale for choosing $q(\thb)$ in (\ref{eq:R2OMC-proposal}) as proposal distribution is as follows: it is a mixture of uniform distributions, so sampling and evaluating is easy. It seamlessly combines the contribution from multiple data points, allowing future data points to be included without re-computation. If each $q_i^n$ encloses the corresponding acceptance region $C_{\epsilon}^{i,n}$, then $q(\thb) > 0$ in areas of the parameter space where $\prod_n^N \sum_i^S \mathbb{1}_{C_{\epsilon}^{i,n}}(\thb) > 0$ and no posterior mass is discarded. Finally, regions of high likelihood are characterized by regions where multiple $C_{\epsilon}^{i,n}$ overlap. As the proposal distribution $q(\thb)$ consists of a mixture of the $q_i^n(\thb)$ with equal weights, areas where multiple $C_{\epsilon}^{i,n}$ overlap will be sampled more often, so that areas of high likelihood will be represented with more samples.

Figure~\ref{fig:concept_image}, steps (ii–iii), illustrates the proposed procedure to obtain posterior samples $\thb_p$. We have two observed data points and the $\thb_p$ come from regions that match at least one observation; the blue-dotted area for $\data_1$ and the green-dotted region for $\data_2$. However, only the ones that match both will get a positive weight (red dots) and become posterior samples.

\begin{figure*}[!t]
  \centering
  \begin{subfigure}[b]{0.245\textwidth}
    \includegraphics[width=\linewidth]{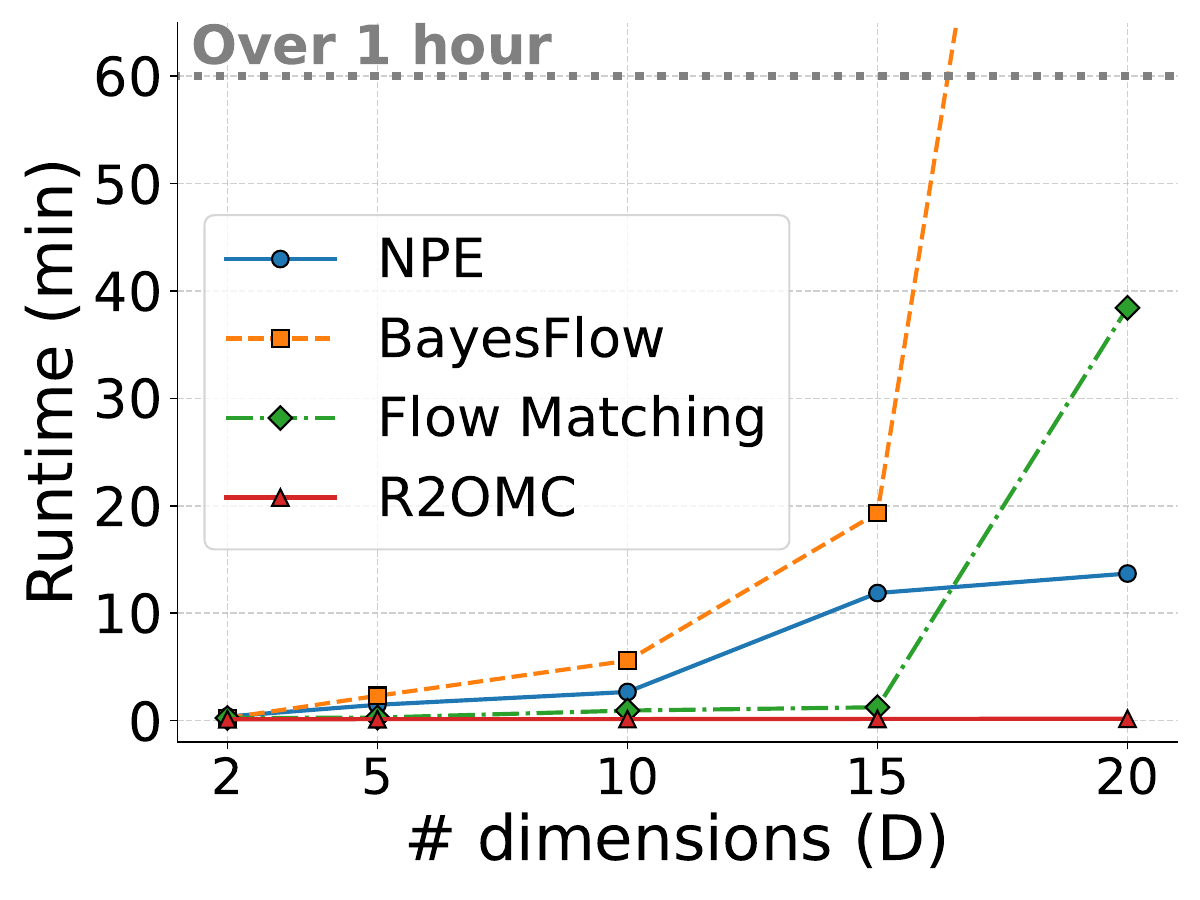}
    \caption{$S_{\mathtt{base}}$}
  \end{subfigure}
  \begin{subfigure}[b]{0.245\textwidth}
    \includegraphics[width=\linewidth]{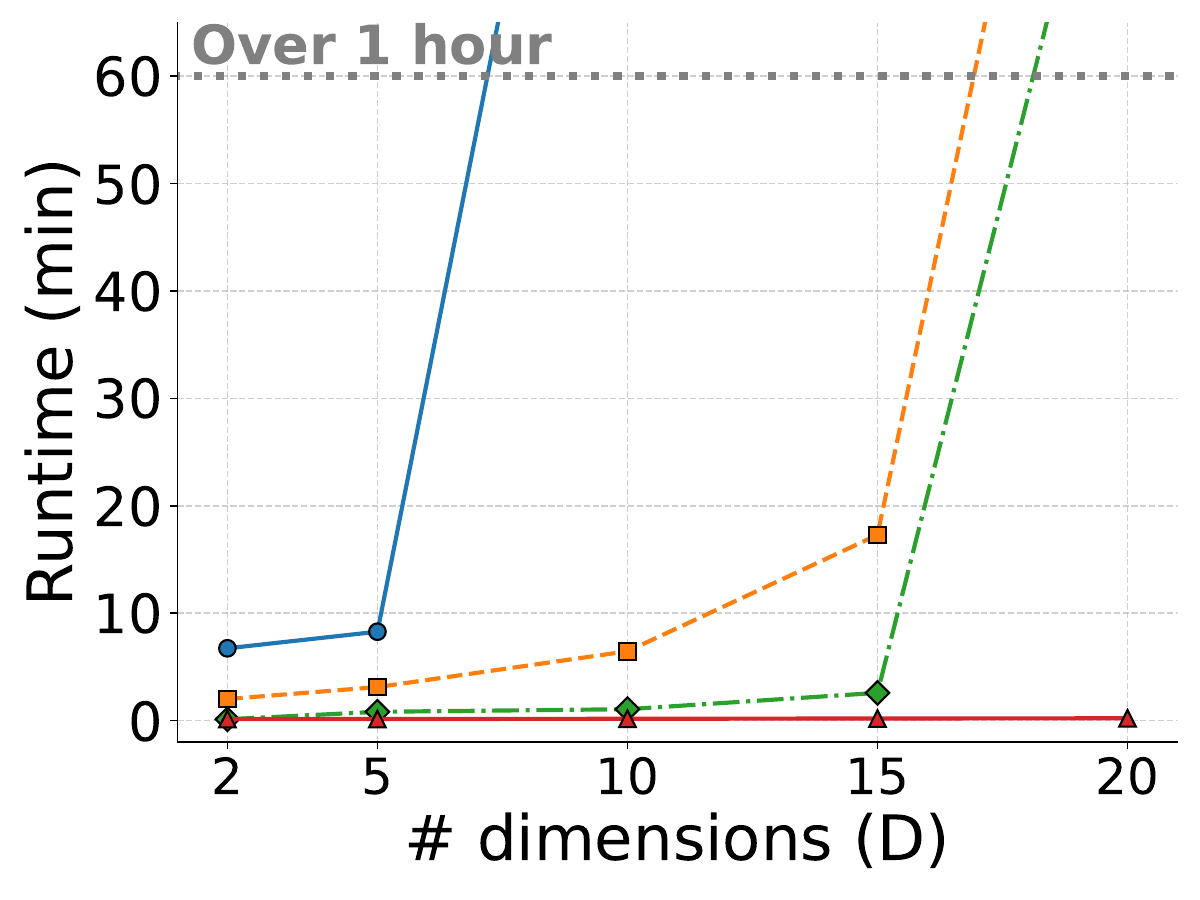}
    \caption{$S_{\mathtt{base}}^{\mathtt{dist}}$}
  \end{subfigure}
  \begin{subfigure}[b]{0.245\textwidth}
    \includegraphics[width=\linewidth]{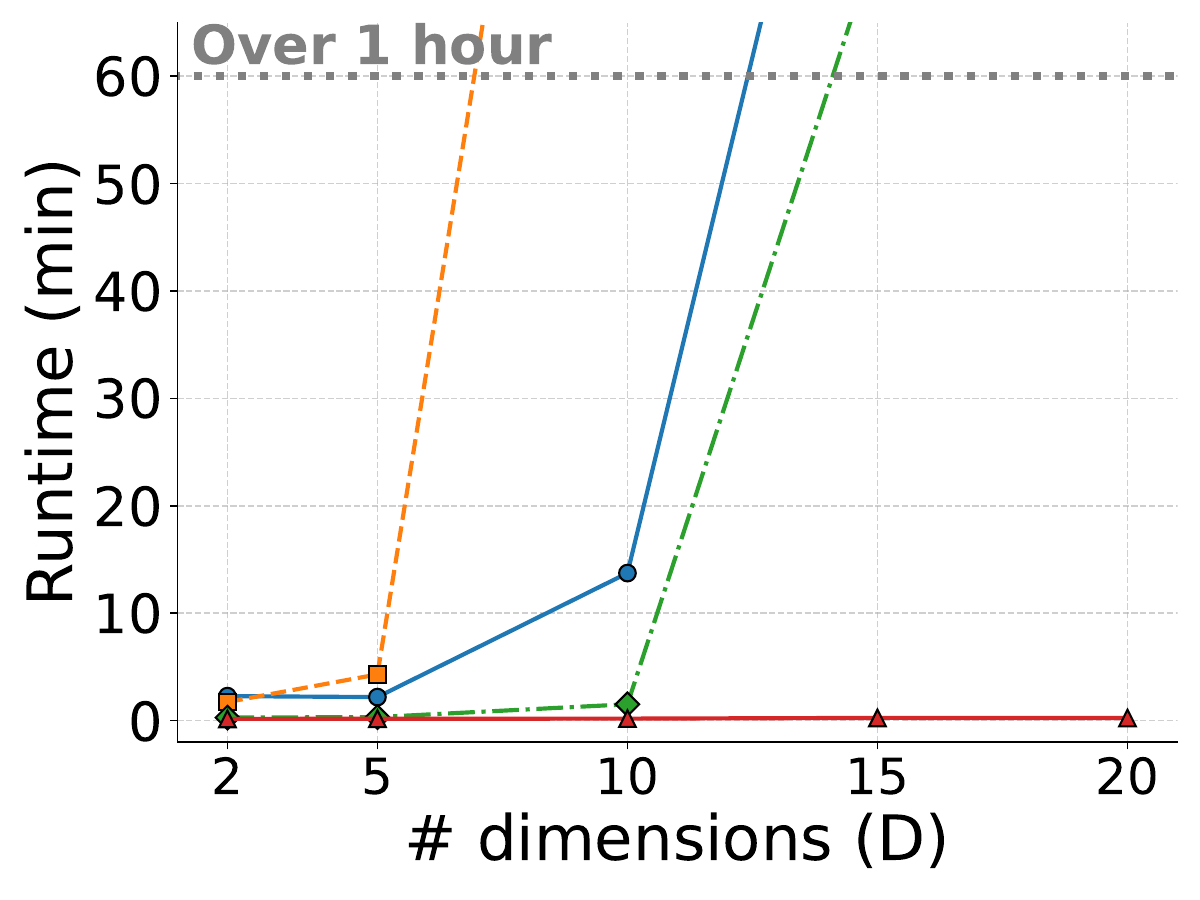}
    \caption{$S_{\mathtt{MoG}}$}
  \end{subfigure}
  \begin{subfigure}[b]{0.245\textwidth}
    \includegraphics[width=\linewidth]{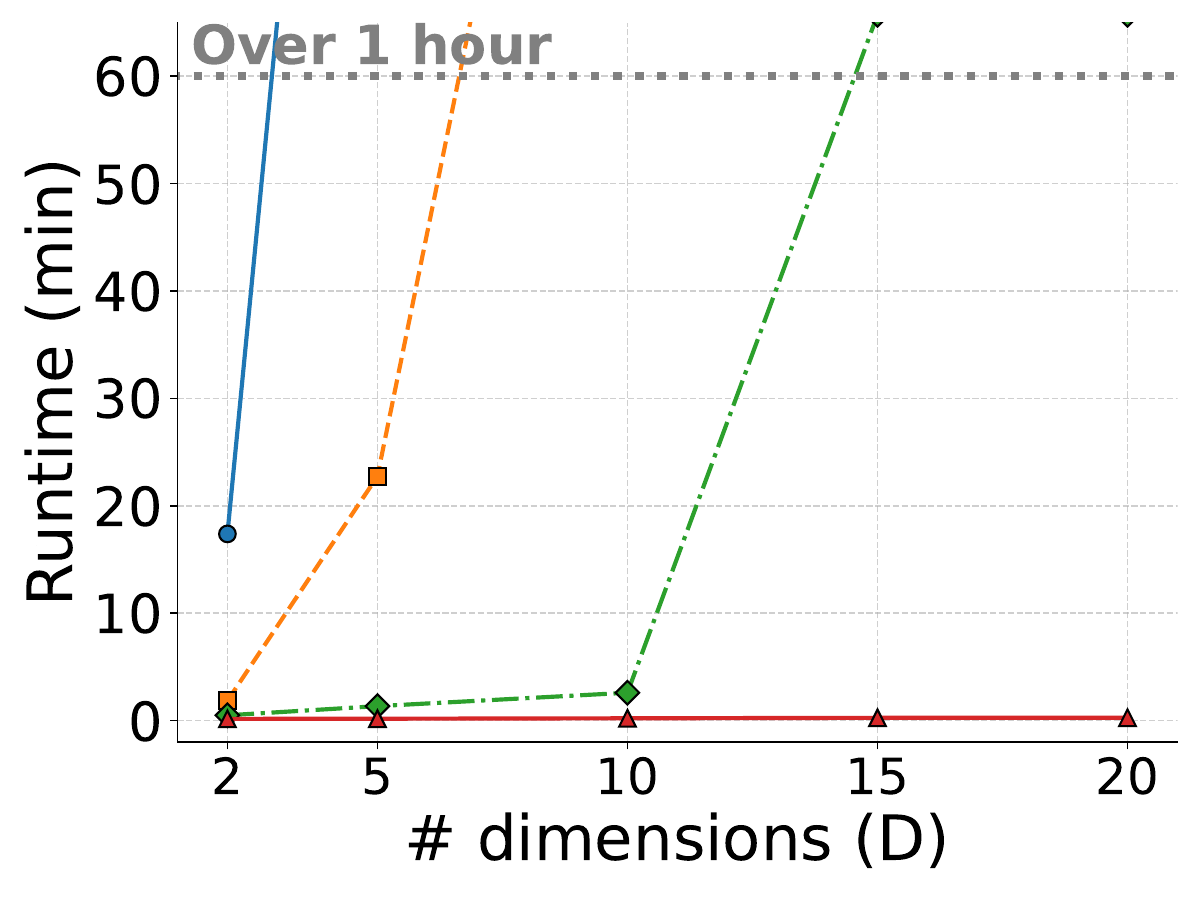}
    \caption{$S^{\mathtt{dist}}_{\mathtt{MoG}}$}
  \end{subfigure}
  \caption{MoG benchmark: Success Frontier plots showing the minimum runtime required to achieve a mean C2ST score $\leq 0.75$ across varying $D$. 
  Corresponding Budget vs. Dimension plots are provided in Appendix~\ref{app-sec:experiments}.}
  \label{fig:high_dim_benchmark}
\end{figure*}

\subsection{\texttt{JAX} implementation}
\label{subsec:implementation}

Algorithms~\ref{alg:r2omc} summarizes the key steps of \rromc: dealing with distractors (line 2); constructing per-data point proposal distributions $\{q_i^n\}_{i=1}^S$ (lines 3-6); combining them to form the overall proposal distribution $q(\thb)$ (line 7) and generating weighted samples (line 8).

\begin{algorithm}[!ht]
  \caption{\texttt{R2OMC}. Requires: $p(\thb), g(\thb, \ub), \{\data_n\}_{n=1}^{N}$}
  \label{alg:r2omc}
  \begin{algorithmic}[1]
    \Procedure{\rromc}{$\tau, S, P$}
      \State $\mask = (I_1, \ldots, I_{D_{y}})$ using~(\ref{eq:informative_dimensions}) with $\tau$ 
      \For{\(n \gets 1 \textrm{ to } N\)}
      \State $\thb_{i,n}^* = \text{argmin}_{\thb} d_i^n(\thb)$ using (\ref{eq:rromc-masked-distance}) for all $i$
      \State Compute hyperboxes and $\{q_i^n\}_{i=1}^S$
    \EndFor
    \State $\thb_p \sim q(\thb)$, where $ q(\thb)= \frac{1}{N} \frac{1}{S} \sum_n \sum_i q_i^n (\thb)$
    \State Compute $\{w_p\}_p^P $ using (\ref{eq:rromc_weight})
    \State \Return $\{w_p, \thb_p\}_{p=1}^{P}$ 
    \EndProcedure
  \end{algorithmic}
\end{algorithm}

We offer an efficient \texttt{JAX} implementation benefiting from automatic differentiation (\texttt{grad}), vectorization (\texttt{vmap}), and just-in-time compilation (\texttt{jit}). 
In many cases, \rromc~executes in seconds where most SBI methods require minutes (see Section~\ref{sec:experiments}). 
This happens because, \rromc~can run using (approximately) only $N * S$ unique calls to the vectorized \texttt{JAX}-based simulator, as we explain below.

For the distractors (line 2), \texttt{grad} computes $\partial g_i/\partial \theta_j$ for all $i$ and $j$ in one call, and \texttt{vmap} evaluates multiple $\thb$ per seed, at once.
Repeating for all seeds costs $S$. 

Obtaining the per-data point proposal distributions $\{q_i^n\}_{i=1}^S$ (lines 3-6) consists of optimization (line 4) and hyperbox building (line 5) for each data point. Each of $S$ optimizations requires $T$ gradient descent steps, where $T$ is a user-defined hyperparameter. \texttt{jit} allows compiling multiple consecutive steps to optimize execution, in many cases achieving cost similar to a single execution, so the cost reduces to $\sim S$. Hyperbox building also requires of the order of $S$ steps when the required line searches are parallelized over each parameter dimensions using \texttt{vmap} (see Appendix \ref{app-sec:methods}). 
Thus the total cost of obtaining the proposal distributions is of the order of $N*S$. 

For generating the weighted samples, (\ref{eq:rromc_weight}) is evaluated on $P$ samples (line 8).
The simulator is only used for computing $\prod_n^N \sum_i^S \mathbb{1}_{C_{\epsilon}^{i,n}}(\thb_p)$ which requires evaluating $S * N$ different distance functions on $P$ points each. Using \texttt{vmap} in each distance function, the total cost is $S * N$. Evaluating $q(\cdot)$ and $p(\cdot)$ does not require calling the simulator. Finally, if the simulator is expensive, the $S *N$ cost of checking whether the samples are in the acceptance regions can be avoided by using the hyperbox or another model for the acceptance region, see \citet{ikonomov2020robust}. 

\subsection{Limitations and Discussion}

While R2OMC enables efficient inference in many scenarios, it has several limitations.
First, it requires differentiable simulators, making it a gray-box rather than a fully black-box method. 
Second, its performance depends critically on the success of the underlying optimization: if optimization fails, proposal samples will poorly approximate the posterior.
In that case, increasing the budget ($S$ in Algorithm~\ref{alg:r2omc}) will not necessarily improve accuracy as is the case with most neural-based methods.

% Other limitations arise in specific inference scenarios. 
In the multiple-observations setting, \rromc \,optimizes each observation independently. 
As a result, each proposed sample is $\epsilon$-close to at least one observation but not equally-close to others, and this cannot be improved since optimization is per observation. 
In such cases, accepting enough proposal samples ($w_p > 0$ in Eq.~\eqref{eq:rromc_weight}) requires increasing the $\epsilon$-threshold, which can broaden the posterior beyond the ground truth. 
However, under the assumption that all observations originate from a single (unknown) parameter configuration, many proposal samples are expected to make all observations plausible, so $\epsilon$ need not become large.

Finally, the optimization does not explicitly incorporate the prior. This enables cheap prior sensitivity analysis but proposal regions can occasionally fall in areas with low or no prior support, which may degrade inference quality.

% These factors should be considered when applying \rromc~to SBI tasks.

\section{Experiments}
\label{sec:experiments}

% Experiments Overview
We assess \rromc~on: 
(a) a Mixture of Gaussians (MoG) benchmark, 
(b) the synthetic SBI benchmark~\citep{lueckmann2021benchmarking},
(c) two image-based inference tasks and
(d) the Lotka-Volterra model.
Together, they span diverse challenges, including high-dimensionality, distractors, multiple observations, complex posteriors and real-world applications.

% Methods used for comparison and problem set-up
\textbf{Methods.} 
We evaluate \rromc~across all benchmarks.
For MoG, it is compared against three established neural methods, \NPE~\citep{greenberg2019automatic}, BayesFlow~\citep{radev2020bayesflow} and FlowMatching~\citep{wildberger2023flow}, using their official implementations provided by~\citet{tejero-cantero2020sbi}.
For SBI benchmark, it is compared to all methods reported by \citet{lueckmann2021benchmarking}.
For image-based tasks, we assess its applicability to high-dimensional problems without baseline comparisons. 
All experiments were run on a 12-core Dell XPS laptop (2.60GHz). 

% Metrics used
\textbf{Metrics.}
We evaluate all methods using the \CtwoST~score~\citep{gutmann2018likelihood, lueckmann2021benchmarking}, which measures similarity between two sets: samples from the approximate posterior and samples from the ground truth. 
A score of $0.5$ indicates indistinguishable sets (best), while $1.0$ indicates perfect separation (worst). 
As a classifier, we use a fully connected neural network and we report the mean \CtwoST~score over, at least, five independent runs.

Appendix~\ref{app-sec:experiments} provides additional details on the setup, methods, and metrics.

\begin{figure*}
    \centering
    \includegraphics[width=.19\linewidth]{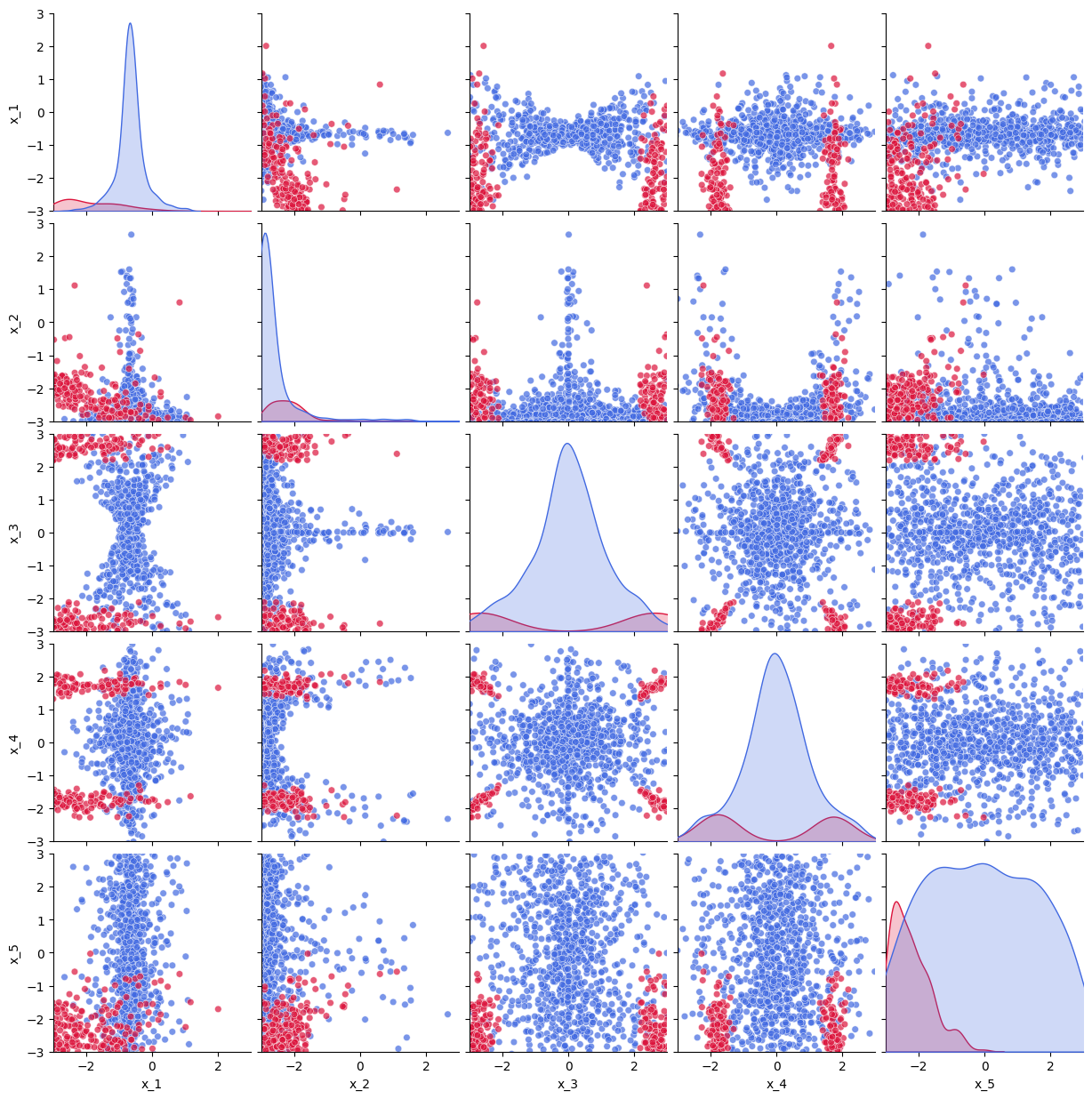}
    \includegraphics[width=.19\linewidth]{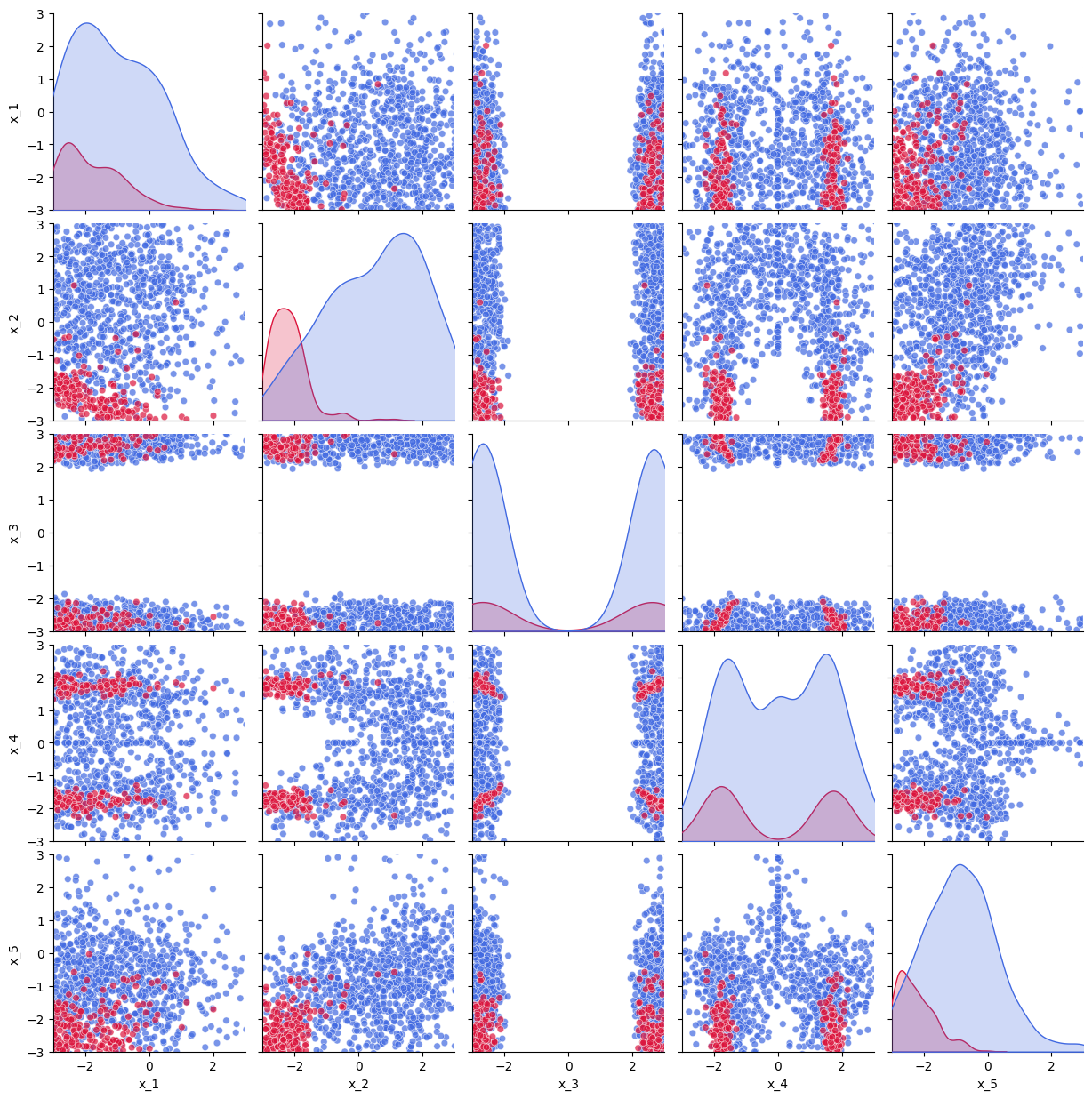}
    \includegraphics[width=.19\linewidth]{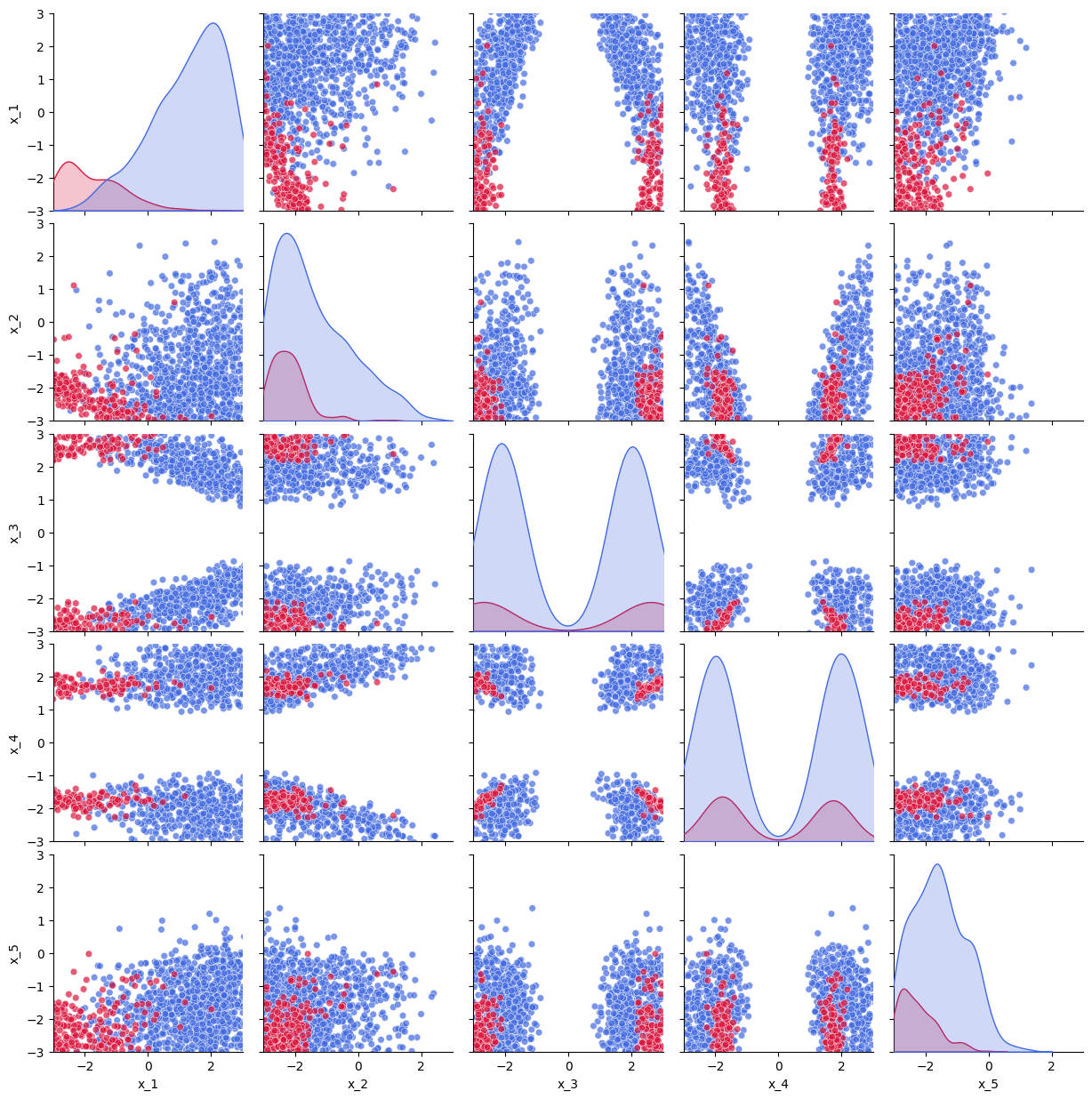}
    \includegraphics[width=.19\linewidth]{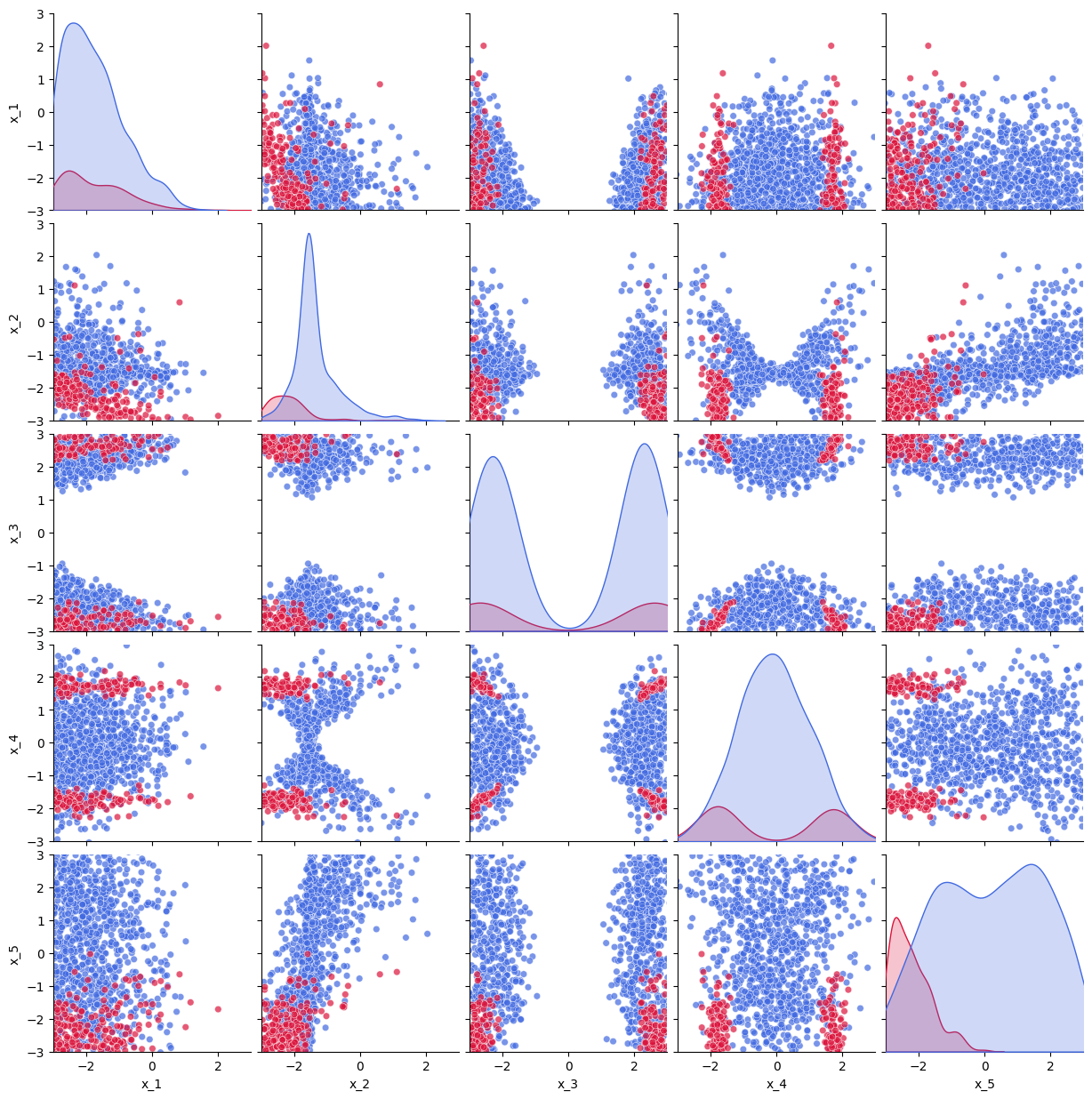}
    \includegraphics[width=.19\linewidth]{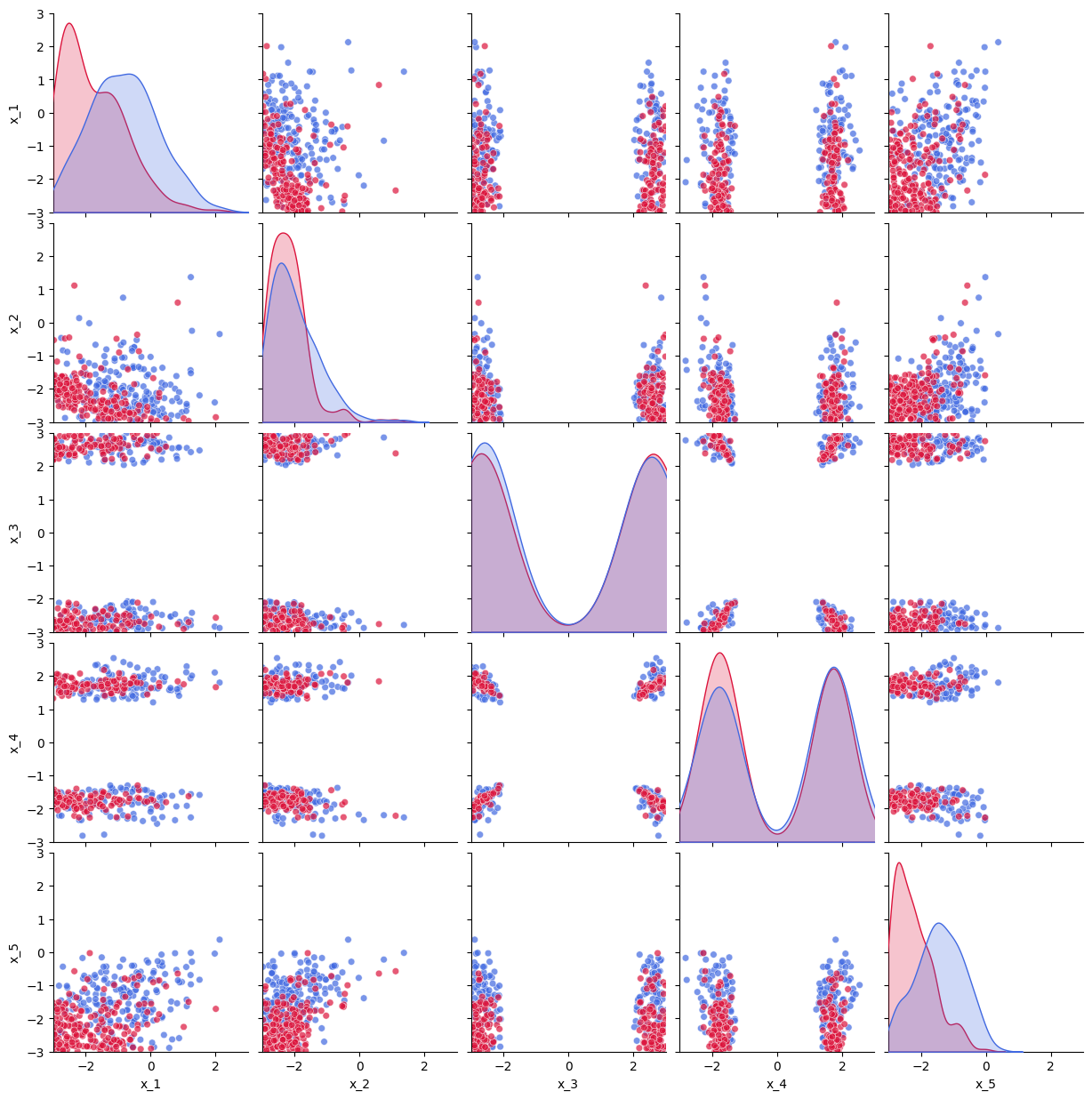}
    \caption{SLCP pairwise posteriors for multiple observations; blue: \rromc~samples, red: ground-truth samples. The four left panels show proposal samples per observation within $\epsilon$-distance, while the rightmost panel shows the subset of samples selected as closest across all observations using the \rromc~weighting scheme (Eq.~\ref{eq:rromc_weight}).
    }
    \label{fig:slcp_posterior}
\end{figure*}

\subsection{Mixture of Gaussians (MoG) benchmark}
\label{subsec:gaussian}

% Why we use the MoG benchmark
The MoG benchmark systematically evaluates the trade-off between accuracy and runtime for SBI methods as the parameter dimensionality $D$ increases, while keeping the underlying problem structure fixed.

% MoG setup
We consider two simulators:
\begin{itemize}
    \item $S_{\mathtt{base}}: \yb \sim \mathcal{N}(\thb, \sigma^2 \mathbf{I})$
    \item $S_{\mathtt{MoG}}: \yb \sim \frac{1}{2} \sum_{s \in \{+1, -1\}} \mathcal{N}(\thb + s\boldsymbol{\mu}, \sigma^2 \mathbf{I})$
\end{itemize}
which yield uni-modal and bi-modal posteriors, respectively.
For each simulator, we consider (i) a simple setting, where the output dimensionality matches the parameter vector ($D_y = D$), and (ii) a distractor setting, where $18$ uninformative dimensions drawn from $\mathcal{U}(-3, 3)$ are appended to the output ($D_y = D + 18$), resulting in four simulator configurations, in total.

Observations are vectors of zeros with small noise, matching the simulator's output dimension ($D$ or $D+18$).
For each simulator, we evaluate \rromc~against the three neural baselines introduced above, varying $D$ from 2 to 20 and the simulation budget from $1{,}000$ to $100{,}000$. 
A uniform prior $\mathcal{U}(\mathbf{-3}, \mathbf{3})$ is used in all cases.

Figure~\ref{fig:high_dim_benchmark} summarizes the results using success frontier plots, which show the minimum runtime required for successful inference across $D$. Inference is considered successful when the mean \CtwoST~score is $\leq 0.75$.
% Runtime measures both data generation and training the network.

All neural-based methods follow a consistent trend: as $D$ increases, the required simulation budget grows rapidly, often exceeding $100{,}000$ simulations, resulting in runtimes of over an hour (see Appendix~\ref{app-sec:experiments} for corresponding Budget vs. Dimension plots).
Runtimes increase more steeply for $S_{\mathtt{MoG}}$ than for $S_{\mathtt{base}}$ and more in the distractor settings than in the simple ones.
In contrast, \rromc~achieves successful inference in a few seconds across all $D$ and simulators.

\begin{figure}
    \centering
    \includegraphics[width=.49\linewidth]{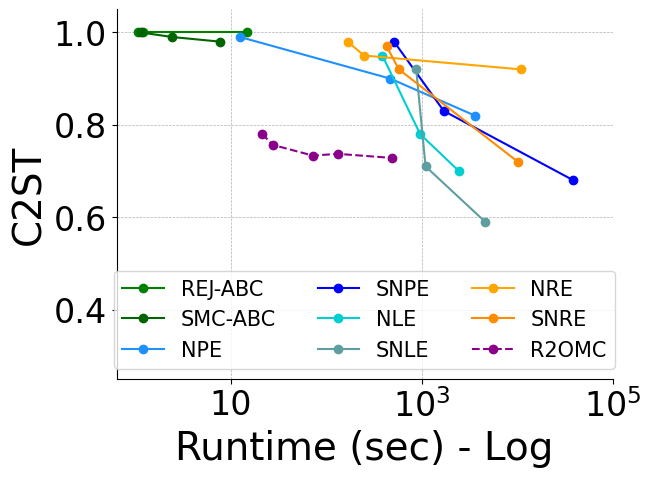}
    \includegraphics[width=.49\linewidth]{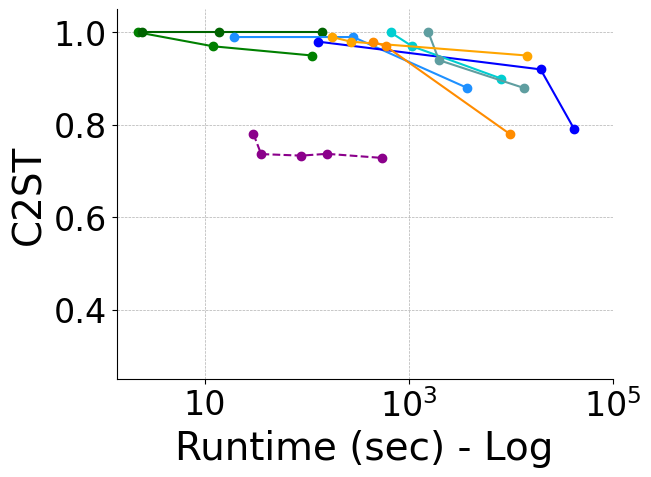}
    \caption{SLCP (left) and SLCP with distractors (right): C2ST score vs. runtime.}
    \label{fig:slcp_c2st_runtime}
\end{figure}

\vspace{-1ex}
\subsection{SBI benchmark}
\label{subsec:sbi-benchmark}

The SBI benchmark~\citep{lueckmann2021benchmarking} is widely used to evaluate SBI methods.
We select: SLCP (T.3) to test inference with multiple observations, SLCP with distractors (T.4) to assess robustness to uninformative dimensions and 2-Moons (T.8) for a bimodal posterior with narrowly-concentrated modes.

\vspace{-1ex}
\subsubsection{SLCP (T.3) and plus distractors (T.4)}

\begin{figure}
    \centering
    \includegraphics[width=.49\linewidth]{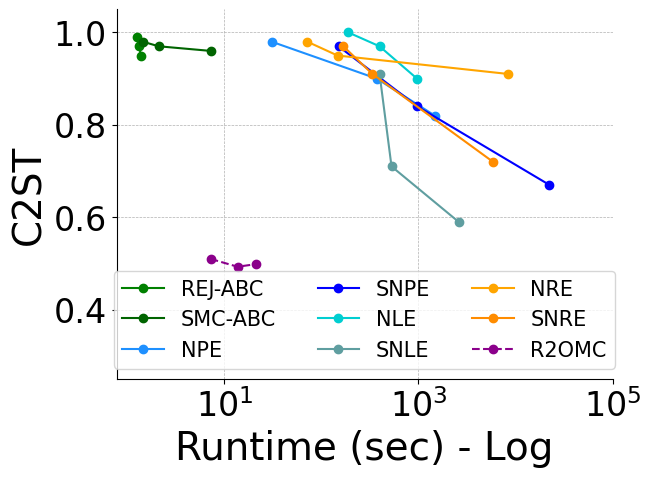}
    \includegraphics[width=.49\linewidth]{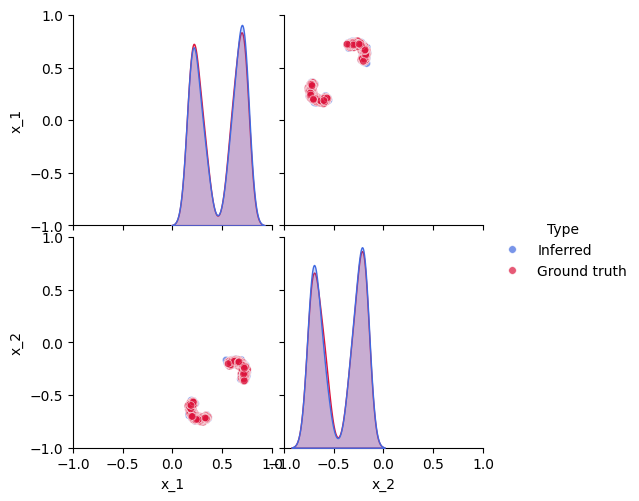}
    \caption{
      2-moons task. Left: C2ST score vs. runtime. Right: pairwise posterior.
    }
    \label{fig:sbi_two_moons}
\end{figure}

The SLCP (Simple Likelihood, Complex Posterior) task has a uniform prior over five parameters $\thb \in \mathcal{U}(-3, 3)^5$ and
four two-dimensional observations drawn from a Gaussian distribution whose mean and variance are nonlinear functions of $\thb$.
This results in a posterior characterized by four symmetric modes and sharp vertical boundaries.
T.4 augments the original SLCP task by appending 23 non-informative dimensions to each observation.

Figure~\ref{fig:slcp_c2st_runtime} reports the \CtwoST~score versus runtime for \rromc~and the methods of \citet{lueckmann2021benchmarking} on both SLCP variants.
\rromc~approximates the posterior accurately (\CtwoST~score $0.7$--$0.8$) in a few seconds, whereas other methods require substantially longer runtimes (often hours), because they require budgets of at least $10^4$ (often $10^5$) samples for similar performance.
The gap is especially pronounced in the presence of distractors, where most methods fail to achieve \CtwoST~scores below $0.8$ regardless of budget.

Figure~\ref{fig:slcp_posterior} illustrates how \rromc~handles multiple observations.
Proposal samples are first selected per observation (within $\epsilon$-distance) and then weighted according to Eq.~\ref{eq:rromc_weight} to retain those closest with respect to all observations.
We speculate that this is why \CtwoST~scores never reach the optimum of 0.5 and start to flatten out: increasing the budget ensures samples are $\epsilon$-close to individual observations, but not necessarily to the joint set. 
In our case, all proposed samples satisfy $\epsilon < 1$ per observation, yet the maximum distance across all observations reaches $\epsilon = 5$.

\vspace{-1ex}
\subsubsection{Two-moons (T.8)}

The 2-moons evaluates SBI methods for bimodal, crescent-shaped posteriors.
The simulator is $\yb|\thb \sim  (r\cos(\alpha) + 0.25, r \sin(\alpha)) + (-|\theta_1 + \theta_2| / \sqrt{2}, (-\theta_1 + \theta_2) / \sqrt{2})$ with $\alpha \sim \mathcal{U}(-\pi/2, \pi/2)$ and $r \sim \mathcal{N}(0.1, 0.01^2)$.
The problem is two-dimensional with a uniform prior $\thb \sim \mathcal{U}(\mathbf{-1}, \mathbf{1})$, 
producing a posterior with two half-moon modes.

Figure~\ref{fig:sbi_two_moons} compares \rromc~to the methods of \citet{lueckmann2021benchmarking}.
\rromc~achieves a near-optimal \CtwoST score ($\approx 0.5$) in a few seconds,
whereas competing methods require minutes to hours (budgets of $10^5$ samples) to reach comparable performance.
The pairwise posterior further demonstrates that \rromc~accurately captures both crescent-shaped modes. Additional results versus budget are reported in Appendix~\ref{app-sec:experiments}.

\vspace{-1ex}
\subsection{Image-based inference}
\label{subsec:image-based}

\begin{figure}
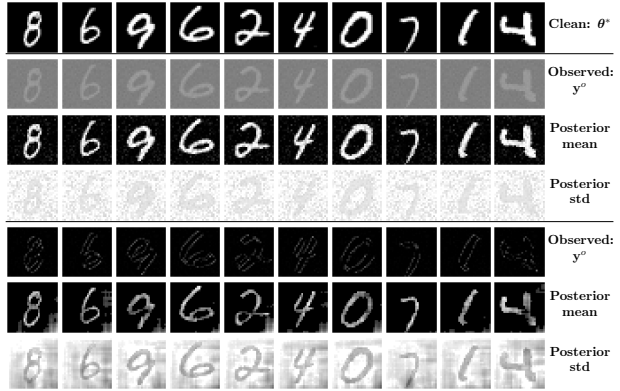

  \centering
  \resizebox{\linewidth}{!}{%
      \setlength{\tabcolsep}{1pt}
      \renewcommand{\arraystretch}{2}
      \begin{tabular}{*{10}{c} >{\centering\arraybackslash}m{0.1\textwidth}}
          \imagerow{images/image_pixelwise}{image_clean}{Clean: $\thb^*$}
          \hline
          \imagerow{images/image_pixelwise}{image_observation}{Observed: $\data$}
          \imagerow{images/image_pixelwise}{posterior_mean}{Posterior mean}
          \imagerow{images/image_pixelwise}{posterior_std}{Posterior std}
          \hline
          \imagerow{images/image_checkerboard}{image_observation}{Observed: $\data$}
          \imagerow{images/image_checkerboard}{posterior_mean}{Posterior mean}
          \imagerow{images/image_checkerboard}{posterior_std}{Posterior std}
      \end{tabular}
    }
  \caption{
    Image-based inference with \rromc.
    Top row: clean ground truth image $\thb^*$. 
    Rows 2--4: distorted observation, posterior mean, and posterior standard deviation 
    for pixel-wise intensity distortion.
    Rows 5--7: the same results for edge-detection filter distortion.
  }
  \label{fig:pixelwise}
\end{figure}

\vspace{-1ex}
We evaluate \rromc~on high-dimensional parameter spaces (784 dimensions) using image-based inference in noisy camera models.
Two simulators are considered on MNIST observations~\citep{lecun1998gradient}:
the first applies pixel-wise intensity changes, $y_{ij} = a\theta_{ij} + b + \boldsymbol{\epsilon}$,
with $a, b$ controlling contrast and brightness;
the second applies edge-detection filtering, $\yb = \text{filter}(\thb) + \boldsymbol{\epsilon}$,
via convolution with a checkerboard filter.
Both use Gaussian noise $\boldsymbol{\epsilon} \sim \mathcal{N}(\mathbf{0}, 0.1^2 \mathbf{I})$ and
uninformative prior $\thb \sim \mathcal{U}(0,1)^{28 \times 28}$.

Figure~\ref{fig:pixelwise} shows that \rromc~recovers posterior means that visually match the clean images, indicating the posterior mode is close to the true generating parameters
$\thb^*$, despite the completely uninformative prior.

While direct comparisons are not performed here, prior work highlights that SBI on images is challenging due to the curse of dimensionality. 
Methods like GATSBI~\citep{ramesh2022gatsbi} can succeed but require complex generative models with long, often unstable, training. 
In contrast, \rromc~achieves accurate inference with just 100 simulations and a runtime of a few seconds.

\vspace{-1ex}
\subsection{Lotka-Volterra}
\label{subsec:lotka-volterra}

\begin{figure*}[!t]
  \centering
  \begin{minipage}[t]{0.62\linewidth}
    \centering
    \vspace{2em}
    \small
    \captionof{table}{Lotka-Volterra SBC results (150 repetitions, budget\,=\,1{,}000).}
    \label{tab:lv_sbc}
    \begin{tabular}{lccccc}
      \toprule
      Parameter & SBC $\chi^2_{100}$ & $p$-value & Cov.\ 50\% & Cov.\ 90\% & Cov.\ 95\% \\
      \midrule
      $\alpha$ (prey birth)      & 97.79 & 0.544 & 0.500 & 0.900 & 0.953 \\
      $\beta$  (predation)       & 84.32 & 0.870 & 0.500 & 0.927 & 0.940 \\
      $\gamma$ (predator death)  & 81.63 & 0.910 & 0.527 & 0.907 & 0.960 \\
      $\delta$ (reproduction)    & 89.71 & 0.760 & 0.480 & 0.907 & 0.987 \\
      \bottomrule
    \end{tabular}
  \end{minipage}%                        % <-- % immediately after \end{minipage}
  \hfill%
  \begin{minipage}[t]{0.26\linewidth}
    \vspace{0pt}                         % <-- realigns [t] baseline after \captionof
    \centering
    \includegraphics[width=\linewidth]{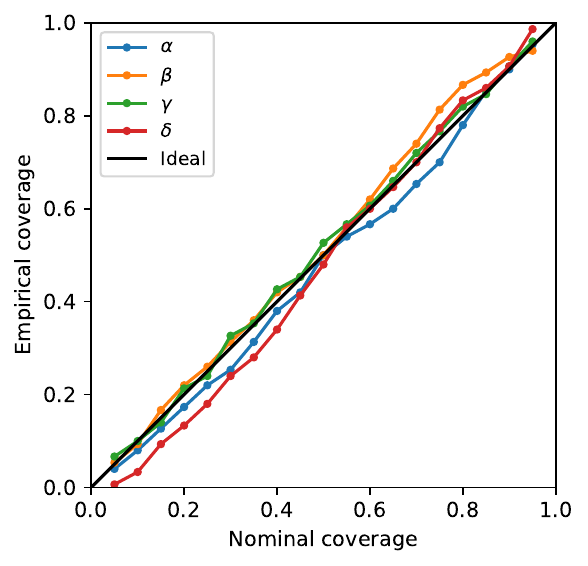}
  \end{minipage}
  \caption{
    Lotka-Volterra SBC summary.
    \textbf{Left (table):} $\chi^2$ uniformity tests ($k=100$ degrees of freedom) and empirical coverage suggest well-calibrated posteriors.
    \textbf{Right:} coverage calibration curve---empirical vs.\ nominal level across 150 repetitions; all parameters closely follow the diagonal.
  }
  \label{fig:lv_summary}
\vspace{-2ex}
\end{figure*}

\begin{figure}[!t]
  \centering
  \includegraphics[width=0.99\linewidth]{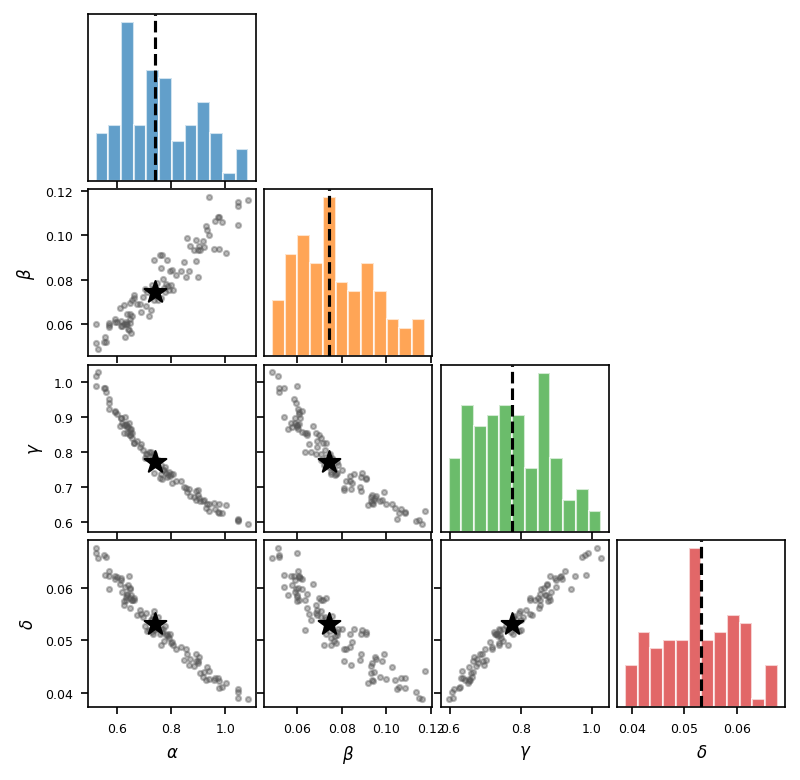}
  \caption{
    Lotka-Volterra pairwise posterior for a representative observation.
    The black star marks the true parameter $\thb^*$, which falls well within the posterior mass for all four parameters.
  }
  \label{fig:lv_posterior}
  \vspace{-3ex}
\end{figure}

The Lotka-Volterra predator-prey model is a standard benchmark in the SBI literature~\citep{lueckmann2021benchmarking}.
It provides a realistic test case where the data-generating process is governed by a system of ODEs.

\textbf{Setup.}
$X$ and $Y$ denote the prey and predator populations, with parameters $\alpha, \beta, \gamma, \delta$ governing their dynamics:
\[
\frac{dX}{dt} = \alpha X - \beta XY, \quad \frac{dY}{dt} = -\gamma Y + \delta XY
\]
Starting from $(X(0), Y(0)) = (30, 1)$ and integrating over $T = 20$ iterations,
the simulator produces a 20-dimensional time-series observation
$\yb = (x_{1,1}, x_{2,1}, \ldots, x_{1,10}, x_{2,10})$ corrupted by log-normal noise.
Parameters follow independent log-normal priors:
$\alpha, \gamma \sim \mathrm{LogNormal}(-0.125, 0.5)$,
$\beta, \delta \sim \mathrm{LogNormal}(-3, 0.5)$.
The simulator is implemented in \texttt{JAX} for compatibility with \rromc's gradient-based optimisation.

\textbf{Evaluation via SBC.}
Since our \texttt{JAX}-based ODE solver differs slightly from the Julia implementation of~\citet{lueckmann2021benchmarking},
ground-truth posterior samples are not available for a direct \CtwoST~comparison.
We therefore assess calibration via Simulation-Based Calibration~\citep[SBC,][]{hermans2021trust}, running 150 independent repetitions with a budget of 1{,}000 simulator calls each.
SBC draws $\thb^* \sim p(\thb)$, simulates $\yb^* \sim p(\yb|\thb^*)$, then draws $K$ posterior samples and records the rank of $\thb^*$ among them.
A well-calibrated posterior yields \emph{uniformly distributed} ranks, which we test with a $\chi^2$ statistic.
We additionally report \emph{empirical coverage}: the fraction of repetitions in which $\thb^*$ falls inside the $\alpha$-level credible interval. A calibrated posterior achieves empirical coverage $\approx \alpha$ for all levels.

\textbf{Results.}
Figure~\ref{fig:lv_summary}
(Table~\ref{tab:lv_sbc} on the left and coverage calibration curve on the right)
report the SBC $\chi^2$ statistics and the empirical coverage.
All four parameters were consistent with a uniform distribution under the $\chi^2$ test (p\,$>\,$0.54) and achieve near-nominal coverage, confirming well-calibrated posteriors within a budget of 1{,}000 calls.
Figure~\ref{fig:lv_posterior} shows the pairwise posterior, where the true parameter $\thb^*$ (black star) falls well within the posterior mass for all parameters, further indicating an accurate inference.

\vspace{-1ex}
\section{Conclusion}
\label{sec:conclusion}
\vspace{-1ex}
We have presented \rromc, a new Bayesian inference method for differentiable simulator models that addresses two major challenges in simulation-based inference:
scalability to high-dimensional parameter spaces and robustness to uninformative dimensions (distractors), all within fast runtimes.

Building on the ROMC framework, \rromc~recasts inference as a set of deterministic optimization problems solvable via gradient descent, enabling efficient inference in high-dimensional spaces. 
A gradient-based filtering step further masks distractors, while a novel adaptive importance sampling scheme selects posterior samples consistent with \emph{all} available iid observations.

Our \texttt{JAX}-based implementation leverages auto-differentiation and vectorization, enabling many parameter–seed combinations to be executed at the cost of a single call.
% Our \texttt{JAX}-based implementation leverages auto-differentiation and vectorization to significantly accelerate computation. 
Across diverse benchmarks, \rromc~consistently achieves high accuracy while requiring only a fraction of the computational time of state-of-the-art methods.

\section*{Acknowledgments}

This research was funded from the European Union’s Horizon Europe research and
innovation program under Grant Agreement No: 101135826 (\href{https://www.ai-dapt.eu/}{ai-dapt.eu}).

\bibliographystyle{unsrtnat}
\bibliography{references}

\section*{Checklist}

 \begin{enumerate}

 \item For all models and algorithms presented, check if you include:
 \begin{enumerate}
   \item A clear description of the mathematical setting, assumptions, algorithm, and/or model. [Yes]
   \item An analysis of the properties and complexity (time, space, sample size) of any algorithm. [Yes]
   \item (Optional) Anonymized source code, with specification of all dependencies, including external libraries. [Yes]
 \end{enumerate}

 \item For any theoretical claim, check if you include:
 \begin{enumerate}
   \item Statements of the full set of assumptions of all theoretical results. [Not Applicable]
   \item Complete proofs of all theoretical results. [Not Applicable]
   \item Clear explanations of any assumptions. [Not Applicable]
 \end{enumerate}

 \item For all figures and tables that present empirical results, check if you include:
 \begin{enumerate}
   \item The code, data, and instructions needed to reproduce the main experimental results (either in the supplemental material or as a URL). [Yes]
   \item All the training details (e.g., data splits, hyperparameters, how they were chosen). [Yes]
   \item A clear definition of the specific measure or statistics and error bars (e.g., with respect to the random seed after running experiments multiple times). [Yes]
   \item A description of the computing infrastructure used. (e.g., type of GPUs, internal cluster, or cloud provider). [Yes]
 \end{enumerate}

 \item If you are using existing assets (e.g., code, data, models) or curating/releasing new assets, check if you include:
 \begin{enumerate}
   \item Citations of the creator If your work uses existing assets. [Yes]
   \item The license information of the assets, if applicable. [Yes]
   \item New assets either in the supplemental material or as a URL, if applicable. [Not Applicable]
   \item Information about consent from data providers/curators. [Not Applicable]
   \item Discussion of sensible content if applicable, e.g., personally identifiable information or offensive content. [Not Applicable]
 \end{enumerate}

 \item If you used crowdsourcing or conducted research with human subjects, check if you include:
 \begin{enumerate}
   \item The full text of instructions given to participants and screenshots. [Not Applicable]
   \item Descriptions of potential participant risks, with links to Institutional Review Board (IRB) approvals if applicable. [Not Applicable]
   \item The estimated hourly wage paid to participants and the total amount spent on participant compensation. [Not Applicable]
 \end{enumerate}

 \end{enumerate}

\clearpage
\appendix
\onecolumn
\input{./appendix}

\end{document}

%% file: appendix.tex
\section{Simulation Budget}
\label{app-sec:compute-budget}

In simulation-based inference (SBI), an important consideration is the ability of a method to infer the posterior distribution under a restricted simulation budget.
The term \textit{simulation budget} typically refers to the number of allowed simulator evaluations — that is, the number of times the simulator is called to generate synthetic data.
Each simulator call produces a pair $(\boldsymbol{\theta}_i, \mathbf{y}_i)$, where $\mathbf{y}_i$ denotes the simulated observation corresponding to parameters $\boldsymbol{\theta}_i$.

Recently, with the advent of neural-based SBI methods, the simulation budget is often associated with the number of unique parameter–observation pairs used to train the neural density estimator.
In multi-round or active learning approaches, simulations from previous rounds can be reused for training, so the total number of simulator calls and the size of the training dataset are not necessarily identical.
Nevertheless, the number of unique simulations remains limited by the simulation budget, and neural estimators rely on these unique instances to learn an accurate posterior.

The simulation budget is an important consideration because it directly relates to computational cost. This arises for two main reasons: (a) each simulator call can be computationally expensive, and (b) training the neural density estimator on the resulting simulated dataset requires additional computation, which grows with dataset size. A common third cost is the time required to draw samples from the trained estimator; however, this cost is typically independent of the simulation budget and is therefore not counted here.
Therefore, simulation budget typically affects costs incured by (a) and/or (b).

For neural-based SBI methods, the simulation budget impacts both sources of computational cost: (a) the number of simulator calls and (b) the cost of training the neural density estimator. In contrast, for more traditional likelihood-free inference methods, such as approximate Bayesian computation (ABC), the simulation budget primarily affects (a), since these methods do not explicitly train a separate estimator.

Lately, modern computational frameworks, such as \texttt{JAX}, allow simulators to be implemented in a way that accelerates computation through vectorization. For example, a batch of parameters ${\boldsymbol{\theta}_i}$ for $i=1, \ldots, N$ can be evaluated on the same random seed $s$, producing outputs ${\mathbf{y}i}{i=1}^N$ at roughly the cost of a single simulator evaluation.
More generally, a grid of parameter–seed pairs ${\boldsymbol{\theta}_i, s_j}$ for $i = 1, \dots, N$ and $j = 1, \dots, S$ can also be computed efficiently, effectively generating $N \times S$ simulated pairs $(\boldsymbol{\theta}, \mathbf{y})$ at the cost of a `vectorized' evaluation, i.e., in significantly reduced computational overhead compared to $N \times S$ sequential evaluations.

These advancements can substantially reduce computational cost (a), as a single simulator call can now produce multiple $(\boldsymbol{\theta}, \mathbf{y})$ pairs through vectorized evaluation. However, they do not reduce cost (b), since training the neural density estimator still scales with the total number of simulated instances.
Therefore, although, this advancements benefits neural-based methods in terms of cost incured by (a), does not help them for (b).
This motivates the development of a method that avoids the computational cost of training an external neural estimator.

Therefore, we clarify that for \rromc, the term simulation budget is not directly applicable as a metric and, when we use it for comparison purposes, it refers to the number of independent calls to a vectorized simulator, where each call can produce multiple $(\boldsymbol{\theta}, \mathbf{y})$ pairs.
This definition emphasizes that for \rromc, the budget counts independent simulator evaluations rather than the total number of resulting dataset instances.

% However, as it becomes clear from runtime plots, it is indeed that case that a call to a vectorized simulator does nt 

\clearpage

\section{Methods}
\label{app-sec:methods}

Below, we provide information and the default hypeparameter setting that we use in the four methods we compare in the Introductory Example (Figure~\ref{fig:concept_example}) and the Mixture of Gaussians (MoG) benchmark~\ref{subsec:gaussian}.
% we compare \rromc~(our proposal) against three established neural-based methods—Neural Posterior Estimator~\citep[NPE,][]{greenberg2019automatic}, BayesFlow~\citep{radev2020bayesflow}, and Flow Matching Posterior Estimator~\citep[FMPE,][]{wildberger2023flow}.
% Below we provide 

\paragraph{R2OMC.}

A detailed description of the method can be found in Algorithm~\ref{alg:r2omc}. 

In Step 2 of Algorithm~\ref{alg:r2omc}, i.e., computing the uninformative dimensions, we typically use around 50 samples from the distribution \(p(u)\) and another 50 samples from \(p(\thb)\), with the parameter \(\tau\) set to machine precision, i.e., \(\tau \approx 0\). Similar results can be obtained with fewer samples, such as 10 or 20.

For Step 3, i.e., computing the optimal points \(\thb_{i,n}^*\), we use the Adam optimizer, with a learning rate between 0.01 and 0.2, depending on the specific problem. Our default choice is 0.01. The number of optimization steps varies between 50 and 400, with 50 as the default. The distance function \(d(\thb)\) is defined as the squared Euclidean distance between the simulator output and the observation, i.e., \(d(\thb) = \|\yb - \data_n\|^2_2\).
The number of seeds \(S\) used ranges from 100 to 10{,}000, with 1000 as the default.
The number of seeds corresponds to the simulation budget; for example, if a budget of 10,000 runs is available, we use (at most) 10,000 seeds.
Typically, we select 80\% of the seeds based on the best \(d_i^n(\thb_{i,n}^*)\).
However, the percentage of accepted seeds can vary between 50\% and 100\%, depending on the absolute values of \(d_i^n(\thb_{i,n}^*)\) for \(i = 1, \dots, S\).

For Step 5, i.e., constructing the hyperboxes, we use Algorithm~\ref{alg:region_construction}. We normally set \(\epsilon\) to be twice the distance of the `worst' accepted seed, i.e., \(\epsilon = 2 \max_i d_i^n(\thb_{i,n}^*)\), where \(i\) is an index over the accepted seeds. Alternatively, \(\epsilon\) can be set to a fixed value such as $0.1$. The typical settings for step size, number of steps, and refinements are \(\eta = 0.1\), \(L = 100\), and \(R = 1\), respectively.

For Step 7, i.e., sampling from the proposal distribution, we generate between 1000 and 100,000 candidate samples, with the default being 2000. From these candidates, we select 1000 via weighted sampling.
For computing the weights, $\{w_p\}_{p=1}^{P}$, we set $\epsilon$ to a value about twice the distance of the `worst' accepted seed, i.e., $\epsilon = 2 \max_{i,n} d_i^n(\thb_{i,n}^*)$. However, this can vary depending on the problem, and by testing different values.

% By default, we keep 1000 posterior samples, which are used to compute the \texttt{C2ST} score using another 1000 samples drawn from the true posterior.

\begin{algorithm}[!ht]
  \caption{Hyperbox Computation \newline
    \textit{Input:} Optimal point \(\thb^*\), distance function \(d(\thb)\), Jacobian \(\jac\) at \(\thb^*\) \newline
    \textit{Parameters:} Step size \(\eta\), number of refinements \(R\), number of steps \(L\), distance threshold \(\epsilon\)}
  \label{alg:region_construction}
  \begin{algorithmic}[1]
    \State Compute eigenvectors \(\vb_d\) of \(\jac^T\jac\) {\scriptsize (\(d = 1,\ldots,D)\)}
    \For{\(d = 1\) to \(D\)}
      \State Initialize endpoint \(\Tilde{\thb} \gets \thb^*\)
      \State Set refinement counter \(r \gets 0\)
      \Repeat{}
        \State Set step counter \(l \gets 0\)
        \Repeat{}
          \State Increment step counter \(l \gets l + 1\)
          \State Update \(\Tilde{\thb} \gets \Tilde{\thb} + \eta \vb_d\)
        \Until{\(d(\thb) > \epsilon\) or \(l = L\)}
        \State Move back one step \(\Tilde{\thb} \gets \Tilde{\thb} - \eta \vb_d\)
        \State Half the step size \(\eta \gets \eta/2\)
        \State Increment refinement counter \(r \gets r + 1\)
      \Until{\(r = R\)}
      \State Set \(\Tilde{\thb}\) as the endpoint
      \State Repeat steps 3-15 for \(\vb_d = - \vb_d\) and set \(\Tilde{\thb}\) as the negative endpoint
    \EndFor
    \State Define a uniform distribution \(q\) over the hyperbox
    \State \Return \(q\)
  \end{algorithmic}
\end{algorithm}

\paragraph{NPE}

For Neural Posterior Estimation (NPE), we use the variant denoted as \texttt{NPE\_C}~\citep{greenberg2019automatic} in the SBI benchmark~\citep{lueckmann2021benchmarking}.
\texttt{NPE\_C} estimates the posterior by training a neural network \( F(\mathbf{y}, \boldsymbol{\phi}) \) to approximate \( p(\boldsymbol{\theta} \mid \mathbf{y}) \) through a density estimator \( q_F(\mathbf{y}, \boldsymbol{\phi})(\boldsymbol{\theta}) \).
In the experiments, we use a neural spline flow as density estimator (\texttt{sbi.utils.get\_nn\_models.posterior\_nn()}).
The NPE configuration is summarized in Table~\ref{tab:npe_config}.

\begin{table}[h!]
\centering
\caption{Configuration of the Neural Posterior Estimation (NPE\_C) method.}
\label{tab:npe_config}
\begin{tabular}{lll}
\toprule
\textbf{Component} & \textbf{Parameter} & \textbf{Value} \\
\midrule
\multirow{5}{*}{\textbf{Density estimator} (\texttt{posterior\_nn})}
 & \texttt{model} & \texttt{'nsf'} \\
 & \texttt{hidden\_features} & 100 \\
 & \texttt{num\_transforms} & 8 \\
 & \texttt{num\_bins} & 10 \\
 & \texttt{z\_score\_x} & \texttt{'independent'} \\
 & \texttt{z\_score\_theta} & \texttt{'independent'} \\
\midrule
\multirow{3}{*}{\textbf{Training configuration} (\texttt{.train()})}
 & \texttt{training\_batch\_size} & 500 \\
 & \texttt{max\_num\_epochs} & 1000 \\
 & \texttt{force\_first\_round} & \texttt{True} \\
\bottomrule
\end{tabular}
\end{table}

\paragraph{BayesFlow.}

BayesFlow~\citep{radev2020bayesflow} is equivalent to Neural Posterior Estimation (NPE) with the key contribution of adding an
embedding network that automatically learns from the data appropriate summary statistics.
The \texttt{sbi} package provides access to these embeddings through the \texttt{sbi.neural\_nets.embedding\_nets()} interface.
In our experiments, we use the fully connected embedding network (\texttt{FCEmbedding}) and the default BayesFlow density estimator. 
The corresponding configurations are summarized in Table~\ref{tab:bayesflow_config}.

\begin{table}[h!]
\centering
\caption{Configuration of the BayesFlow embedding network and density estimator.}
\label{tab:bayesflow_config}
\begin{tabular}{lll}
\toprule
\textbf{Component} & \textbf{Parameter} & \textbf{Value} \\
\midrule
\multirow{3}{*}{\textbf{Embedding network} (\texttt{FCEmbedding})} 
 & \texttt{output\_dim} & 20 \\
 & \texttt{num\_layers} & 2 \\
 & \texttt{num\_hiddens} & 50 \\
\midrule
\multirow{5}{*}{\textbf{Density estimator} (\texttt{posterior\_nn})}
 & \texttt{model} & \texttt{'nsf'} \\
 & \texttt{hidden\_features} & 100 \\
 & \texttt{num\_transforms} & 8 \\
 & \texttt{z\_score\_x} & \texttt{'independent'} \\
 & \texttt{z\_score\_theta} & \texttt{'independent'} \\
\midrule
\multirow{3}{*}{\textbf{Training configuration} (\texttt{.train()})}
 & \texttt{training\_batch\_size} & 500 \\
 & \texttt{max\_num\_epochs} & 1000 \\
 & \texttt{force\_first\_round} & \texttt{True} \\
\bottomrule
\end{tabular}
\end{table}

\paragraph{FMPE.}

Flow Matching Posterior Estimation (FMPE)~\citep{{wildberger2023flow}} builds upon advances in generative modeling with continuous normalizing flows.
Similar in spirit to diffusion models, FMPE replaces discrete flow transformations with a continuous-time formulation based on flow matching.
In our experiments, we use the \texttt{sbi} implementation (\texttt{.FMPE()}) with a multilayer perceptron (\texttt{mlp}) as vector field estimator (\texttt{vf\_estimator}). The default configuration is summarized in Table~\ref{tab:fmpe_config}.

\begin{table}[h!]
\centering
\caption{Configuration of the Flow Matching Posterior Estimation (FMPE) method.}
\label{tab:fmpe_config}
\begin{tabular}{lll}
\toprule
\textbf{Component} & \textbf{Parameter} & \textbf{Value} \\
\midrule
\multirow{1}{*}{\textbf{Vector field estimator}}
 & \texttt{vf\_estimator} & \texttt{'mlp'} \\
\midrule
\multirow{3}{*}{\textbf{Training configuration} (\texttt{.train()})}
 & \texttt{training\_batch\_size} & 500 \\
 & \texttt{max\_num\_epochs} & 1000 \\
 & \texttt{force\_first\_round\_loss} & \texttt{True} \\
\bottomrule
\end{tabular}
\end{table}

\clearpage

\section{Introductory Example}
\label{app-sec:intro-example}

\subsection{Problem Definition}

We provide further details on the introductory example of Figure~\ref{fig:concept_example}.  
The example comes with three variants, that we refer to as: (1) \textbf{Simple}, (2) \textbf{High-dimensional}, and (3) \textbf{Distractors}.
Simple (1) and High-dimensional (2) share the same setup.
Between them, the only difference is the parameter dimension $D$ and the output dimension $D_y$ which are: $D=D_y=2$ in (1) and $D=D_y=10$ in (2).

In both (1) and (2), the simulator is:
\begin{equation}
  \label{eq:intro-simulator}
  \yb \sim \frac{1}{2} \sum_{s \in \{+1, -1\}} \mathcal{N}(\thb + s\boldsymbol{\mu}, \sigma^2 \mathbf{I}), 
  \quad \boldsymbol{\mu} = \mathbf{1}, \ \sigma = 0.2.
\end{equation}

In the Distractors (3) variant, the parameter space is $D = 2$, but we add $D_{\text{dist}} = 18$ distractor dimensions to the output, yielding $D_y = 20$.
Thus, in (3), the simulator becomes:

\begin{equation}
  \label{eq:intro-simulator-distractor}
  \yb = (\mathbf{y}^{(1)}, \mathbf{y}^{(2)}), \quad
  \mathbf{y}^{(1)} \sim \frac{1}{2} \sum_{s \in \{+1, -1\}} \mathcal{N}(\thb + s\boldsymbol{\mu}, \sigma^2 \mathbf{I}), \quad
  \mathbf{y}^{(2)} \sim \mathcal{U}(-\mathbf{3}, \mathbf{3}),
\end{equation}
where $\boldsymbol{\mu} = \mathbf{1}$ and $\sigma=0.2$.  
For all variants, the prior is $\mathcal{U}(-\mathbf{3}, \mathbf{3})$ and the observation is $\data = \mathbf{0} \in \mathbb{R}^{D_y}$.

\subsection{Ground Truth Posterior}

\paragraph{Simple and High-dimensional Case.}

The posterior is given by
\begin{align}
  p(\thb \mid \data) &\propto p(\thb) p(\data \mid \boldsymbol{\theta}) \\
  &\propto p(\thb) \left[ \mathcal{N}(\data; \boldsymbol{\theta} - \boldsymbol{\mu}, \sigma^2\mathbf{I}) + \mathcal{N}(\data; \boldsymbol{\theta} + \boldsymbol{\mu}, \sigma^2\mathbf{I}) \right] \label{eq:intro-simple-1} \\
  &\propto p(\thb) \left[ \mathcal{N}(\boldsymbol{\theta}; \boldsymbol{\mu}, \sigma^2\mathbf{I}) + \mathcal{N}(\boldsymbol{\theta}; -\boldsymbol{\mu}, \sigma^2\mathbf{I}) \right] \label{eq:intro-simple-2} \\
  &\propto
    \begin{cases}
      \mathcal{N}(\boldsymbol{\theta}; \boldsymbol{\mu}, \sigma^2\mathbf{I}) + \mathcal{N}(\boldsymbol{\theta}; -\boldsymbol{\mu}, \sigma^2\mathbf{I}), & \boldsymbol{\theta} \in [-3, 3]^D, \\
      0, & \text{otherwise}.
    \end{cases}
    \label{eq:intro-simple}
\end{align}

From~(\ref{eq:intro-simple-1}) to~(\ref{eq:intro-simple-2}), we use the property: $\mathcal{N}(\yb; \thb - \boldsymbol{\mu}, \Sigma) = \mathcal{N}(\thb; \yb + \boldsymbol{\mu} , \Sigma)$.

\paragraph{Distractors Case.}

In the distractors variant, the posterior over $\boldsymbol{\theta}$ remains unchanged.  
Formally:
\begin{align}
  p(\boldsymbol{\theta} \mid \data)
  &\propto p(\boldsymbol{\theta}) p(\data \mid \boldsymbol{\theta}) \\
  &\propto p(\boldsymbol{\theta}) p((\mathbf{y}^{o,(1)}, \mathbf{y}^{o,(2)}) \mid \boldsymbol{\theta}) \label{eq:intro-distractors-1} \\
  &\propto p(\boldsymbol{\theta}) \, p(\mathbf{y}^{o,(1)} \mid \boldsymbol{\theta}) \, p(\mathbf{y}^{o,(2)} \mid \boldsymbol{\theta}) \label{eq:intro-distractors-2} \\
  &\propto p(\boldsymbol{\theta}) \, p(\mathbf{y}^{o,(1)} \mid \boldsymbol{\theta}) \, p(\mathbf{y}^{o,(2)}) \label{eq:intro-distractors-3} \\
  &\propto
    \begin{cases}
      \mathcal{N}(\boldsymbol{\theta}; \boldsymbol{\mu}, \sigma^2\mathbf{I}) + \mathcal{N}(\boldsymbol{\theta}; -\boldsymbol{\mu}, \sigma^2\mathbf{I}), & \boldsymbol{\theta} \in [-3, 3]^D, \\
      0, & \text{otherwise}.
    \end{cases}
  \label{eq:intro-distractors}
\end{align}

Steps (\ref{eq:intro-distractors-1}) to (\ref{eq:intro-distractors-2}) use independence between $\mathbf{y}^{o,(1)}$ and $\mathbf{y}^{o,(2)}$, and steps (\ref{eq:intro-distractors-2}) to (\ref{eq:intro-distractors-3}) use that $\mathbf{y}^{o,(2)}$ is independent of $\thb$.

\subsection{Experimental Setup and Hyperparameter Configuration}

All methods were evaluated for simulation budgets of
$
[1{,}000, 10{,}000, 30{,}000, 40{,}000, 50{,}000]
$.
For each method, we drew 1{,}000 samples from the approximate posterior and compared them against 1{,}000 ground-truth samples using the C2ST score.  
Each experiment was repeated across 5 independent random seeds, so we report the average C2ST scores among the independent runs.

The experiment goal is to analyze the performance of each method in two regimes.
The \textbf{low-budget/low-runtime} regime, where the budget is always $1{,}000$ and the
\textbf{high-budget/high-runtime} regime, where the budget is between $10{,}000$ and $50{,}000$.
In the low-runtime regime, we simply report the mean C2ST score and runtime for a budget of $1{,}000$.  
In the high-runtime regime, we select the minimum budget that achieves an average C2ST below 0.65. If no budget meets this threshold, we select the budget with the lowest C2ST (which is always 50{,}000 in this case).
We report the mean C2ST and runtime of the selected budget

Below, we detail the hyperparameter configuration for each method, highlighting the deviations from the default settings.

\paragraph{ROMC} 
For \rromc, we used at most a maximum of $S = 1{,}000$ seeds, corresponding to roughly 1{,}000 independent simulator calls, as this budget was enough for a near-optimal C2ST score.
The modified fitting parameters are: \texttt{\{"pcg\_to\_keep":1.\}}, i.e., we keep all optimization ending points $\{\thb_i^*\}_{i=1}^{S}$.

\paragraph{NPE-C} 
The modified fitting parameters are:
\texttt{\{"batch\_size":100, "training\_batch\_size":100\}}

\paragraph{BayesFlow} 
The modified fitting parameters are: \texttt{\{"embedding\_net\_output\_dim": 2\}} for the Simple and Distractor variants, and \texttt{10} for the High-dimensional variant.

\paragraph{Flow-Matching} All arguments were default.

\subsection{Conclusions}

Figure~\ref{fig:concept_example} illustrates the key-findings presented below.

In the simple scenario (1), most methods perform well across both low- and high-budget regimes. Only FMPE requires a substantial simulation budget and runtime to succeed, whereas the remaining methods already achieve good performance under low-budget conditions.

However, the picture changes as the task becomes more challenging—either through the addition of distractors (2) or by increasing dimensionality (3). In these settings, all neural-based methods fail to perform adequately with a low simulation budget and require significantly higher budgets, and consequently higher runtimes, to succeed. BayesFlow, in particular, fails to converge even under the high-budget regime. In contrast, \rromc~achieves competitive performance using only 1,000 simulations, resulting in substantially lower runtime requirements.

Overall, the results highlight an important trend: as dimensionality increases or distractor dimensions are introduced, neural-based methods demand increasingly larger simulation budgets to achieve reliable performance. While a budget of 50,000 simulations is not prohibitive—modern infrastructure can handle even 100,000 simulations within reasonable runtimes—the observed trend is clear. With increases in dimensionality, additional distractors, or more complex simulators (beyond simple Gaussian noise), the required number of simulations grows rapidly. Consequently, under strict budget and runtime constraints, these neural-based approaches are likely to face significant limitations.

\clearpage
\section{Experiments: Additional Information}
\label{app-sec:experiments}

We here provide some additional information on the experiments presented in the main text.

\subsection{Mixture of Gaussians (MoG) Benchmark}

The Mixture of Gaussians (MoG) benchmark evaluates how simulation-based inference (SBI) methods scale with increasing problem dimensionality.
Specifically, it assesses whether higher-dimensional problems require larger simulation budgets to achieve accurate inference, i.e., whether there exists a correlation between dimensionality and the required number of simulations.

To this end, we consider four simulator variants:
$S_{\texttt{base}}$ (1), $S_{\texttt{base}}^{\texttt{dist}}$ (2), $S_{\texttt{MoG}}$ (3), and $S_{\texttt{MoG}}^{\texttt{dist}}$ (4).
Each simulator is tested across different dimensionalities, $D = {2, 5, 10, 15, 20}$.
For each configuration, we evaluate every method’s performance across multiple simulation budgets using the C2ST metric.

\subsubsection{Problem definition and ground truth posterior}

\textbf{Base simulator - without distractors (1):}

\begin{tabular}{@{}ll@{}}
  \textbf{Prior} & $\thb \sim \mathcal{U}(-\mathbf{3}, \mathbf{3})$ \\ [5pt]
  \textbf{Simulator} & $\yb \sim \mathcal{N}(\thb + \boldsymbol{\mu}, \sigma^2 \mathbf{I}) \label{eq:mog_base}$, with $\boldsymbol{\mu} = \mathbf{1}, \sigma^2 = 0.2^2$ \\ [5pt]
  \textbf{Dimensionality} & $\thb \in \mathbb{R}^{D}$, $\yb \in \mathbb{R}^{D}$ \\ [5pt]
  \textbf{Observation} & $\data = (0, \ldots, 0)$ where $\data \in \mathbb{R}^D$ \\ [5pt]
  \textbf{Posterior} & $ p(\thb | \data) \propto
                                    \begin{cases}
                                      \mathcal{N}(\thb; -\boldsymbol{\mu}, \sigma^2 \mathbf{I}), & \text{if } \thb \in [-3, 3]^D, \\
                                      0, & \text{otherwise}.
                                    \end{cases} $ \\ [5pt]
\end{tabular}

\textbf{Proof of the posterior:}
\begin{align}
  p(\thb | \data) &\propto p(\thb) p(\data | \thb) \label{eq:mog_base_posterior-1} \\
                  &\propto p(\thb) \mathcal{N}(\data; \thb + \boldsymbol{\mu}, \sigma^2 \mathbf{I}) \label{eq:mog_base_posterior-2} \\
                  &\propto p(\thb) \mathcal{N}(\thb; \data - \boldsymbol{\mu}, \sigma^2 \mathbf{I}) \label{eq:mog_base_posterior-3} \\
                  &\propto
                    \begin{cases}
                      \mathcal{N}(\thb; - \boldsymbol{\mu}, \sigma^2 \mathbf{I}), & \text{if } \thb \in [-3, 3]^D, \\
                      0, & \text{otherwise}.
                    \end{cases} \label{eq:mog_base_posterior}
\end{align}

From~(\ref{eq:mog_base_posterior-2}) to~(\ref{eq:mog_base_posterior-3}), we apply the property: $\mathcal{N}(\yb; \thb + \boldsymbol{\mu}, \Sigma) = \mathcal{N}(\thb; \yb - \boldsymbol{\mu} , \Sigma)$.

\textbf{Base simulator - with distractors (2):}

\begin{tabular}{@{}ll@{}}
  \textbf{Prior} & $\thb \sim \mathcal{U}(-\mathbf{3}, \mathbf{3})$ \\ [5pt]
  \textbf{Simulator} & $\yb = (\mathbf{y^{(1)}}, \mathbf{y^{(2)}})$, \\ [5pt]
  & $\mathbf{y^{(1)}} \sim \mathcal{N}(\thb + \boldsymbol{\mu}, \sigma^2 \mathbf{I})$, $\boldsymbol{\mu} = \mathbf{1}$, $\sigma^2=0.2^2$, \\ [5pt]
  & $\mathbf{y^{(2)}} \sim \mathcal{U}(\mathbf{-3}, \mathbf{3})$ \\ [5pt]
  \textbf{Dimensionality} & $\thb \in \mathbb{R}^D,  \yb \in \mathbb{R}^{D + 18}$, $\yb^{(1)} \in \mathbb{R}^{D}$, $\yb^{(2)} \in \mathbb{R}^{18}$\\ [5pt]
  \textbf{Observation} & $\data = (\mathbf{y^{o, (1)}}, \mathbf{y^{o, (2)}})$, where both $\mathbf{y^{o, (1)}}$ and $\mathbf{y^{o, (2)}}$  are $(0, \ldots, 0) $ \\ [5pt]
  \textbf{Posterior} & $ p(\thb | \data) \propto
                                    \begin{cases}
                                      \mathcal{N}(\thb; -\boldsymbol{\mu}, \sigma^2 \mathbf{I}), & \text{if } \thb \in [-3, 3]^D, \\
                                      0, & \text{otherwise}.
                                    \end{cases} $ \\ [5pt]
\end{tabular}

The ground truth posterior is the same as in~(\ref{eq:mog_base_posterior}), as the distractors do not affect the posterior, see (\ref{eq:intro-distractors}).

\textbf{MoG simulator - without distractors (3):}

\begin{tabular}{@{}ll@{}}
  \textbf{Prior} & $\thb \sim \mathcal{U}(-\mathbf{3}, \mathbf{3})$ \\ [5pt]
  \textbf{Simulator} & $\yb \sim \frac{1}{2} \sum_{s \in \{-1, +1\}} \mathcal{N}(\thb + s\boldsymbol{\mu}, \sigma^2\mathbf{I})$, $\boldsymbol{\mu} = \mathbf{1}$ and $\sigma^2=0.2^2$ \\ [5pt]
  \textbf{Dimensionality} & $\thb \in \mathbb{R}^{D}$, $\yb \in \mathbb{R}^{D}$ \\ [5pt]
  \textbf{Observation} & $\data = (0, \ldots, 0)$ where $\data \in \mathbb{R}^D$ \\ [5pt]
  \textbf{Posterior} & $ p(\thb | \data) \propto
                                    \begin{cases}
                                      \mathcal{N}(\thb; \sum_{s \in \{-1, +1\}} \mathcal{N}(s\boldsymbol{\mu}, \sigma^2 \mathbf{I}), & \text{if } \thb \in [-3, 3]^D, \\
                                      0, & \text{otherwise}.
                                    \end{cases} $ \\ [5pt]
\end{tabular}

The MoG simulator is the same as in the introductory example, so the proof is equivalent to Section~\ref{app-sec:intro-example}.

\textbf{MoG simulator - with distractors (4):}

\begin{tabular}{@{}ll@{}}
  \textbf{Prior} & $\thb \sim \mathcal{U}(-\mathbf{3}, \mathbf{3})$ \\ [5pt]
  \textbf{Simulator} & $\yb=(\mathbf{y^{(1)}}, \mathbf{y^{(2)}})$,\\[5pt]
  &$\mathbf{y^{(1)}} \sim \frac{1}{2} \sum_{s \in \{-1, +1\}} \mathcal{N}(\thb + s\boldsymbol{\mu}, \sigma^2\mathbf{I})$, $\boldsymbol{\mu} = \mathbf{1}$ and $\sigma^2=0.2^2$ \\[5pt]
  &$\mathbf{y^{(2)}} \sim \mathcal{U}(\mathbf{-3}, \mathbf{3})$ \\ [5pt]
  \textbf{Dimensionality} & $\thb \in \mathbb{R}^{D}$, $\yb \in \mathbb{R}^{D + 18}$, $\yb^{(1)} \in \mathbb{R}^{D}$, $\yb^{(2)} \in \mathbb{R}^{18}$ \\ [5pt]
  \textbf{Observation} & $\data = (\mathbf{y^{o, (1)}}, \mathbf{y^{o, (2)}})$, where both $\mathbf{y^{o, (1)}}$ and $\mathbf{y^{o, (2)}}$  are $(0, \ldots, 0) $ \\ [5pt]
  \textbf{Posterior} & $ p(\thb | \data) \propto
                       \begin{cases}
                         \mathcal{N}(\thb; \sum_{s \in \{-1, +1\}} \mathcal{N}(s\boldsymbol{\mu}, \sigma^2 \mathbf{I}), & \text{if } \thb \in [-3, 3]^D, \\
                                      0, & \text{otherwise}.
                                    \end{cases} $ \\ [5pt]
\end{tabular}

The MoG simulator is the same as in the introductory example, so the proof is equivalent to Section~\ref{app-sec:intro-example}.

\subsubsection{Experimental Setup and Hyperparameter Configuration}

For each of the four problems (1)--(4) and across all dimensionalities 
$D \in \{2, 5, 10, 15, 20\}$, we evaluated all methods under simulation budgets of
\(
[1{,}000, 5{,}000, 10{,}000, 50{,}000, 100{,}000],
\)
recording both the C2ST score and the runtime.

To reduce experimental cost, we adopted an early-stopping criterion: if a method achieved a C2ST score below 0.75 for a given dimensionality $D$ at a certain budget, we did not evaluate it at higher budgets. Each experiment was repeated with three independent random seeds, and we report the average C2ST scores across these runs.
For each method, we drew 1{,}000 samples from the approximate posterior and compared them against 1{,}000 ground-truth samples using the C2ST metric.

The goal of this experiment is to assess whether the required simulation budget---and consequently, the runtime---increase with problem dimensionality. To this end, we identify the minimum budget at which each method achieves successful inference, defined as an average C2ST score below 0.75. We report both the mean C2ST and mean runtime across the three runs.

Below, we detail the hyperparameter configuration for each method, emphasizing deviations from the default settings.

\paragraph{ROMC} 
For \rromc, we used at most a maximum of $S = 1{,}000$ seeds, corresponding to roughly 1{,}000 independent simulator calls, as this budget was enough for a near-optimal C2ST score.
The modified fitting parameters are: \texttt{\{"pcg\_to\_keep":1., "epochs": 10, "alpha": 0.1\}} and the modified sampling parameters are: \texttt{\{"samples\_per\_region": 2\}}.

\paragraph{NPE-C} 
The modified fitting parameters are:
\texttt{\{"num\_transforms":16, "num\_bins":16 "hidden\_features":200, "training\_batch\_size": 500\}}

\paragraph{BayesFlow} 

The modified fitting parameters are:
\texttt{\{
  "batch\_size": 500,
  "training\_batch\_size": 500,
  "embedding\_net\_num\_layers": 2,
  "embedding\_net\_num\_hiddens": 32
  \}}.
Finally, the parameter
\texttt{\{"embedding\_net\_output\_dim": dim\}} is for the non-distractor variants ((1), (3)) and \texttt{dim + 18} for the distractor variants ((2), (4)).

\paragraph{Flow-Matching}
The modified fitting parameters are:
\texttt{\{"batch\_size": 100, "training\_batch\_size": 100\}}.

\subsubsection{Results}

\begin{figure*}[!t]
  \centering
  \begin{subfigure}[b]{0.245\textwidth}
    \includegraphics[width=\linewidth]{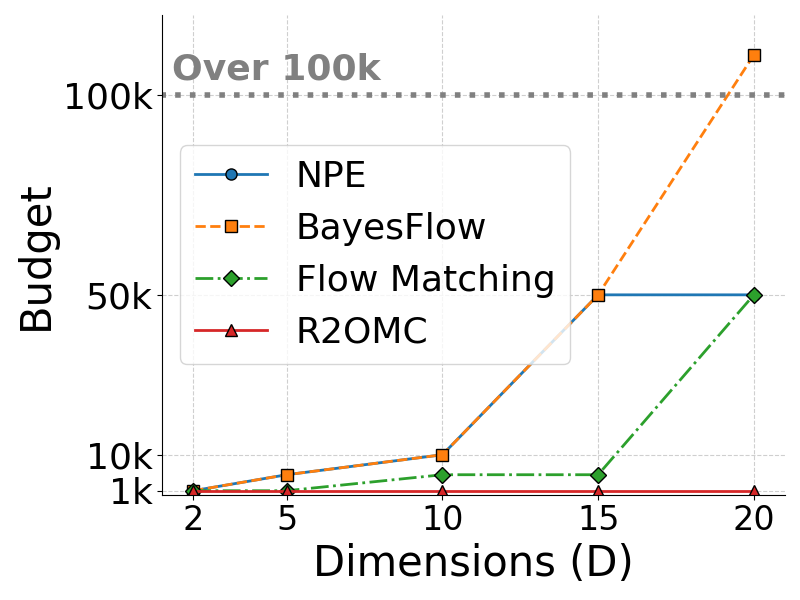}
    \caption{$S_{\mathtt{base}}$}
  \end{subfigure}
  \begin{subfigure}[b]{0.245\textwidth}
    \includegraphics[width=\linewidth]{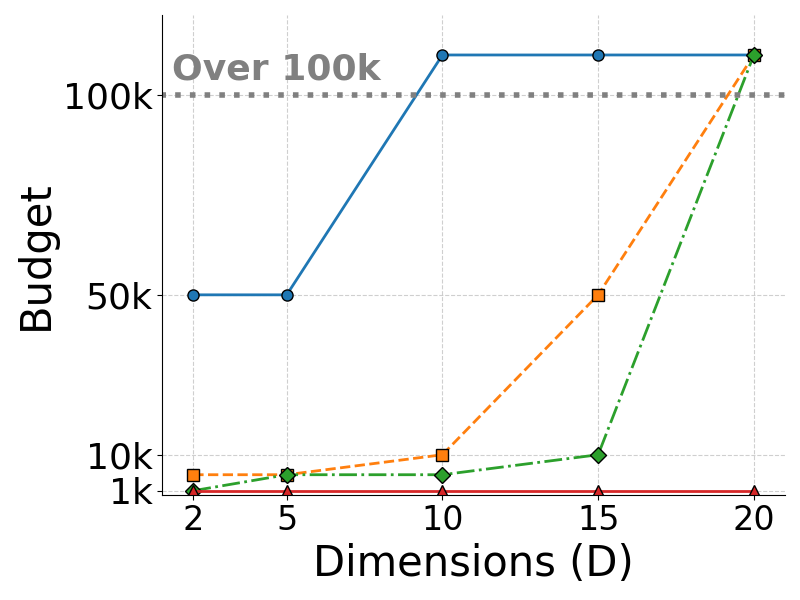}
    \caption{$S_{\mathtt{base}}^{\mathtt{dist}}$}
  \end{subfigure}
  \begin{subfigure}[b]{0.245\textwidth}
    \includegraphics[width=\linewidth]{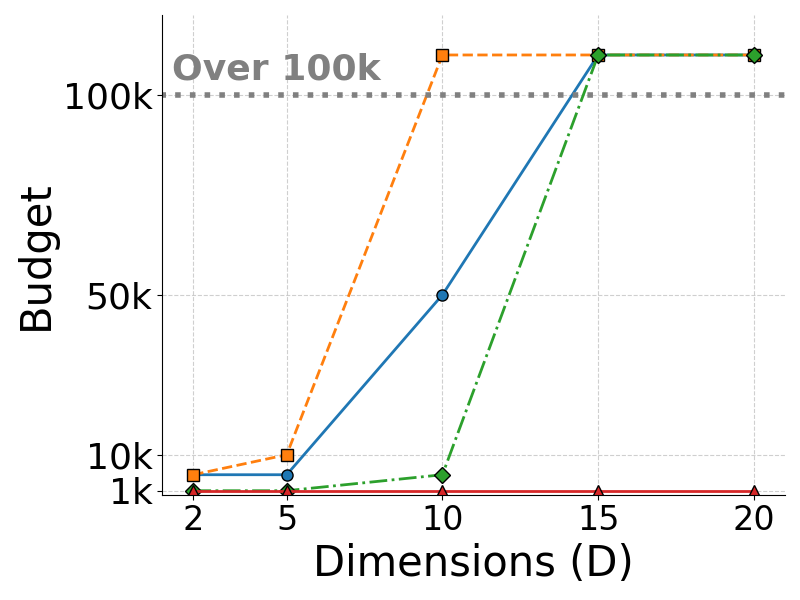}
    \caption{$S_{\mathtt{MoG}}$}
  \end{subfigure}
  \begin{subfigure}[b]{0.245\textwidth}
    \includegraphics[width=\linewidth]{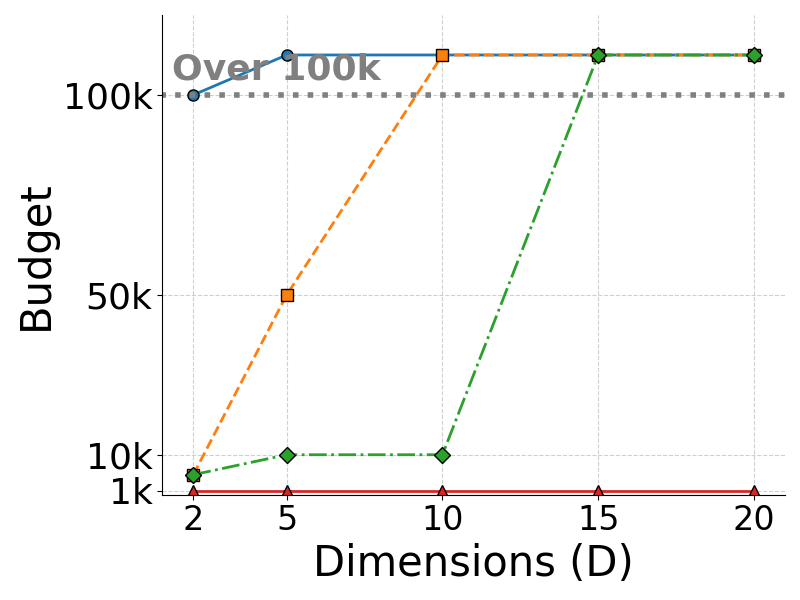}
    \caption{$S^{\mathtt{dist}}_{\mathtt{MoG}}$}
  \end{subfigure}
  \caption{MoG benchmark: Success Frontier plots showing the lowest budget required to reach a C2ST score $\leq 0.75$ for varying $D$.}
  \label{fig:mog_benchmark_frontier_budget}
\end{figure*}

Figure~\ref{fig:high_dim_benchmark} in the main paper summarizes the results using success frontier plots, which depict the minimum runtime required to achieve successful inference across different dimensions $D$. 
Inference is deemed successful when the mean \CtwoST~score is less than or equal to 0.75. 
The corresponding plots using the simulation budget (instead of runtime) are shown in Figure~\ref{fig:mog_benchmark_frontier_budget}. 
This figure complements Figure~\ref{fig:high_dim_benchmark} validating that the factor that incurs longer runtimes is the increased simulation budget required at higher dimensionalities.

\begin{figure}
    \centering
    \includegraphics[width=.24\linewidth]{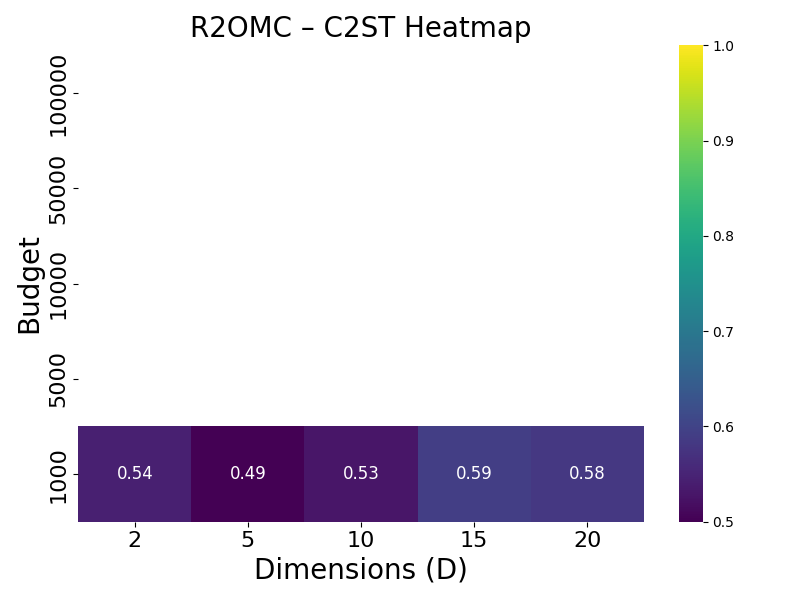}
    \includegraphics[width=.24\linewidth]{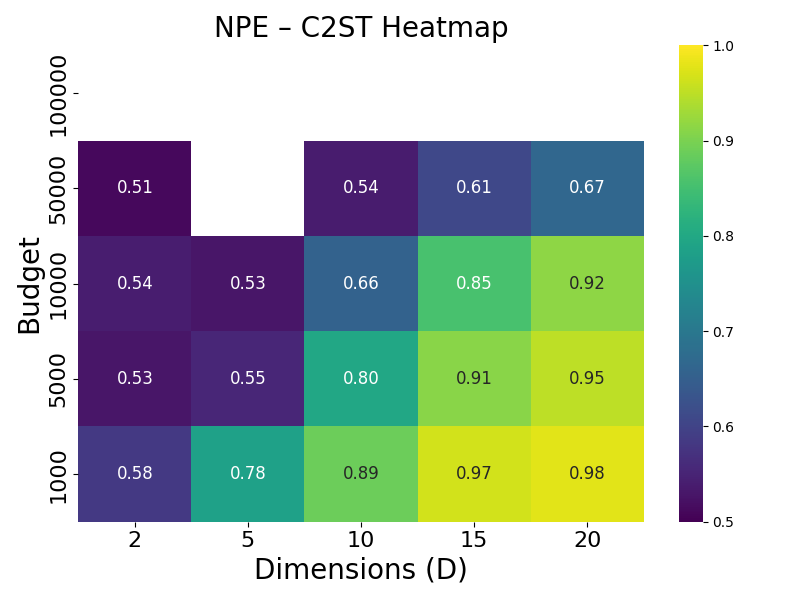}
    \includegraphics[width=.24\linewidth]{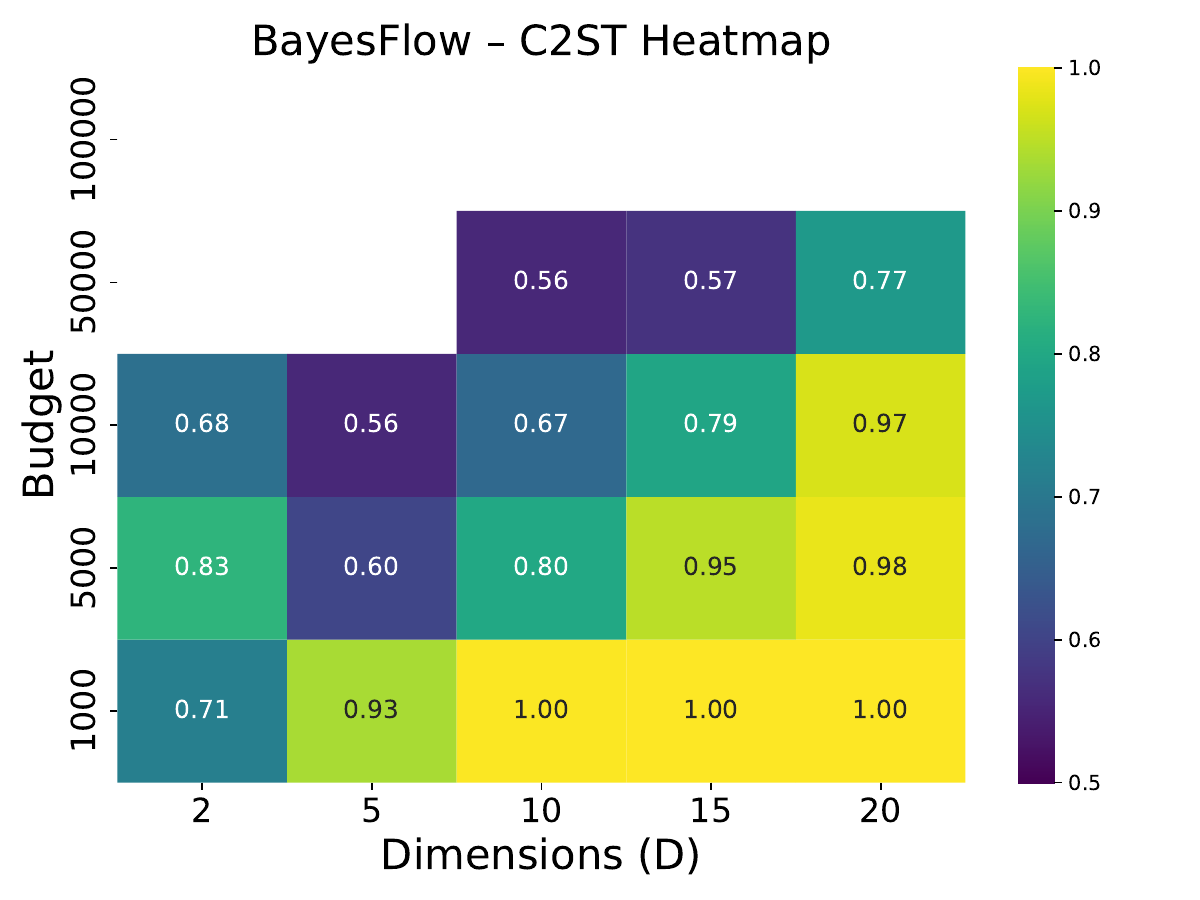}
    \includegraphics[width=.24\linewidth]{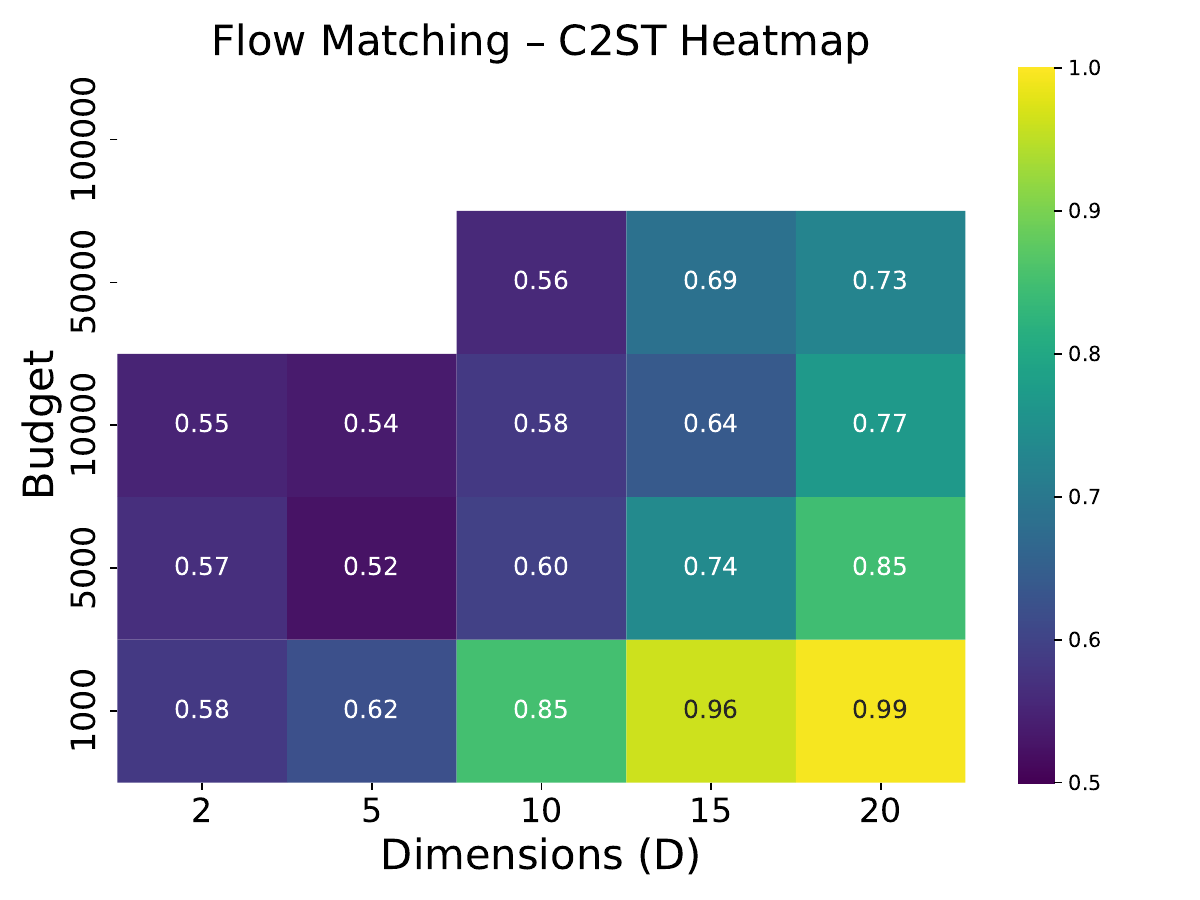}
    \caption{MoG Benchmark - Base simulator: C2ST heatmaps for R2OMC, NPE, BayesFlow, and Flow Matching.}
    \label{fig:mog_benchmark_base_heatmaps}
\end{figure}

\begin{figure}
    \centering
    \includegraphics[width=.24\linewidth]{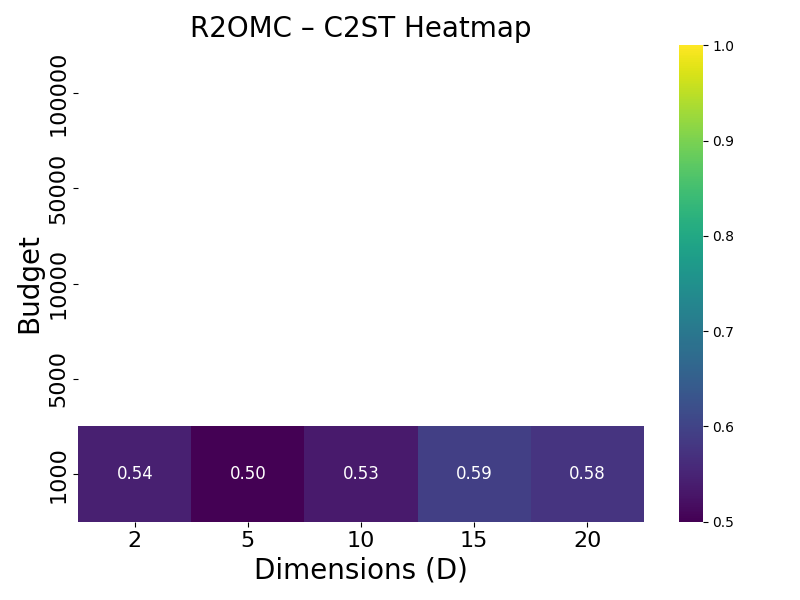}
    \includegraphics[width=.24\linewidth]{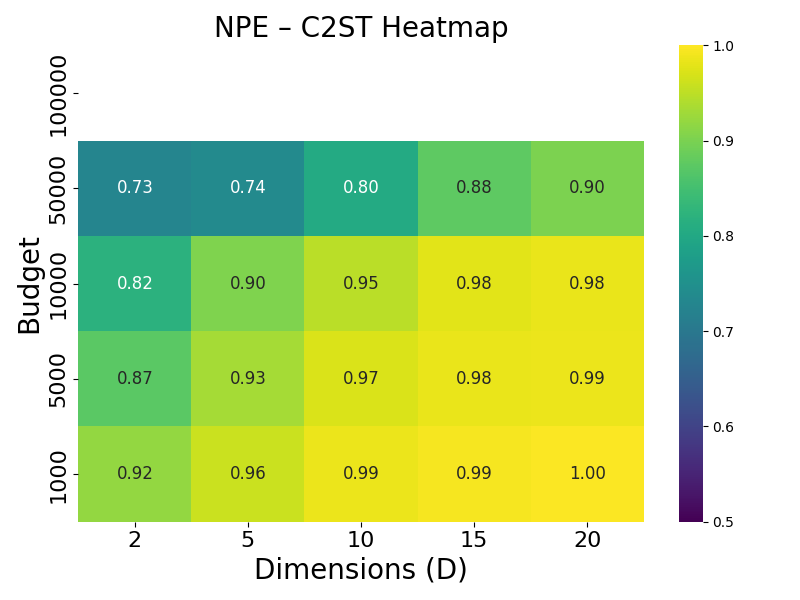}
    \includegraphics[width=.24\linewidth]{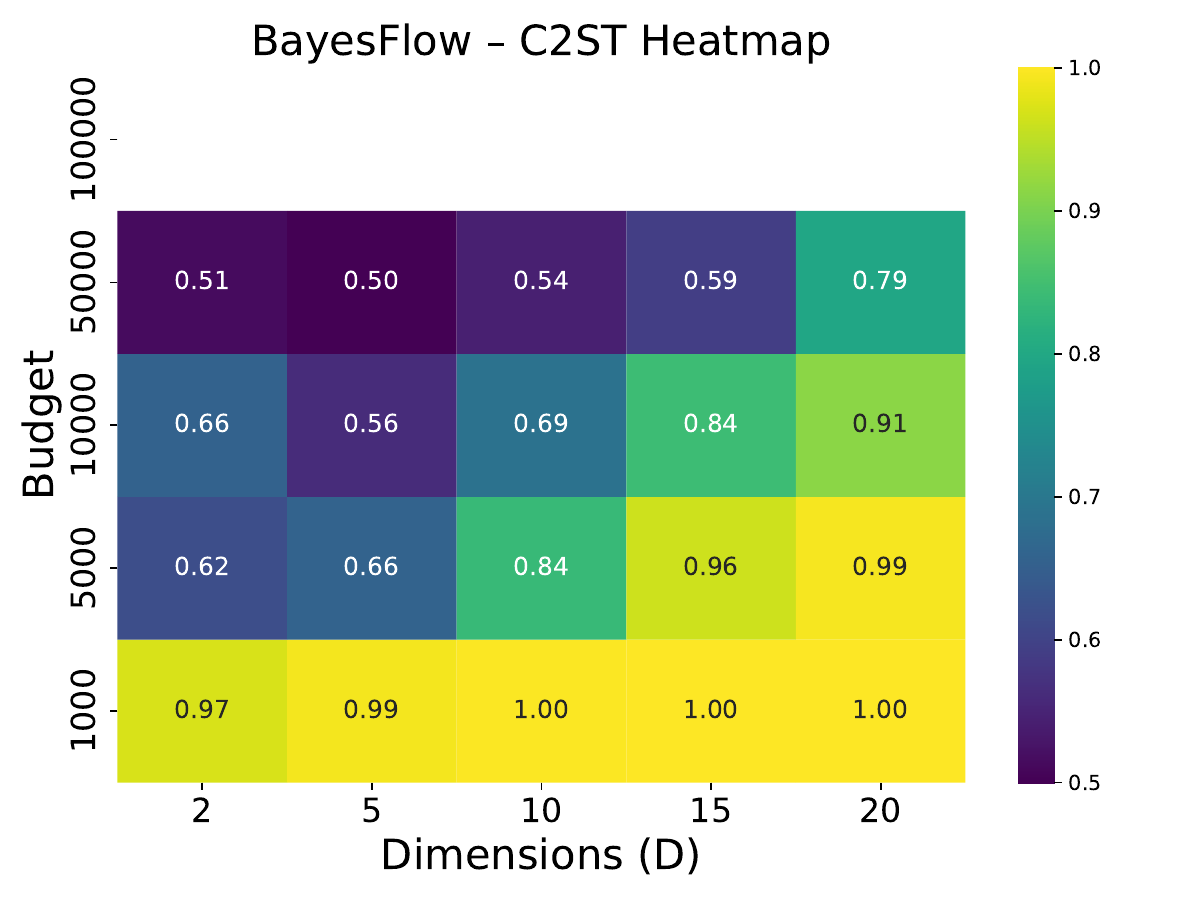}
    \includegraphics[width=.24\linewidth]{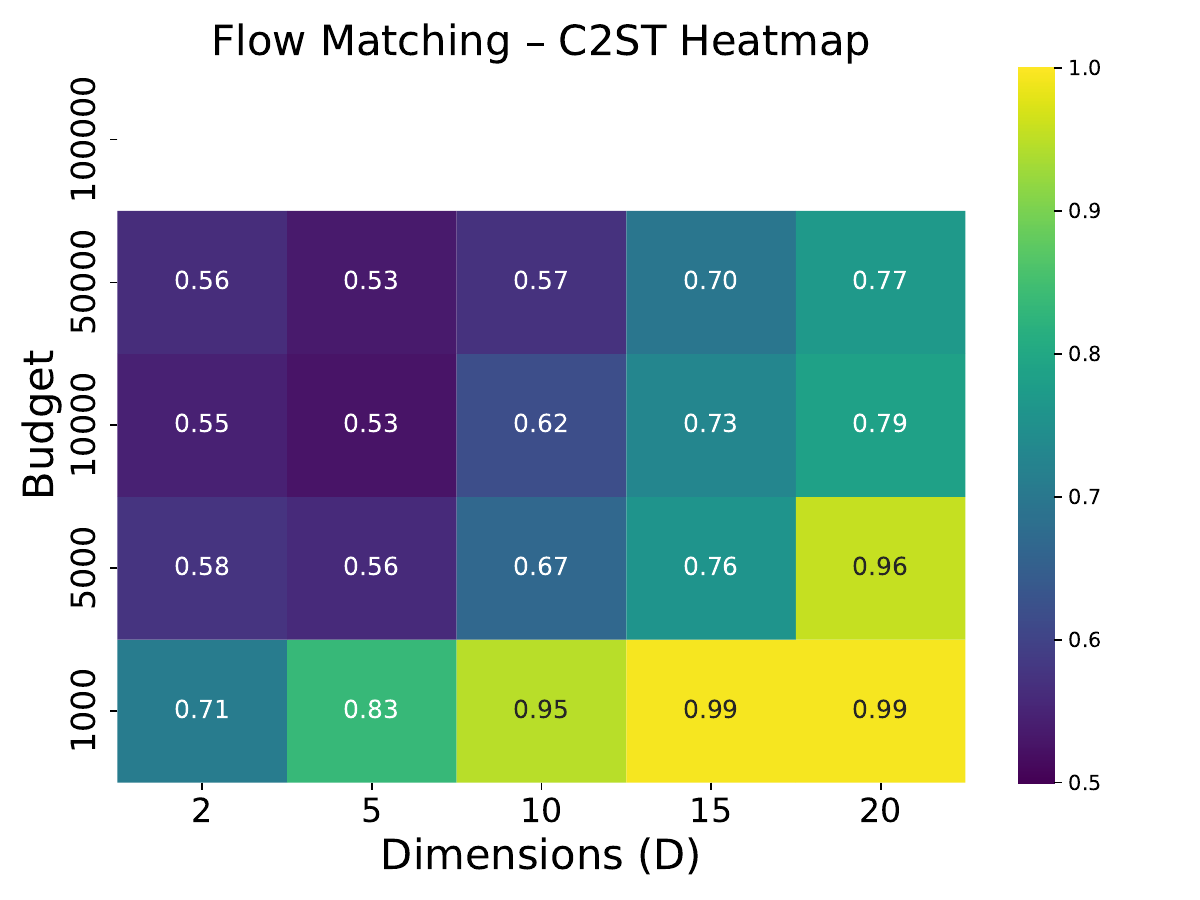}
    \caption{MoG Benchmark - Base simulator with distractors: C2ST heatmaps for R2OMC, NPE, BayesFlow, and Flow Matching.}
    \label{fig:mog_benchmark_base_distractors_heatmaps}
\end{figure}

\begin{figure}
    \centering
    \includegraphics[width=.24\linewidth]{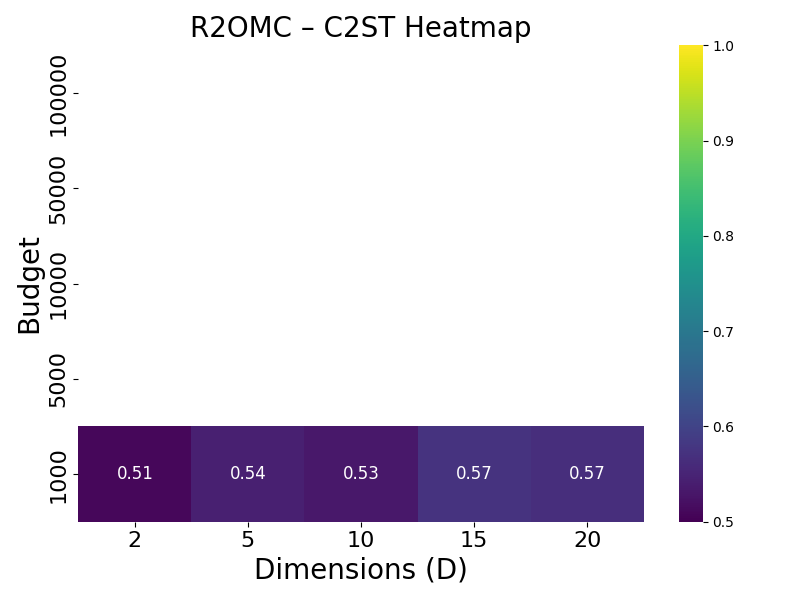}
    \includegraphics[width=.24\linewidth]{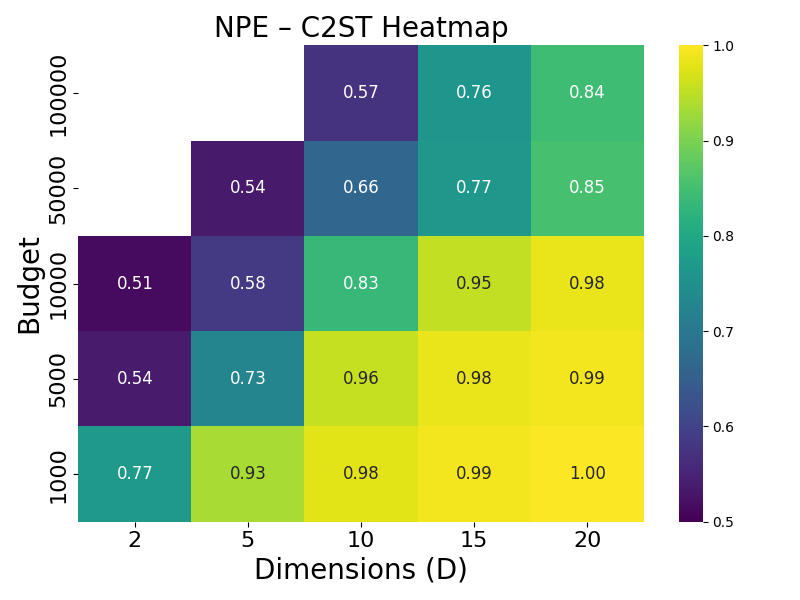}
    \includegraphics[width=.24\linewidth]{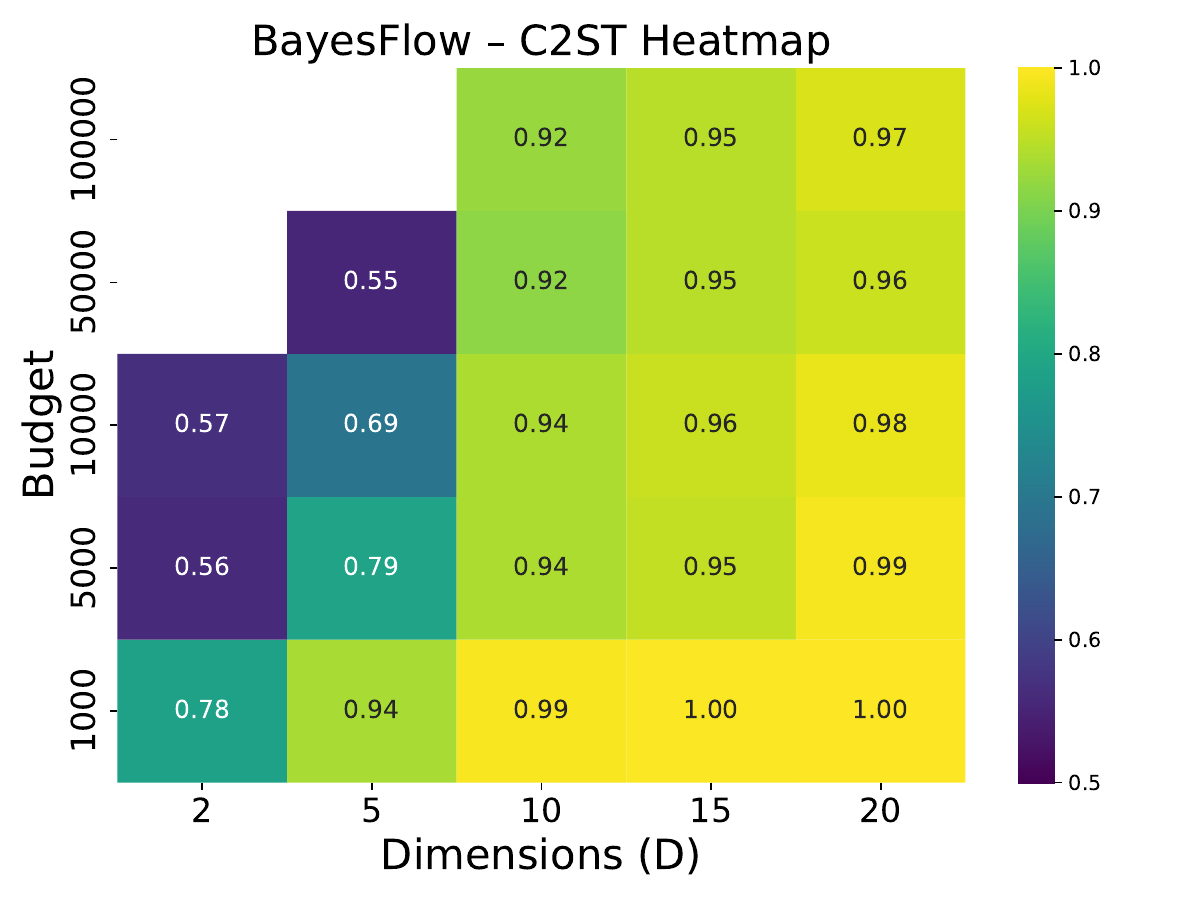}
    \includegraphics[width=.24\linewidth]{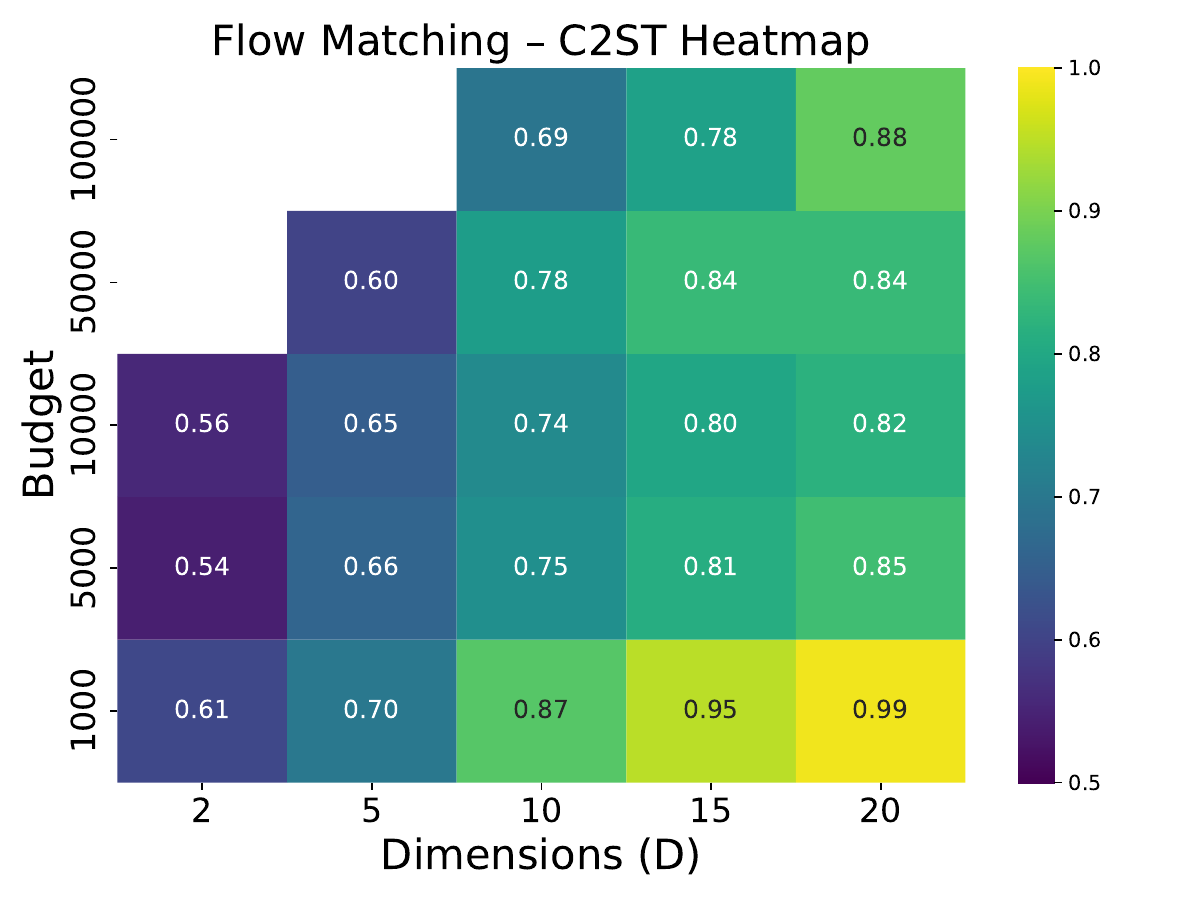}
    \caption{MoG Benchmark - MoG simulator: C2ST heatmaps for R2OMC, NPE, BayesFlow, and Flow Matching.}
    \label{fig:mog_benchmark_mog_heatmaps}
\end{figure}

\begin{figure}
    \centering
    \includegraphics[width=.24\linewidth]{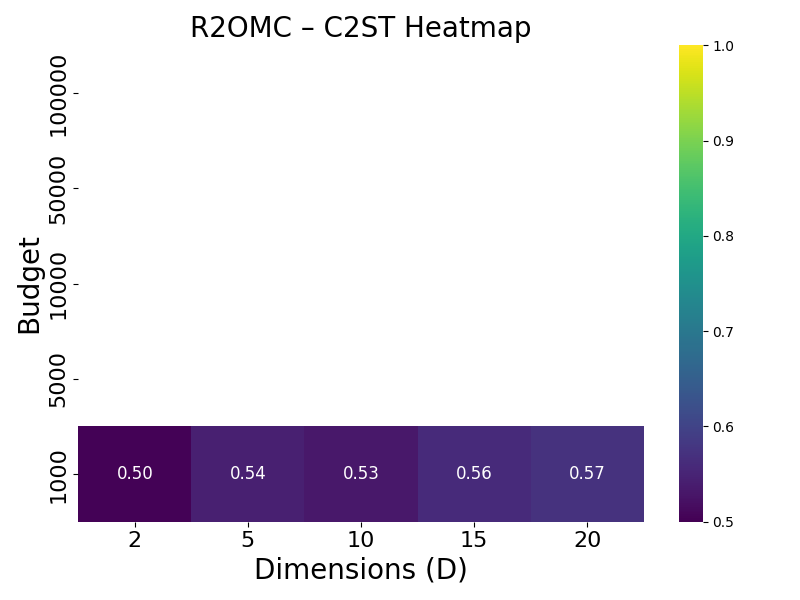}
    \includegraphics[width=.24\linewidth]{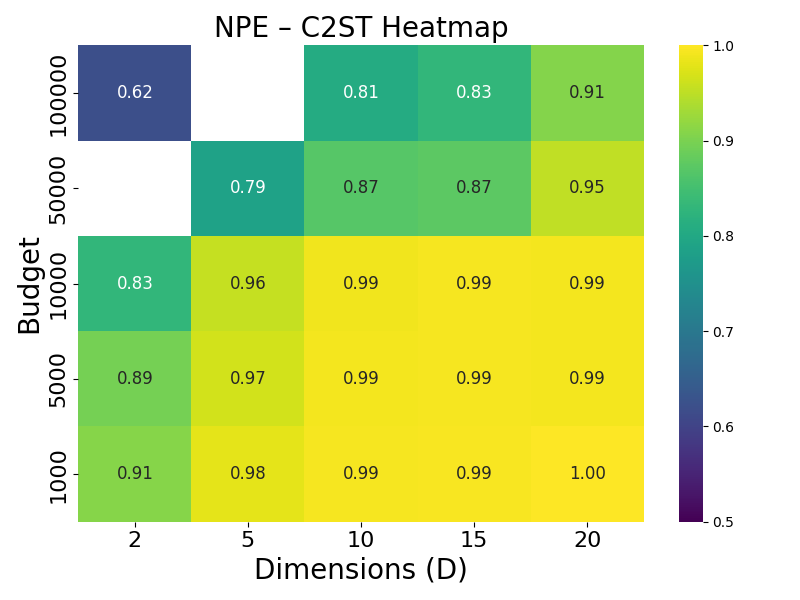}
    \includegraphics[width=.24\linewidth]{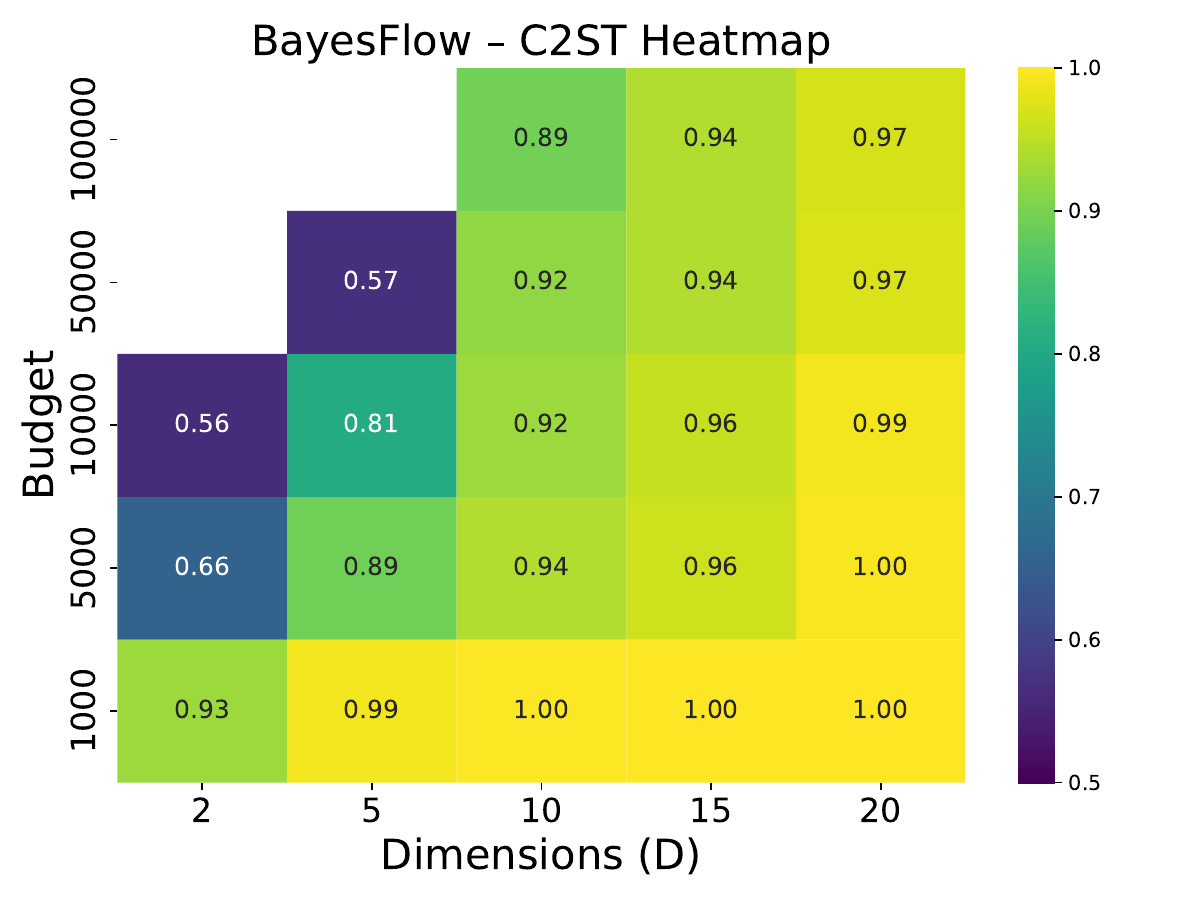}
    \includegraphics[width=.24\linewidth]{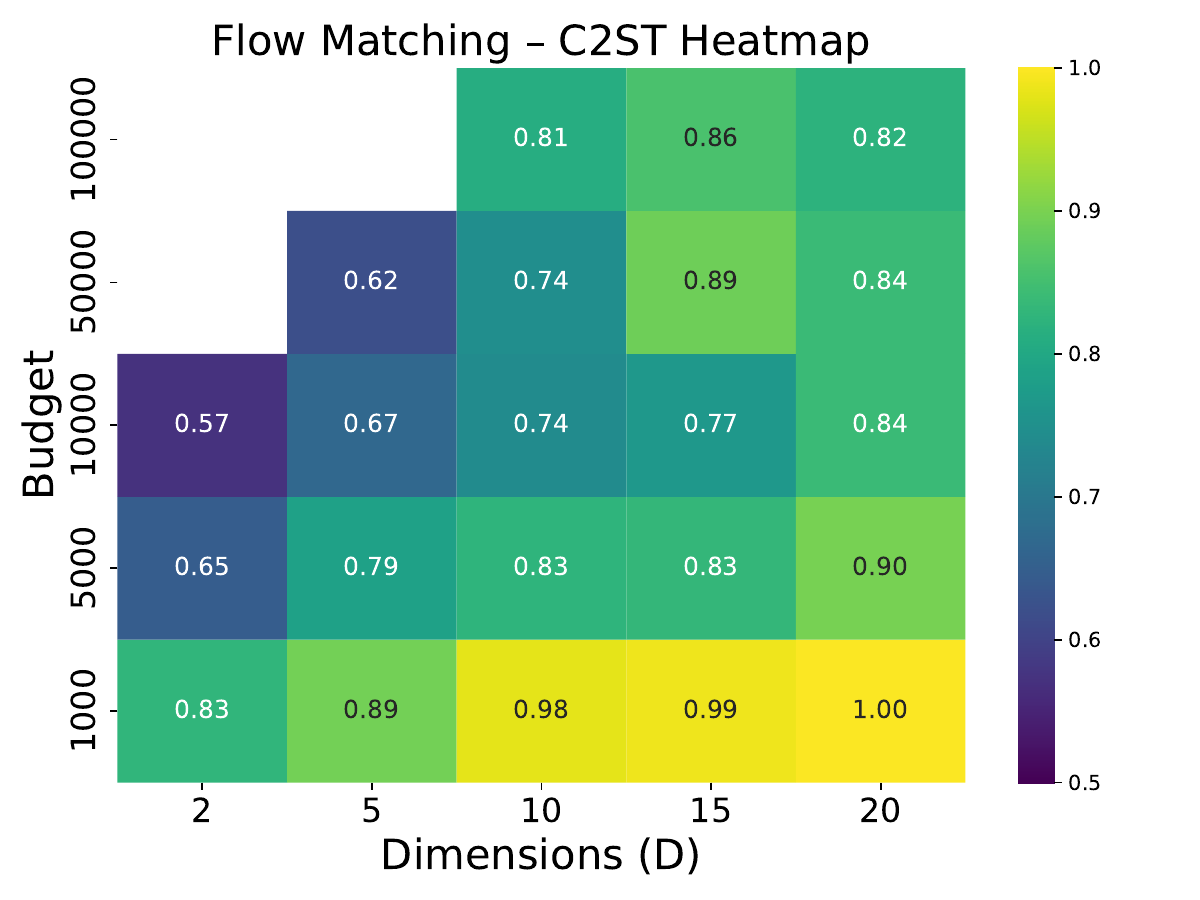}
    \caption{MoG Benchmark - MoG simulator with distractors: C2ST heatmaps for R2OMC, NPE, BayesFlow, and Flow Matching.}
    \label{fig:mog_benchmark_mog_distractors_heatmaps}
\end{figure}

For a more detailed view, Figures~%
\ref{fig:mog_benchmark_base_heatmaps} ($\mathcal{S}_{\mathtt{base}}$), 
\ref{fig:mog_benchmark_base_distractors_heatmaps} ($\mathcal{S}_{\mathtt{base}}^{\mathtt{dist}}$), 
\ref{fig:mog_benchmark_mog_heatmaps} ($\mathcal{S}_{\mathtt{MoG}}$), and 
\ref{fig:mog_benchmark_mog_distractors_heatmaps} ($\mathcal{S}_{\mathtt{MoG}}^{\mathtt{dist}}$) 
report the mean \CtwoST~score for each method across all dimensionalities and budgets.

All neural-based methods exhibit a consistent trend: as $D$ increases, the simulation budget required for accurate posterior inference grows rapidly, often exceeding $100{,}000$ simulations and resulting in runtimes that often exceed one hour. 
Runtimes increase more sharply for $S_{\mathtt{MoG}}$ compared to $S_{\mathtt{base}}$, and are higher in the distractor settings than in the simpler ones. 

In contrast, \rromc~achieves successful inference within a few seconds across all dimensionalities and simulator configurations.

\subsection{SBI - SLCP (T.3) and SLCP Distractors (T.4)}

We here provide additional information on example T.3 and T.4 of the SBI benchmark~\cite{lueckmann2021benchmarking}.

SLCP (T.3):

\begin{tabular}{@{}ll@{}}
  \textbf{Prior} & $\thb \sim \mathcal{U}(-\mathbf{3}, \mathbf{3})$ \\ [5pt]
  \textbf{Simulator} & $\yb = (\boldsymbol{y_1}, \ldots, \boldsymbol{y_4}), \boldsymbol{y_i} \sim \mathcal{N}(\boldsymbol{m_{\theta}}, \boldsymbol{S_{\theta}}) \in \mathbb{R}^2 $ \\ [5pt]
                 & where: $\boldsymbol{m_{\theta}} \sim \begin{bmatrix}  \theta_1 \\ \theta_2 \end{bmatrix},
                   \boldsymbol{S_{\theta}} = \begin{bmatrix} s_1 & \rho s_1 s_2 \\ \rho s_1 s_2 & s_2^2  \end{bmatrix}$,
                   where $s_1 = \theta_3^2$, $s_2 = \theta_4^2$ and $\rho = \tanh \theta_5$ \\ [10pt]
  \textbf{Dimensionality} & $\thb \in \mathbb{R}^{5}$, $\boldsymbol{y_i} \in \mathbb{R}^{2}$, $\yb \in \mathbb{R}^{8}$ \\ [5pt]
\end{tabular}

SLCP with distractors (T.4):

\begin{tabular}{@{}ll@{}}
  \textbf{Prior} & $\thb \sim \mathcal{U}(-\mathbf{3}, \mathbf{3})$ \\ [5pt]
  \textbf{Simulator} & $\yb = (\boldsymbol{y_1}, \ldots, \boldsymbol{y_4})$, $\boldsymbol{y_i} = (\boldsymbol{y_i^{(0)}}, \boldsymbol{y_i^{(1)}})$ where $\boldsymbol{y_i^{(0)}} \sim \mathcal{N}(\boldsymbol{m_{\theta}}, \boldsymbol{S_{\theta}}) $ \\ [5pt]
                 & where: $\boldsymbol{m_{\theta}} \sim \begin{bmatrix}  \theta_1 \\ \theta_2 \end{bmatrix},
                   \boldsymbol{S_{\theta}} = \begin{bmatrix} s_1 & \rho s_1 s_2 \\ \rho s_1 s_2 & s_2^2  \end{bmatrix}$,
                   $s_1 = \theta_3^2$, $s_2 = \theta_4^2$ and $\rho = \tanh \theta_5$ \\ [10pt]
                 & and $\boldsymbol{y_i^{(1)}}$ comes from a distribution (independent of $\thb$) analyzed in \citep{lueckmann2021benchmarking}. \\ [5pt]
  \textbf{Dimensionality} & $\thb \in \mathbb{R}^{5}$, $\boldsymbol{y_i} \in \mathbb{R}^{100}$, $\boldsymbol{y_i} \in \mathbb{R}^{25}$, $\boldsymbol{y_i^{(0)}} \in \mathbb{R}^{2}$, $\boldsymbol{y_i^{(0)}} \in \mathbb{R}^{23}$ \\ [5pt]
\end{tabular}

The ground truth posterior is not available in closed form. Instead a set of $10{,}000$ posterior samples is used

\subsubsection{Experimental Setup and Results}

We evaluate \rromc~using simulation budgets (i.e., numbers of seeds) of 1{,}000, 1{,}500, 5{,}000, 10{,}000, and 30{,}000. 
The SBI benchmark provides 10{,}000 ground-truth posterior samples for each of eight distinct observations. 
We run \rromc~on all observations and report the mean \CtwoST~score and mean runtime across them.

\begin{figure}
\centering
\includegraphics[width=.24\linewidth]{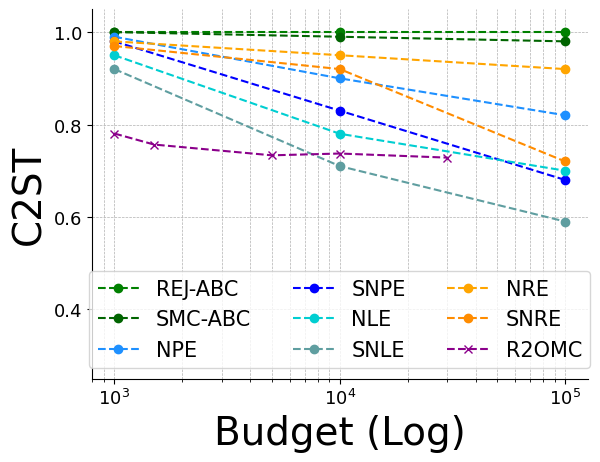}
\includegraphics[width=.24\linewidth]{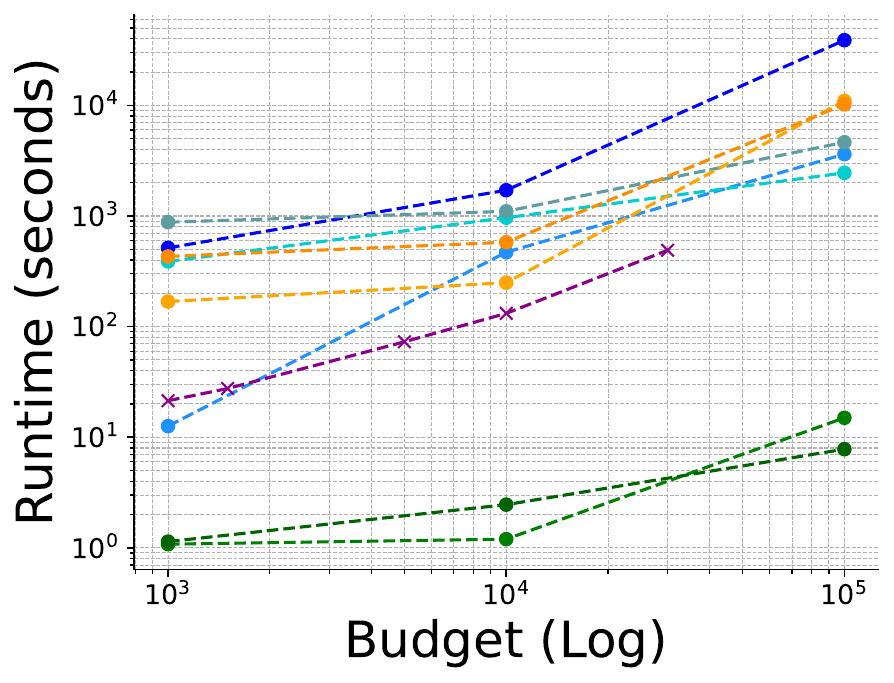}
\includegraphics[width=.24\linewidth]{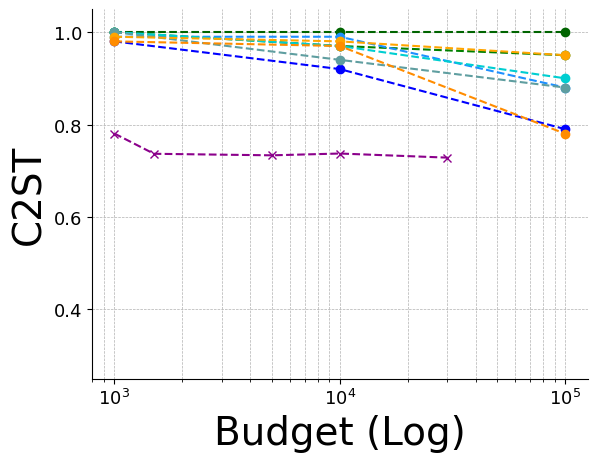}
\includegraphics[width=.24\linewidth]{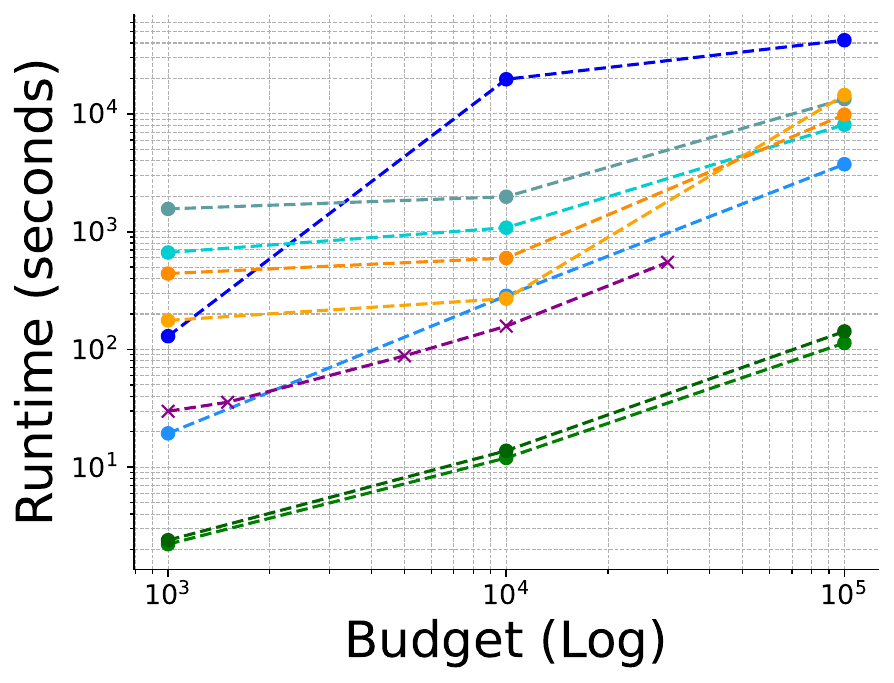}
\caption{From left to right: \texttt{C2ST} vs. budget, runtime vs. budget. for T.3: SLCP and same figures for (T.4: SLCP Distractors)}
\label{fig:exp_sbi_slcp}
\end{figure}

Figure~\ref{fig:slcp_c2st_runtime} in the main paper presents the \CtwoST--runtime plots for both T.3 and T.4 tasks. 
To complement these, Figure~\ref{fig:exp_sbi_slcp} shows the \CtwoST--budget and runtime--budget relationships for the same tasks. 
These results confirm that the increase in runtime is primarily driven by the larger simulation budgets required for accurate inference.

\begin{figure}
    \centering
    \includegraphics[width=.32\linewidth]{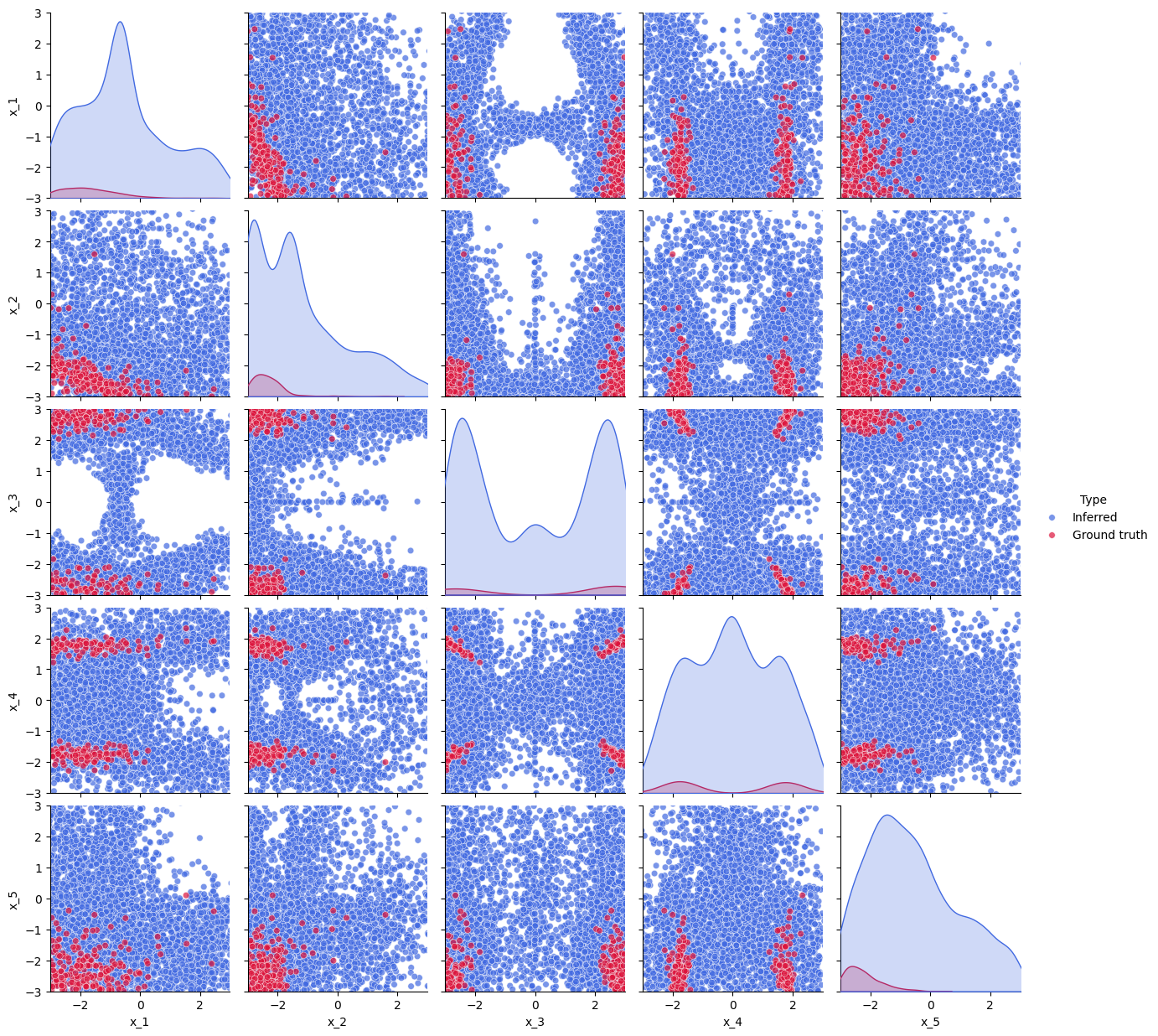}
    \includegraphics[width=.32\linewidth]{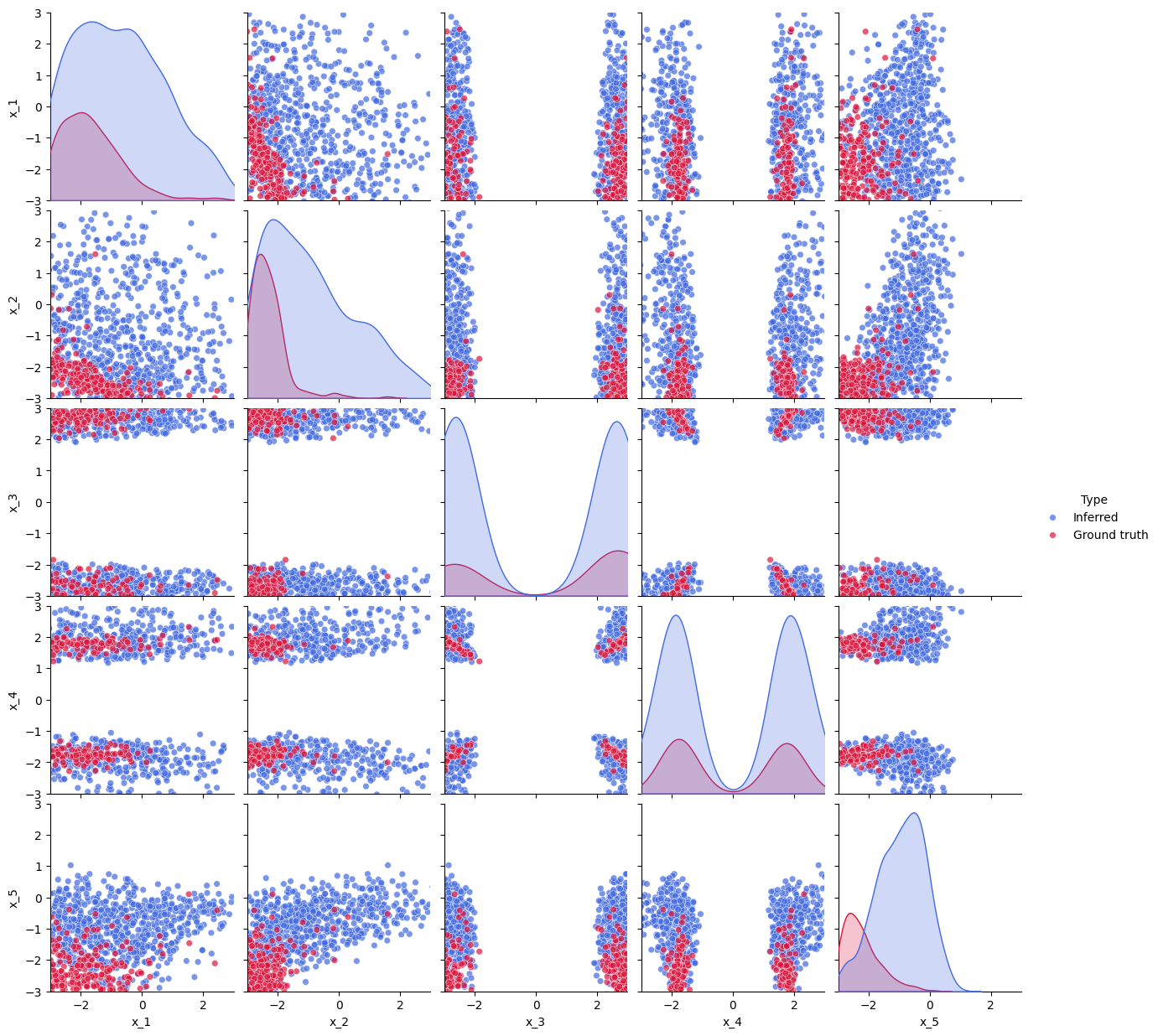}
    \includegraphics[width=.32\linewidth]{./figures/sbibm/slcp/pairwise_posterior_selected}
    \caption{T.3: SLCP. From left to right: total samples, accepted samples, and selected samples by R2OMC.}
    \label{fig:exp_sbi_slcp_samples}
\end{figure}

Figure~\ref{fig:slcp_posterior} in the main paper illustrates how \rromc~handles multiple observations. 
Proposal samples are first selected per observation (within an $\epsilon$-distance) and then reweighted according to Eq.~\ref{eq:rromc_weight} to retain those most consistent across all observations. 
In Figure~\ref{fig:exp_sbi_slcp_samples}, we visualize all proposal samples from all observations (left panel), the subset that receives positive weights (middle panel), and the final accepted samples (right panel).

Overall, results demonstrate that \rromc~accurately approximates the posterior, achieving mean \CtwoST~scores in the range of 0.7--0.8 within only a few seconds. 
In contrast, competing methods require substantially higher budgets (typically $10^4$--$10^5$ simulations) and correspondingly longer runtimes—often several hours—to reach comparable performance. 
The performance gap becomes particularly pronounced in distractor settings, where most competing methods fail to achieve \CtwoST~scores below 0.8 regardless of the simulation budget.

\subsection{SBI - Two Moons (T.8)}

We here provide additional information on example T.8 of the SBI benchmark~\cite{lueckmann2021benchmarking}.
It tests inference when the posterior exhibits both global (bimodality) and local (crescent shape) structure to illustrate how algorithms deal with multimodality:

\begin{tabular}{@{}ll@{}}
  \textbf{Prior} & $\thb \sim \mathcal{U}(-\mathbf{1}, \mathbf{1})$ \\ [5pt]
  \textbf{Simulator} & $\yb \sim \begin{bmatrix} 
    r \cos(\alpha) + 0.25 \\ 
    r \sin(\alpha)
  \end{bmatrix} + \begin{bmatrix} 
    |\theta_1 + \theta_2|/\sqrt{2} \\ 
    (-\theta_1 + \theta_2)/\sqrt{2}
  \end{bmatrix}$, where $r \sim \mathcal{N}(0.1, 0.01^2)$ and $\alpha \sim \mathcal{U}(-\pi/2, \pi/2)$ \\ [10pt]
  \textbf{Dimensionality} & $\thb \in \mathbb{R}^{2}$, $\yb \in \mathbb{R}^{2}$ \\ [5pt]
\end{tabular}

The ground truth posterior is not available in closed form. Instead a set of $10{,}000$ posterior samples is used

\subsubsection{Experimental Setup and Results}

\begin{figure}[ht]
  \centering
  \begin{subfigure}[b]{0.45\linewidth}
    \centering
    \includegraphics[width=\linewidth]{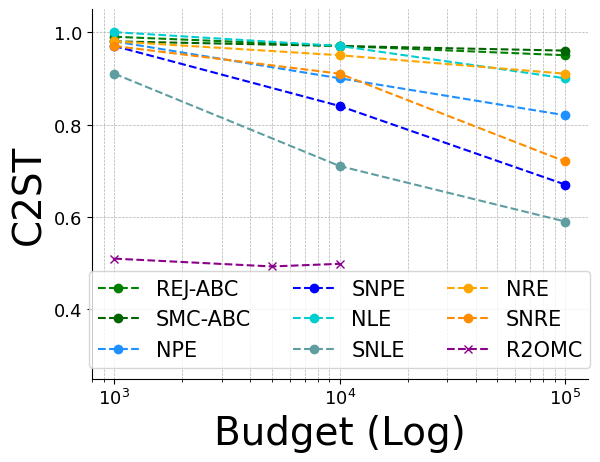}
    \caption{\texttt{C2ST} score}
  \end{subfigure}
  \hfill
  \begin{subfigure}[b]{0.45\linewidth}
    \centering
    \includegraphics[width=\linewidth]{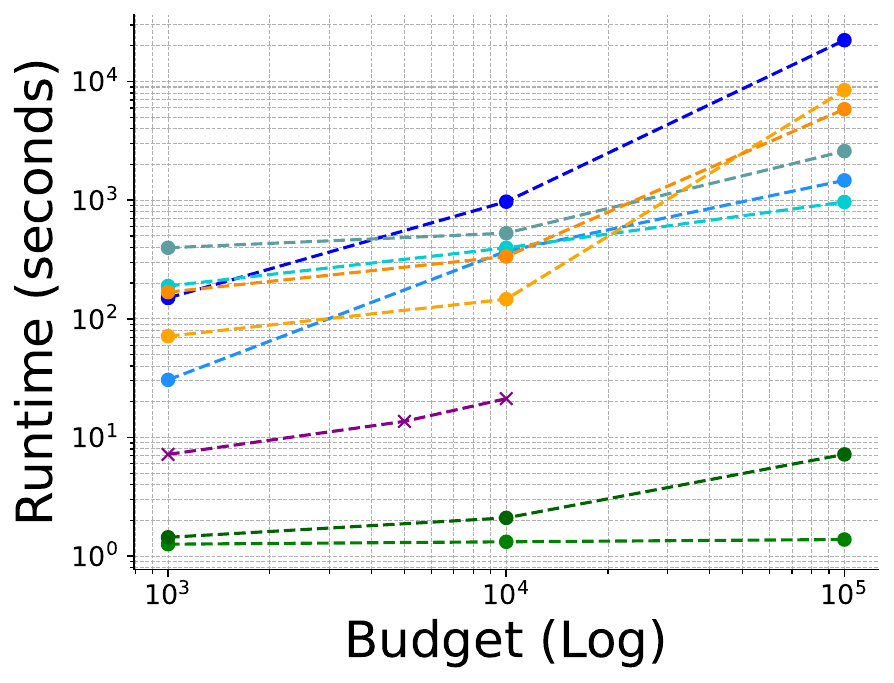}
    \caption{Ground truth samples}
  \end{subfigure}
  \caption{T.8: Two Moons. From left to right: \texttt{C2ST} score, ground truth samples, and samples by R2OMC.}
  \label{fig:exp_sbi_two_moons}
\end{figure}

We test ROMC for simulation budget (number of seeds) equal to 1000, 5000, 10000.
SBI benchamrk provides a set of 10{,}000 gound-truth posterior samples for each of the 8 different observations.
We run ROMC in all observations and we report the mean C2ST score and the mean runtime across themm.
Figure~\ref{fig:sbi_two_moons} in the main paper shows the C2ST vs. runtime plot and the pairwise posterior plot.
To compliment that, in Figure~\ref{fig:exp_sbi_two_moons} we provide the C2T vs. budget plot and the Budget vs. Rutime.
These plot confirm that the reason behind higher duntimes is the increase in the budget. 

\rromc~achieves a near-optimal \CtwoST score ($\approx 0.5$) in a few-seconds runtime as 1000 different seeds $S$ are already enough for a good posterior approximaiont.
Competing methods require minutes to hours (budgets of $10^5$ samples) to reach comparable performance (which never becomes that good)
The pairwise posterior further demonstrates that \rromc~accurately captures both crescent-shaped modes.

\clearpage